\documentclass{article}

\PassOptionsToPackage{hyphens}{url}
\usepackage[english]{babel}
\usepackage[letterpaper,top=2cm,bottom=2cm,left=3cm,right=3cm,marginparwidth=1.75cm]{geometry}
\usepackage{booktabs}
\usepackage{amssymb}
\usepackage{cellspace}
\usepackage{amsmath}
\usepackage{graphicx}
\usepackage{epstopdf}
\AppendGraphicsExtensions{.tif}
\usepackage{comment}
\usepackage{dirtytalk}
\usepackage{epsfig}
\usepackage{subcaption,booktabs}
\usepackage{multirow}
\usepackage{adjustbox}
\usepackage{authblk}
\usepackage{lineno}
\usepackage[version=4]{mhchem}
\usepackage[colorlinks=true, allcolors=blue]{hyperref}
\usepackage{float}

\title{DINOv2 Rocks Geological Image Analysis: Classification, Segmentation, and Interpretability}

\author[1,*, $\dagger$]{Florent Brondolo}
\author[1,*]{Samuel Beaussant}
\affil[1]{Akkodis Research, \texttt{florent.brondolo@akkodis.com}, \texttt{samuel.beaussant@akkodis.com}}

\affil[$\dagger$]{Project lead}
\affil[*]{Equal contributions}

\date{} 

\begin{document}
\maketitle
\sloppy
\begin{abstract}
Recent advancements in computer vision have significantly improved image analysis tasks. Yet, deep learning models often struggle when applied to domains outside their training distribution, such as in geosciences, where domain-specific data can be scarce. This study investigates the classification, segmentation, and interpretability of CT-scan images of rock samples, focusing on the application of modern computer vision techniques to geoscientific tasks. We compare a range of segmentation methods to assess their efficacy, efficiency, and adaptability in geological image analysis. The methods evaluated include Otsu thresholding, clustering techniques (K-means, fuzzy C-means), a supervised machine learning approach (Random Forest), and deep learning models (UNet, ResNet152, and DINOv2), using ten binary sandstone datasets and three multi-class calcite datasets. DINOv2 was selected for its promising results in feature extraction and its potential applicability in geoscientific tasks, prompting further assessment of its interpretability and effectiveness in processing CT-scanned rock data. For classification, a non-fine-tuned DINOv2 demonstrates strong performance in classifying rock images, even when the CT-scans are likely outside its original training set. In segmentation tasks, thresholding and clustering techniques, though computationally efficient, produce subpar results despite preprocessing efforts. In contrast, supervised methods achieve better performance without the need for preprocessing. While deep learning methods demand greater computational resources, they require minimal intervention and offer superior generalization. A LoRA fine-tuned DINOv2, in particular, excels in out-of-distribution segmentation and outperforms other methods in multi-class tasks, even with limited data. Notably, the segmentation masks generated by DINOv2 often appear more accurate than the original targets, based on visual inspection. Code is available at: \url{https://github.com/FloFive/DINOv2-X-Geosciences.git}.
\end{abstract}

\section{Introduction}
Digital rock physics is commonly used in various subsurface engineering fields such as Carbon Capture and Storage \cite{krevor2015, tatomir2016, krevor2017, bui2018} and Gas Storage \cite{henkel2014, jha2021}, which heavily rely on numerical analysis of the physical and chemical properties of samples \cite{iglauer2012, krevor2017, da2021, alhammadi2017, pak2015}. Central to this is micro-computed tomography (µCT), a non-destructive imaging technique that offers invaluable insights into the internal structure and properties of geological formations at micron scales \cite{mostaghimi2012, alhammadi2017, berg2018, al2018, krevor2017, an2023}. This allows for investigations such as heterogeneity characterization, pore network properties, fluid flow studies, and even contact angle measurements \cite{mostaghimi2012, an2023, bui2018, alhammadi2017, watt2024, main2024, blunt2017, pak2015}. Fluid injection and fluid mixing into rock samples can trigger changes within the rock fabric, which include fracturing, dissolution, erosion, precipitation, and mechanical compaction. These processes induce micro- and macro-variabilities within the rock structures, potentially creating preferential flow paths that enhance fluid conductivity or, conversely, obstructing paths and disrupting fluid flow. Utilizing petrophysical models accelerates the qualification of subsurface sites and ensures standardization across qualification studies, promoting a systematic approach. In this context, high-quality segmentation masks of CT-scanned volumes of rocks are the backbone for such models \cite{dong2009, al2018, jouini2022, ibrahim2021, mahmoud2023}. As the need to understand subsurface integrity for projects like \ce{CO2} storage grows, the importance of advanced vision algorithms to improve segmentation of CT-scans in a reliable and reproducible way becomes increasingly evident.

However, translating µCT data into actionable pore-scale modeling results through accurate image segmentation remains a labor-intensive, time-consuming, and subjective task \cite{da2021}. To address these challenges, scientists often rely on machine learning or deep learning approaches for quick and precise segmentations. Yet, obtaining publicly available multi-class datasets that encompass a diverse range of rock types is particularly difficult. Moreover, deep learning approaches typically require a substantial number of training samples to perform and generalize effectively, and they may still depend on spurious correlations for their predictions \cite{sagawa2020investigation, ye2024spurious}. Consequently, this scarcity of datasets prevents the research community from scaling and leveraging deep-learning-based geological segmentation models \cite{chauhan2016, balcewicz2021, da2021}. Therefore, the limited availability of data is a significant bottleneck for developing robust and versatile segmentation methods for geosciences.

Despite these challenges, foundation models demonstrate the capability to perform effectively with minimal data, highlighting their potential for broader adoption in geoscience applications. To the best of our knowledge, very few prior works explore and tune foundation models to fit the specificity of CT-scanned rock samples. Meanwhile, foundation models are seeing rapid and widespread adoption in the medical field \cite{anand2023, chen2023, perez2024, dippel2024, vorontsov2023, zimmermann2024}. In this work, we demonstrate that similar progress can be achieved to make accurate predictions by assessing and fine-tuning DINOv2 specifically for the analysis of CT-scanned rock samples. We perform an empirical evaluation and compare our method against a range of baseline models and traditional segmentation techniques \cite{li2021} such as Otsu's method \cite{otsu1975, liao2001}, K-means \cite{jain2010}, fuzzy C-means (FCM) \cite{dunn1973, belhassen2010}, Random Forest (RF) \cite{mansour2001, liaw2002}, and two Convolutional Neural Networks (CNNs), a customized UNet \cite{ronneberger2015, liang2022} and ResNet152 \cite{he2016}. We select a challenging dataset characterized by high noise levels and limited data availability to explore the options available to researchers facing constraints in training capabilities, whether due to data scarcity or noise challenges.

This study highlights the relevance of high-performing foundation models tuned to geoscientific data for segmentation and classification, especially when faced with limited and varied training data. Our research shows the synergy between DINOv2 and geosciences, demonstrating that DINOv2 significantly outperforms other widely used and popular segmentation methods. Notably, DINOv2 exhibits superior generalizing capability with noisy and scarce data, making it a powerful tool for advancing geological segmentation tasks. The goal of this research is to raise awareness among geoscientists regarding this powerful but underutilized vision model, as well as to provide empirical guidelines on how to best leverage it.

\section{Background}
\label{sec:background}
The methods described in this part are popular segmentation tools that provide a fair trade-off between processing speed and segmentation quality. Although they represent only a small subset of available algorithms, they serve as comparative benchmarks to highlight the capabilities of DINOv2. Segmentation techniques at the pore scale can be broadly categorized into several families, each with its own methodologies, use cases, advantages, and limitations.

\subsection{Threshold-based and unsupervised segmentation techniques}
Unsupervised and threshold-based segmentation techniques are computationally efficient and simple methods widely employed for image segmentation tasks. However, due to their limited capabilities, they often display limited performance when dealing with complex and noisy data.

Among threshold-based techniques, the Otsu method \cite{otsu1975} has been adapted for multiclass problems \cite{liao2001} and is particularly effective in preserving details and edges in CT images \cite{rulaningtyas2021}. The method determines the optimal threshold by maximizing between-class variance, ensuring optimal separation between foreground and background \cite{zhang2017, chang2019}. the Otsu method is widely used in various image processing applications, such as Fiji \cite{schindelin2012}, but its performance in noisy environments remains limited \cite{chang2019}.

Unsupervised approaches like K-means and FCM do not require annotated data during training, learning directly from raw data. K-means partitions data into $k$ clusters by assigning each pixel to the cluster with the closest mean intensity value \cite{macqueen1967, shapiro2001}. Despite its computational efficiency and ease of implementation \cite{jain2010}, K-means may struggle with data displaying a higher level of noise and/or complex gray level distributions that are difficult to distinguish by histogram analysis \cite{lai2009, jouini2015}.

FCM allows data points to probabilistically belong to multiple clusters \cite{dunn1973, belhassen2010}. This method is more robust than K-means in handling noisy data and overlapping clusters, providing nuanced segmentation by considering the degree of membership to multiple clusters. This characteristic is especially beneficial in scan-type data and applications where boundary regions are ambiguous.

\subsection{Supervised approaches} 
This study employs two supervised learning techniques. The first method is an adaptation of the RF algorithm, which is commonly used in the WEKA segmentation tool of Fiji \cite{schindelin2012, arganda2017}. On the deep learning front, after thorough consideration of various CNN-based approaches \cite{da2021, da2021-2, liang2022, liu2024}, we opted for the UNet model introduced by \cite{liang2022}. This decision is influenced by factors such as task compatibility, IT constraints, and code availability. The UNet model is selected for its readily accessible code and proven effectiveness in multi-class rock CT-scanning applications.

The RF algorithm \cite{mansour2001, liaw2002} is a versatile classification method that uses an ensemble of decision trees to improve predictive accuracy and generalization. Each tree is trained on a different bootstrap sample of the data, with randomness in both sample selection and feature choice, ensuring diversity among trees and reducing the risk of overfitting \cite{liaw2002, li2021}. Predictions in an RF are made by aggregating the votes from all trees, with the majority vote determining the final class. This approach leverages individual tree strengths while mitigating weaknesses, enhancing overall performance \cite{mansour2001, li2021}. Modern implementations, such as the WEKA trainable segmentation toolbox, enable parallel training of trees, significantly speeding up the process \cite{arganda2017} while leaving the user in control of the features used for training the method. RFs have been successfully applied in various geoscientific fields, including segmentation of SEM \cite{wu2019, lormand2018, li2021} and CT-scans of rocks \cite{purswani2020, garfi2020}.

The UNet architecture is a form of CNN used for image segmentation tasks. Introduced in \cite{ronneberger2015}, the architecture is characterized by its U-shaped structure, which consists of a contracting path (encoder) and an expansive path (decoder). The encoder captures the context of the input image through multiple layers of convolutional operations and downsampling \cite{lecun1989}, extracting important features such as fractures, pores, and mineral boundaries. The UNet architecture has demonstrated remarkable effectiveness in various segmentation tasks, ranging from fracture segmentation \cite{pham2023} to mineral and petrological segmentation \cite{li2021, da2021, song2020, niu2020, liang2022, chen2020}, and even lithology identification \cite{xu2021}. While it has shown proficiency in these areas, performance can vary depending on the availability and quality of annotated training samples. However, UNet models typically require training from scratch with extensive data, which is often gathered, utilized, and safeguarded within the industry. The limited public datasets that do exist may face compatibility issues across repositories due to class imbalance and discrepancies in represented phases, as they often belong to different projects. Overall, both UNet and CNNs have been instrumental in advancing the field of image segmentation, particularly in the context of digital rock analysis. Their ability to provide precise localization and accurate segmentation results makes them valuable tools for a wide range of applications. Another form of CNN is the ResNet \cite{he2016} which introduces skip connections or "residuals" that help mitigate the vanishing gradient problem, enabling the training of very deep networks. This structure allows the ResNet to learn more complex features without degradation, making it highly effective in a range of computer vision tasks. Due to its very high modeling capacity, it was one of the first model successfully pre-train on large corpus of data and fine-tuned on task specific datasets.

\subsection{The DINOv2 foundation model}
Foundation models are large-scale, pre-trained models designed to serve as versatile bases for a wide range of downstream tasks. These models are exceptionally useful because they can be efficiently fine-tuned. Fine-tuning involves adjusting the pre-trained model on a smaller, task-specific dataset, enhancing its performance for that particular task. This approach often outperforms models trained from scratch, especially in scenarios with limited data. DINOv2 \cite{oquab2024} represents a state-of-the-art (SOTA) self-supervised learning (SSL) method specifically designed for pre-training vision foundation models. It excels in creating robust image encoders by employing vision transformers (ViTs). These transformers leverage the attention mechanism \cite{vaswani2017attention}, enabling the model to process images in a way that captures complex patterns and features. By pre-training this architecture on a vast and diverse dataset of images, DINOv2 generates visual features that are highly effective for various applications, such as image classification, object detection, and segmentation. Notably, DINOv2 models were primarily trained on "natural" images, depicting everyday scenes and objects. However, these models can be fine-tuned for more specific domains, allowing adaptation to specialized applications like medical imaging or satellite image analysis \cite{bou2024exploring, baharoon2023}. Overall, DINOv2 foundation models are pre-trained image encoders capable of extracting meaningful and robust representations from various types of images, making them suitable for a wide range of vision tasks.

\section{Related works}
The performance of deep learning models depends significantly on the size of the dataset and the model. Modern foundation models in natural image domains use millions of images to train models with hundreds of millions (or even billions) of parameters \cite{caron2021, radford2021, kirillov2023, oquab2024}. However, in specialized fields such as medicine and geoscience, datasets of this magnitude are often difficult to collect due to the infrequency of image acquisition and challenges related to data sharing. Despite these limitations, foundation models have gained significant popularity across domain-specific applications due to their versatility and flexibility. They are notably employed in complex tasks such as medical image analysis, computational pathology, and semantic segmentation, demonstrating their adaptability and strong performance in a variety of specialized areas. 

Recent advancements demonstrate that models trained on natural images can perform well on out-of-distribution (OOD) medical images. Foundation models like DINOv2 generate data representations that generalize effectively across various tasks, making them suitable for pathology, where data is often unlabeled and diverse applications are common. These tasks include general pathology, cancer detection, tumor profiling, biomarker evaluation, rare disease identification, and various tissue detection. They are applied across multiple imaging modalities (CT, MRI, Ultrasound) and anatomical regions, demonstrating effectiveness in tasks such as localization, segmentation, depth estimation, and classification \cite{vorontsov2023, anand2023, dippel2024, zimmermann2024, cui2024}. DINOv2 captures rich semantic features from images, enabling high accuracy and robustness in medical imaging tasks, where detailed and precise analysis is crucial \cite{anand2023, baharoon2023}. Several foundation models have been evaluated in studies \cite{tayebi2024, bensaid2024, anand2023, huix2024, huang2024}, and DINOv2 has demonstrated superior adaptability to both in-distribution and OOD data. It generally surpasses methods such as Stable Diffusion, SAM, DINOv1, and the SOTA UniverSeg \cite{butoi2023} in a variety of segmentation tasks.

However, despite its success in the medical field, DINOv2 remains largely overlooked in the geosciences domain. Yet, this represents a promising opportunity to leverage its capabilities for segmenting CT-scanned rock images, potentially enhancing precision in rock segmentation which has been a pivotal task for many years. However, standard machine learning and deep learning approaches have been tested, starting with the well-known WEKA tool from the Fiji software \cite{schindelin2012, arganda2017}. This tool, which is based on an RF algorithm, is widely used by the geoscientific community. Many authors have presented various UNet and CNN architectures, either trained from scratch or utilizing pre-trained backbones to address this task, yielding satisfactory results \cite{niu2020, song2020, chen2020, li2021, da2021, xu2021, liang2022, malik2022, reinhardt2022, manzoor2023, pham2023, liu2024}. Newly introduced foundation models have attracted little attention and very few studies have attempted to adapt them for geoscientific segmentation tasks \cite{shankar2023, shan2024, koeshidayatullah2023, giannakis2024, julka2023, cristobal2024, hu2024}. Notably, only one study has been identified using SAM for CT-scanner segmentation \cite{ma2023_2}, and no studies have been found utilizing DINOv2 for this purpose.

\section{Evaluation Settings}
In this section, we briefly describe the experiments conducted to assess the potential benefits of DINOv2 in a geoscientific context. We present the common settings used in our experiments, including the data used, the pre-processing pipeline and the values of some of the most important hyper-parameters. Readers can refer to our publicly available code for more details and information. All experiments were run on NVIDIA RTX 3060 and Tesla T4 GPUs.

\subsection{Motivations}
The main purpose of our experiments is to analyse and highlight the strengths and limitations of DINOv2 within the specific context of geological image analysis. In Section \ref{sec:dino_raw}, we conduct a comprehensive probing and interpretability analysis of DINOv2's \textit{raw} features (i.e., without any fine-tuning). The model's effective and powerful data representation abilities allow it to capture the underlying structure of the data, promoting better generalization and robustness. By examining how DINOv2 encodes rock data, we gain a deeper understanding of the model's innate suitability for the geoscience domain. Additionally, high-quality features often correlate with efficient fine-tuning. Consequently, in Section \ref{sec:DINOv2-segmentation}, we investigate the ability of a segmentation model leveraging DINOv2 to adapt to our application with minimal data. Furthermore, to better contextualize our performance metric, we evaluate our model and provide a fair comparison with other SOTA approaches benchmarked in the same conditions. 

\subsection{Datasets}\label{sec:dataset}
For this study, we utilize two distinct sets of CT-scanned data for separate analytical tasks. We refer to the classification dataset as the \say{sandstones} set and the segmentation dataset as the \say{carbonates} set. Each raw file is downloaded from public repositories and converted to TIFF format, and then to the NPY data format. We automatically adjust the brightness and contrast of the TIFF files using Fiji to improve the images, particularly given their rather dark appearance. This adjustment improves the sharpness of the features in the images, which in turn boosts our models' predictions. Each dataset includes scanner images and their corresponding ground-truth (GT).

\subsubsection{Classification task}\label{sec:sandstones}
The classification task focuses on a \say{sandstones} dataset comprising 10 CT-scanned samples, each representing a different rock type with unique petro-morphological characteristics, as gathered by \cite{neumann2021}. Each scanner captures distinct petro-morphological features, contributing to variations in contrast, primarily influenced by particle sizes and their arrangement, resulting in a spectrum of gray shades. While typically more interpretable, some scans may exhibit partial noise or artifacts. These characteristics are leveraged for classification purposes in this study.

\subsubsection{Segmentation task}\label{sec:carbonates}
For the segmentation task, the \say{carbonates} dataset consists of three CT-scanned calcite core samples obtained by \cite{alhammadi2017}. These cores, extracted from drilling surveys in the same Middle Eastern reservoir rock. The dataset includes three classes: crude oil, brine, and rock matrix. Despite the use of a high-resolution beam, the CT-scans show high granularity with visible grains but low overall contrast, making it difficult to differentiate some dark areas. This low contrast and resolution limit the observation of fine details, though the rock’s surface texture remains somewhat discernible. These factors affect the clarity and precision of the scans, complicating detailed analysis of the rock’s microstructure. Each of the three calcite core samples is respectively denoted by \textit{S1}, \textit{S2} and \textit{S3}.

\subsection{Data processing and augmentation}\label{sec:data_proc_aug}
In deep learning-based computer vision, data augmentation is commonly used to improve model robustness and generalization. This is typically achieved through geometric and color transformations, such as rotations or color jittering \cite{oquab2024, caron2021}, applied directly to the input data. In our deep learning models, unless otherwise stated, the transformation pipeline consists of random or center cropping, upsampling, random vertical or horizontal flips, and adjustments to contrast, brightness, and gamma. We experiment with multiple crop and upsample sizes and find that a crop size of 224 followed by bilinear upsampling to 560 produces the best results. This resolution is selected after extensive trial and error, with larger resolutions improving performance until we reach our computational limits. These experiments suggest that fine-grained segmentation requires upscaling relatively low-resolution images to extract high-quality features and highlight small but important rock details. For the other approaches (machine learning and threshold-based), we denoise the data using a non-local means filter \cite{berg2018} to improve robustness, without applying any data augmentation. We also employ a hand-engineered basic feature extractor (BFE) to enhance some of our baseline machine learning algorithms (e.g., RF). This BFE operates on denoised data, using a Gaussian kernel to compute 15-dimensional feature vectors, which include pixel intensity, gradient intensity, texture variations, and local structure to capture features at different levels of detail. This approach is adapted from \cite{wu2019} to suit our specific needs.

\subsection{DINOv2 backbones and hyperparameters}\label{sec:DINOv2_hyp}
\begin{figure}[htb]
\begin{center}
\includegraphics[scale=0.6]{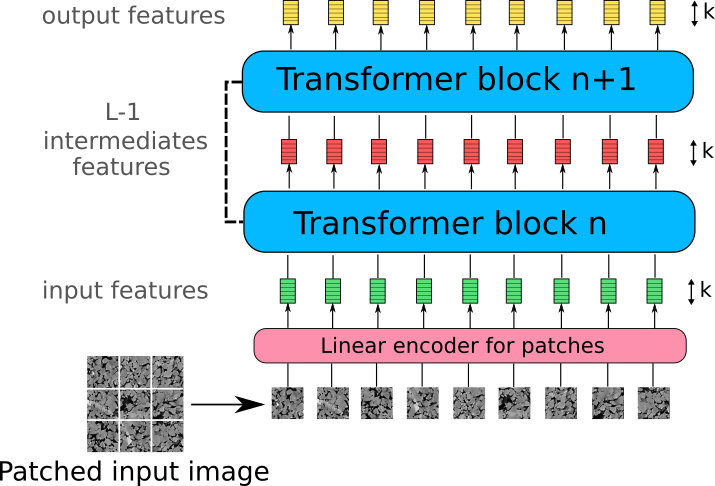}
\caption{Simplified illustration of DINOv2's ViT. Depending on the model size, the number of layers $L$ and the feature size $k$ may vary. We omit positional encoding for simplicity.}
\label{fig:dino_backbone}
\end{center}
\end{figure}

Depending on the experiment, this study leverages three DINOv2 model sizes: DINOv2-small (22M parameters), DINOv2-base (86M parameters) and DINOv2-large (300M parameters), respectively displaying feature vector sizes of 384, 768 and 1,024. Before being processed by the backbone, the input images are divided into fixed-size, non-overlapping patches. Following this, each of these patches is transformed into a corresponding feature vector using DINOv2's ViT (Figure \ref{fig:dino_backbone}). All DINOv2 models are trained using a patch size of 14, which, given our selected image size (560x560), results in feature patches of size 40x40 ($560/14$). 

To address this, we employ a parameter-efficient fine-tuning approach called Low-Rank Adaptation (LoRA), which adapts large models to specific tasks by significantly reducing the number of trainable parameters \cite{hu2021}. LoRA achieves this by freezing the weights of the pre-trained model to avoid retraining the entire model. Instead, it injects trainable low-rank matrices into some layers to approximate the changes required for new tasks. Additionally, LoRA introduces no extra inference latency, as the low-rank matrices can be merged with the pre-trained weights during inference. The rank of the LoRA matrices and the adapted layers are hyperparameters that must be tuned. In our case, we set the rank to $r=32$ and target all the linear layers, including the attention matrices. For instance, with DINOv2-base, we reduce the number of trainable parameters from 86M to only 5M, resulting in significant computational savings.

To further enhance memory efficiency, we implement 4-bit quantization. Quantization reduces the numerical precision of model weights, thereby significantly decreasing the memory footprint and computational requirements while maintaining performance. By combining quantization and LoRA fine-tuning (also known as QLoRA \cite{dettmers2024}), we achieve memory-efficient fine-tuning and inference of large models without sacrificing accuracy.

\section{Exploring DINOv2's Pattern Recognition in Rock CT-Scans}
\label{sec:dino_raw}
Our first experimental studies focus on the interpretability of DINOv2's features and their suitability to the geoscientific domain. We leverage existing probing and visualization methods to inspect whether DINOv2 can natively recognize and discriminate important petrophysical patterns. In other words, we aim to evaluate how effectively DINOv2 can interpret CT-scans of rocks by scrutinizing the amount of useful information encoded in its features. Our experiments are based on two datasets, presented in section \ref{sec:dataset} with significantly different characteristics. Unless otherwise stated, all experiments are run using a non-fine-tuned (frozen) DINOv2, i.e., none of its weights are updated. By doing so, we highlight the intrinsic capabilities of DINOv2 as a backbone for OOD data. 

\subsection{Assessing DINOv2 for Rock Type Classification}
\subsubsection{Experimental settings}
We visualize DINOv2-base features using data from the sandstones dataset (Section \ref{sec:sandstones}) to gain insights into DINOv2’s ability to process CT-scanned data with an \say{out-of-the-shelf} model. Our goal is to determine whether rocks with the same geological nature have similar latent representations in DINOv2’s feature space, which would suggest a zero-shot geological pattern recognition capability. Since DINOv2 generates high-dimensional embedding vectors, we use t-Distributed Stochastic Neighbor Embedding (t-SNE) for visualization. t-SNE is a powerful machine learning algorithm for dimensionality reduction and is particularly effective for visualizing high-dimensional data. It preserves the local structure of the dataset, making it valuable for illustrating clusters and complex relationships.

We further validate DINOv2’s capabilities by training a k-Nearest Neighbors (kNN) model directly on the latent features to classify rock samples. Intuitively, if t-SNE reveals clean and distinct clusters in DINOv2’s latent space, we expect this simple classifier to perform well. We randomly sample 3,000 images from the sandstones dataset and use an 80/20 train/test split strategy, allocating 80\% of the data for training and 20\% for validation. We experiment with different values for the number of neighbors, $k$, and also vary the image resolution.

\subsubsection{Results}
Our experiments with rock classification using a kNN model demonstrate remarkable stability and accuracy. The classification accuracy consistently reaches 100\% across a broad range of parameters, specifically regarding the number of neighbors, $k$, and image resolution. Only at very low image resolutions (below 128x128) or high values of $k$ (> 100) does the accuracy slightly decrease, dropping to a still impressive 96.5\%.

We plot the t-SNE visualization of DINOv2-base features in Figure \ref{fig:tsne-scatterplot}. As shown, scanners from the same dataset (rock sample) cluster together. More interestingly, rock samples with similar characteristics, such as Bandera brown and Bandera gray, also have close feature representations. A similar phenomenon is observed for Berea rocks. Qualitatively, we note that Bentheimer and Bandera gray rocks are petro-morphologically dissimilar, which may explain why their latent representations are far apart (see Figure \ref{fig:tsne-barycenters}).

Samples with similar petro-morphological characteristics exhibit proximity in the feature space, indicating DINOv2’s capability to generalize across subtle geological variations. While the exact components in such analyses are generally unclear, plotting a zoomed-in crop of a representative 2D slice of each dataset, centered around the barycenter of their respective point cloud, allows us to analyze the visual characteristics deemed meaningful in the distribution of geological scanners (see Figure \ref{fig:tsne-barycenters}). In this analysis, the first component appears to relate to pore size, while the second component might focus on porosity density. This observation, based on a similar set of data, highlights: 1) the strength of a non-fine-tuned DINOv2 in meaningfully organizing geological datasets by petro-morphological traits, and 2) the potential of using DINOv2 as a powerful tool for uncovering underlying geological features and patterns not immediately apparent through traditional methods. These findings emphasize the model’s capacity to interpret complex geological features effectively for accurate classification tasks, despite being pretrained on unrelated data. Such adaptability and interpretability enhance its utility in geological analysis and extend its potential for diverse applications in pattern recognition across various domains.

\subsection{Linear and kNN probing for segmentation}
We extend our previous findings and further study the performance of DINOv2 on a more challenging image segmentation task.

\label{sec:probing}
\subsubsection{Probing for feature quality assessment}
When applying deep learning, pixel-wise classification typically requires specialized and sophisticated neural network architectures \cite{da2021, da2021-2, liang2022, cheng2022, xie2021}. In contrast, we use a simple linear classifier on top of an out-of-the-shelf DINOv2 encoder. This evaluation of the linear layer decoder helps isolate DINOv2’s performance from the potential contributions of a more complex segmentation head. A well-structured representation space can effectively capture the underlying structure and relationships of the data, facilitating tasks such as classification and clustering. Therefore, in this experimental setup, linear classification accuracy serves as a proxy to assess the quality of the feature space provided by a frozen backbone. This straightforward procedure, known as \textit{probing}, is commonly used in SSL \cite{caron2021, oquab2024, grill2020, chen2020bootstrap} to quantify the usefulness of features extracted by a pre-trained model for downstream tasks. Similarly, we also apply a kNN model for segmentation and perform kNN probing.

\begin{figure}[htb]
    \centering
    \begin{subfigure}[t]{0.48\textwidth}
        \centering
        \includegraphics[width=\linewidth]{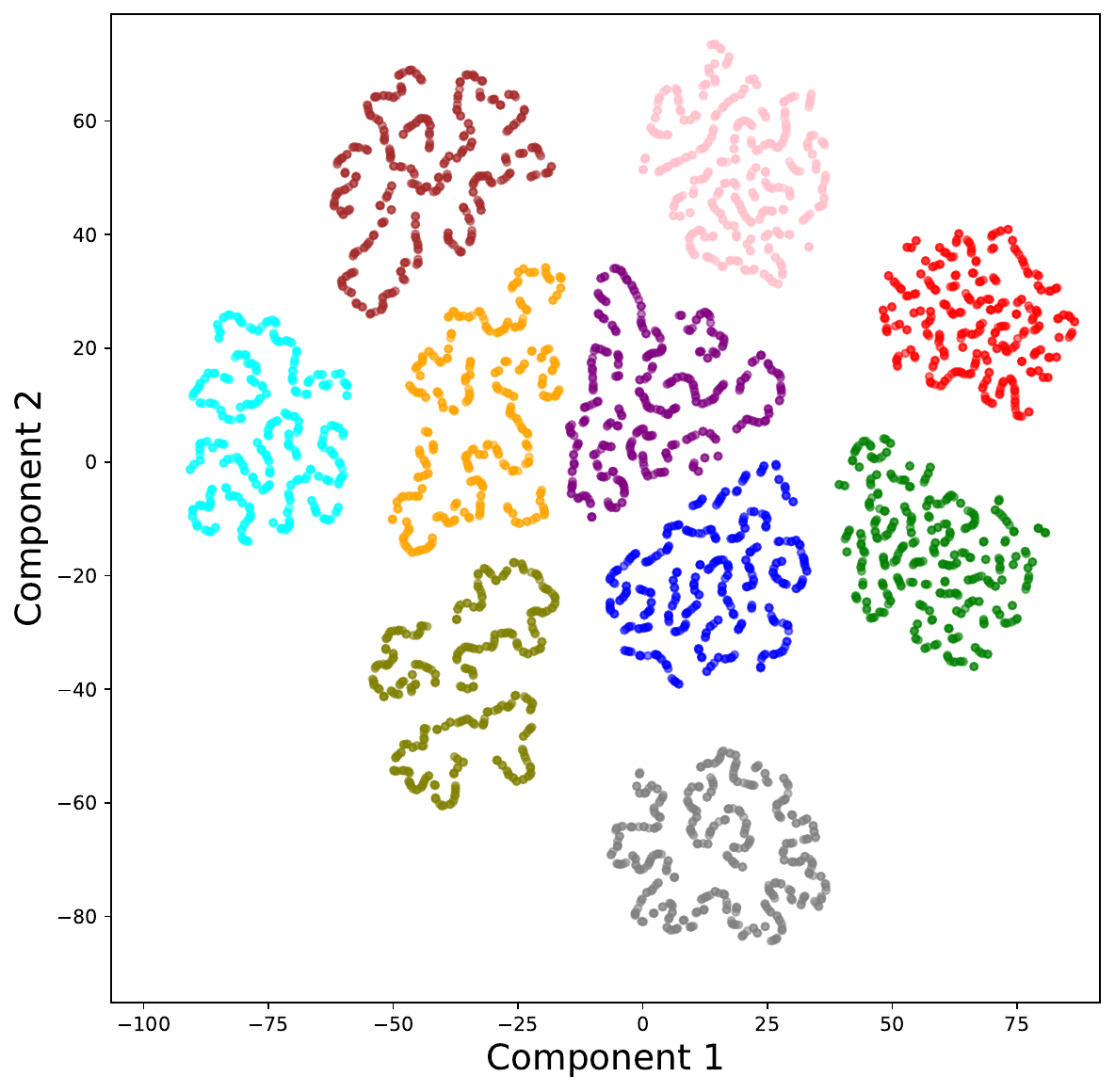}
        \caption{2D representation of DINOv2 clustering for the 10 sandstone samples.}
        \label{fig:tsne-scatterplot}
    \end{subfigure}
    \hfill
    \begin{subfigure}[t]{0.48\textwidth}
        \centering
        \includegraphics[width=\linewidth]{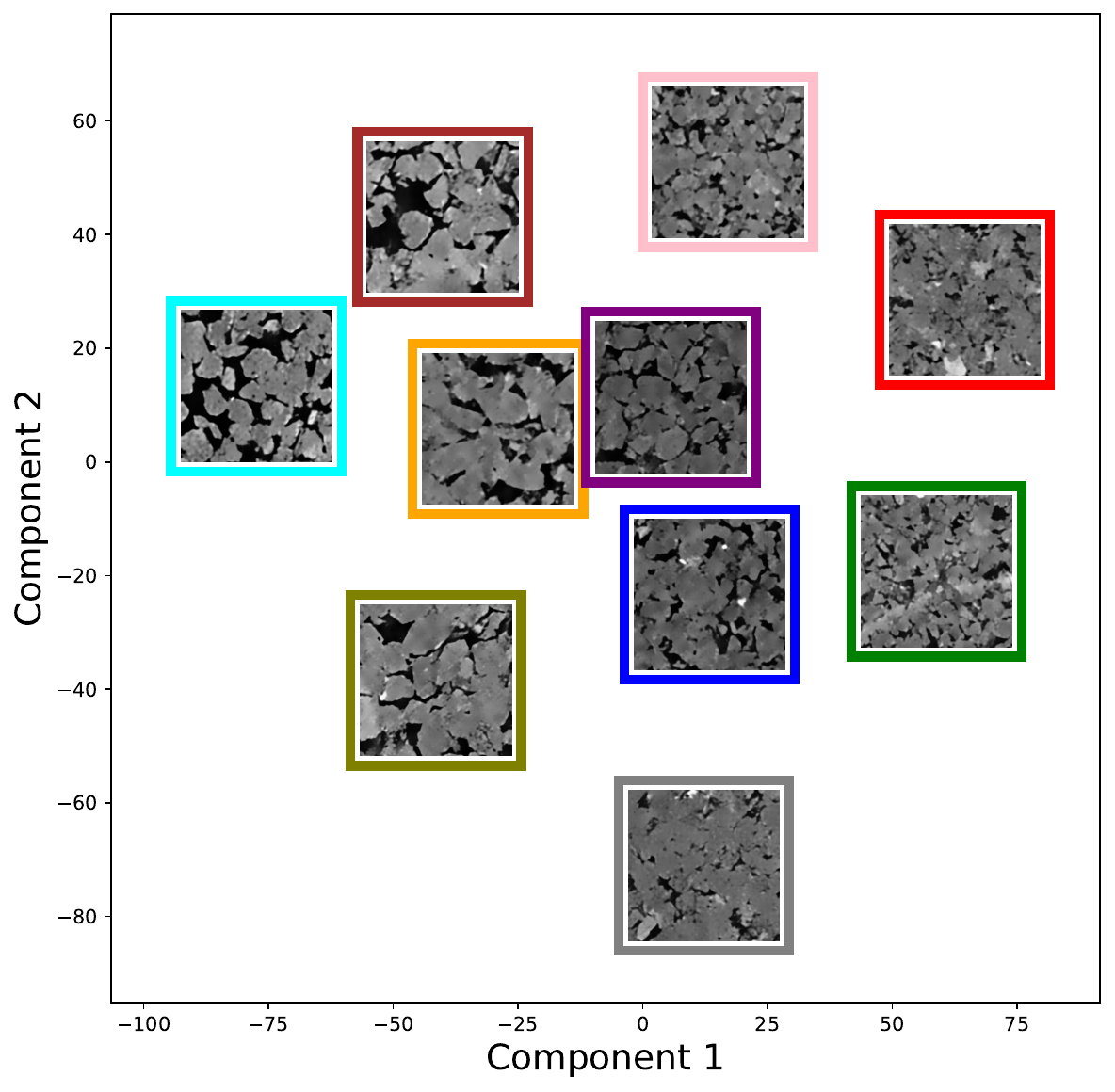}
        \caption{2D representation of DINOv2 clustering represented by a slice sample.}
        \label{fig:tsne-barycenters}
    \end{subfigure}
    \vspace{1em}  
    \begin{minipage}{0.95\textwidth}
        \centering
        \includegraphics[width=\linewidth]{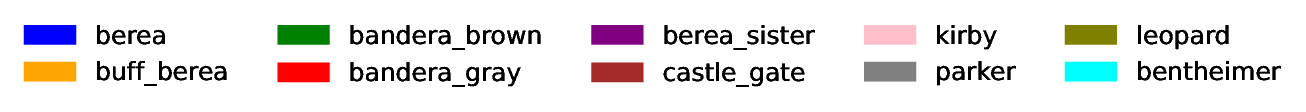}
        \label{fig:tsne-legend}
    \end{minipage}
    \caption{t-SNE visualization of the sandstones dataset embedded in DINOv2 feature space.}
    \label{fig:tsne-combined}
\end{figure}

\subsubsection{Experimental settings}
We experiment with DINOv2-small, DINOv2-base, and DINOv2-large to study how feature size correlates with downstream performance. Additionally, we report probing results on features extracted by BFE (see Section \ref{sec:data_proc_aug}) to highlight the benefits and overall superior quality of features learned by DINOv2. For linear probing, the segmentation head consists of a single convolutional layer with a 1x1 kernel and no padding. This layer is applied to the feature map to predict class logits. For DINOv2, the 40x40 logits map (see Section \ref{sec:DINOv2_hyp}) is bilinearly upsampled to the full resolution (560x560). As in the original DINOv2 paper \cite{oquab2024}, we use an enhanced version of the linear probing setup by concatenating the patch tokens from the last four layers of the backbone before the classifier.

For kNN probing, the feature map is upsampled from the low-resolution 40x40 feature map to a higher resolution that matches the GT resolution. This ensures a one-to-one mapping between the training labels and feature vectors. kNN models scale poorly with the number of pixels because their computational complexity increases significantly with larger datasets due to the need to calculate distances between each point in the dataset. To address this, we upscale DINOv2 features from 40x40 to 128x128 and downscale the GT masks from 224x224 to 128x128, making predictions computationally feasible on our hardware while maintaining sufficiently large images. For the same reasons, we use only 200 images for training and evaluate on 100 images. During prediction, pixel-wise classification is performed by extracting features from the test images and finding the k-nearest pixel’s latent representations in the training set, along with their corresponding labels. In practice, we find that $k=50$ neighbors yield the best results.

\subsubsection{Results}
We perform linear and kNN probing for multi-class image segmentation on the carbonates dataset. In this setup, we train our model on samples \textit{S1} and \textit{S2} and evaluate the model's performance and generalization ability using the IoU metric on the held-out and unseen \textit{S3} data subset. Results are reported in Table \ref{table_probing}. As shown, DINOv2 probing significantly outperforms the regular feature extractor. Interestingly, both the kNN and linear classifiers trained on top of a BFE perform poorly, indicating that this pixel classification task is not trivial. This underscores the essential contribution of DINOv2's powerful features to its success in the task. A sample of the predicted segmentation masks for both methods is provided in Appendix \ref{appendix:probing}.

\begin{figure*}[!ht]
     \centering
     \begin{subfigure}{0.48\textwidth}
         \centering
         \includegraphics[width=1.1\textwidth]{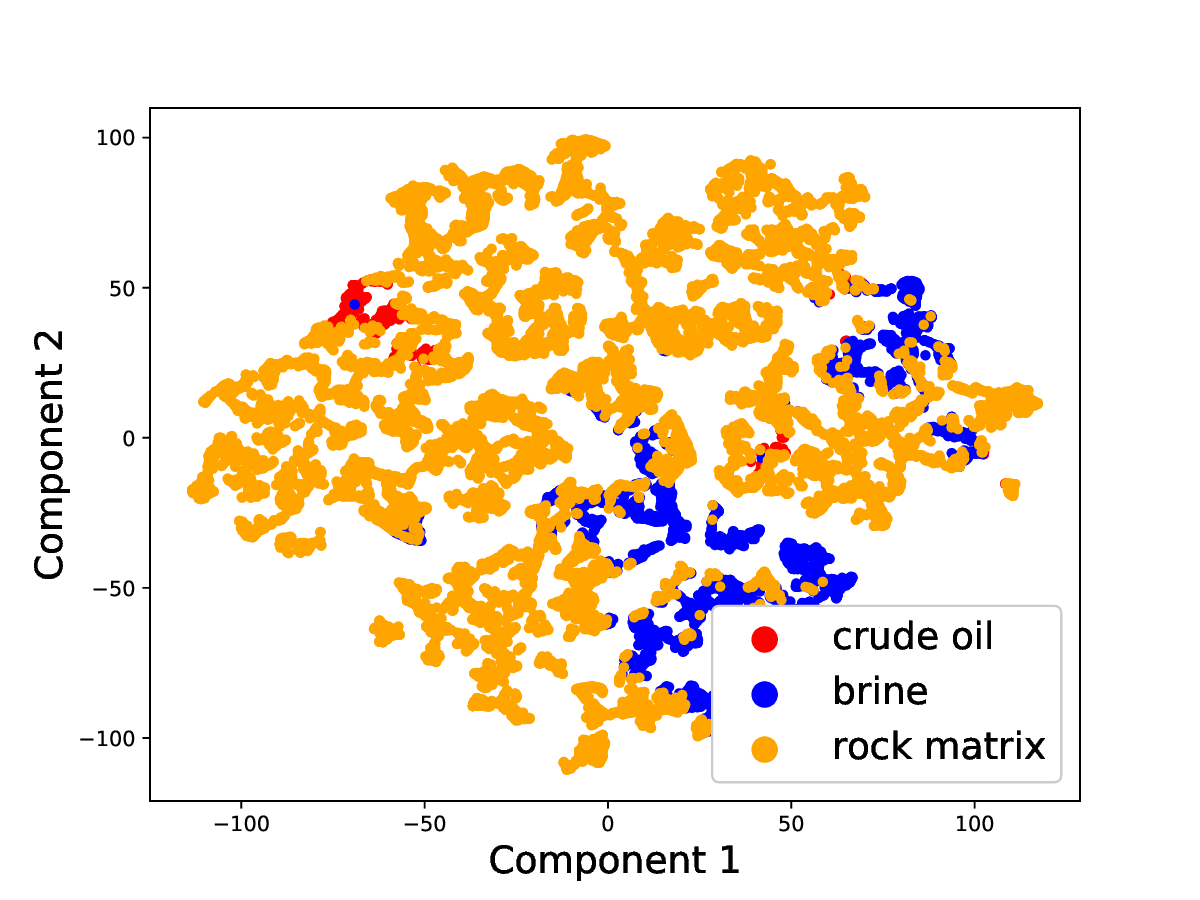}
         \caption{DINOv2}
         \label{fig:tsne-seg-dino}
     \end{subfigure}
     \hspace{0.3cm}
     \begin{subfigure}{0.48\textwidth}
         \centering
         \includegraphics[width=1.1\textwidth]{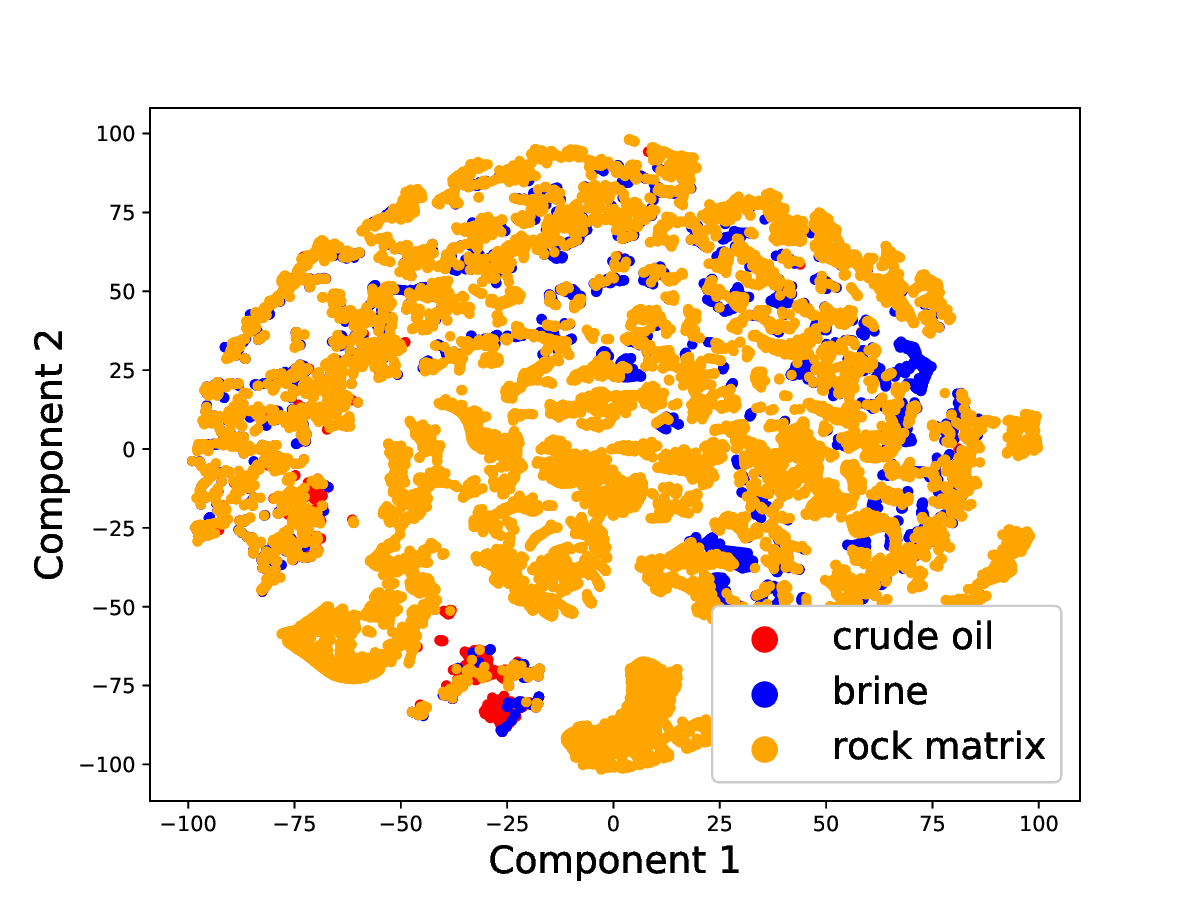}
         \caption{BFE}
         \label{fig:tsne-seg-bfe}
     \end{subfigure}
        \caption{t-SNE visualization of pixel-level feature vectors for a test image from sample \textit{S3}. This figure shows a 2-dimensional representation of DINOv2 and BFE clustering the pixel's latent representation of the same image.}
        \label{fig:tsne-seg}
\end{figure*}

\begin{figure*}[!ht]
     \centering
     \begin{subfigure}{0.48\textwidth}
         \centering
         \includegraphics[width=1.1\textwidth]{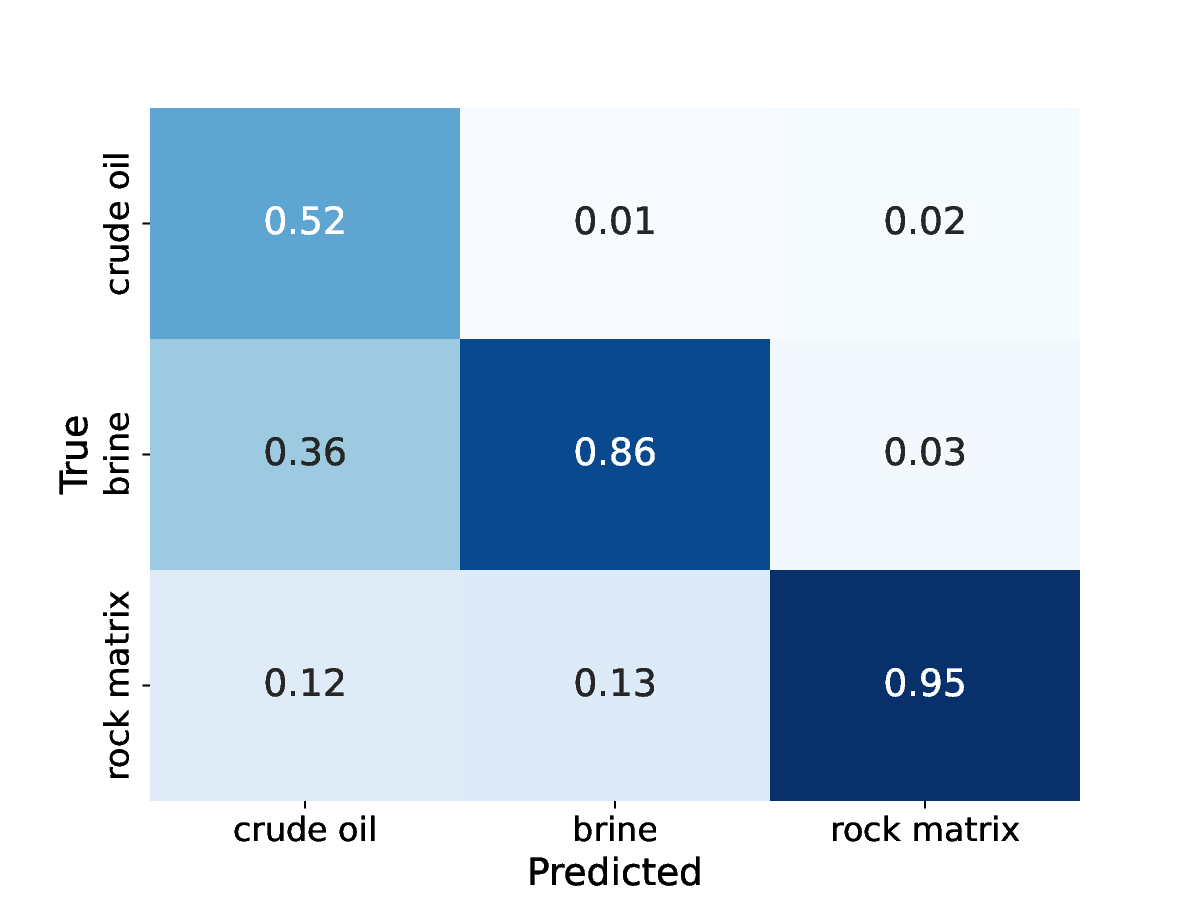}
         \caption{DINOv2}
         \label{fig:confusion_matrix_dino}
     \end{subfigure}
     \hspace{0.3cm}
     \begin{subfigure}{0.48\textwidth}
         \centering
         \includegraphics[width=1.1\textwidth]{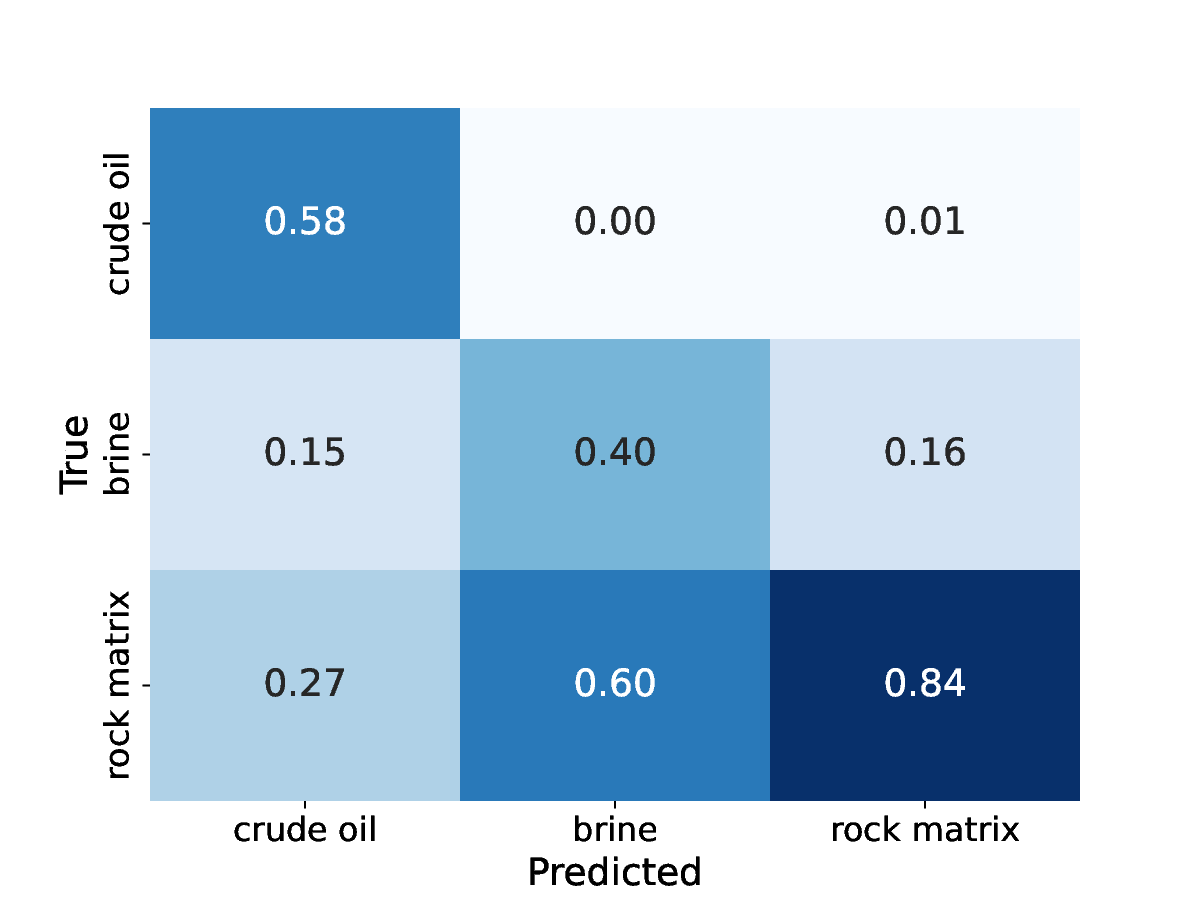}
         \caption{BFE}
         \label{fig:confusion_matrix_bfe}
     \end{subfigure}
        \caption{Confusion matrices for a segmentation prediction with kNN probing on a test image. Row indices of the confusion matrix correspond to the true class labels and column indices correspond to the predicted class labels.}
        \label{fig:confusion_matrix}
\end{figure*}

We use t-SNE visualization to understand why kNN performs better with DINOv2 compared to BFE. We plot the low-dimensional representation of each 128x128 feature vector for the same image using both approaches in Figure \ref{fig:tsne-seg}. It appears that the feature representation of the tested image under BFE is much more disorganized than DINOv2's. With BFE, the pixel feature representations corresponding to brine and rock matrix are mostly blended and overlap. This is qualitatively observed in Appendix \ref{appendix:probing}, which depicts a predicted mask using kNN and BFE. Quantitatively, Figure~\ref{fig:confusion_matrix_bfe} shows that a significant number of pixels from the brine class (79.5\%) are incorrectly classified as rock matrix pixels. In contrast, pixels representing crude oil are better clustered in the t-SNE plot, explaining their higher classification accuracy. DINOv2's latent space contrasts the three classes more effectively, leading to better probing performance. Nevertheless, there are still overlaps between some instances, which explains the confusion and misclassification of certain pixels (see Figure~\ref{fig:confusion_matrix_dino} for more details).

\begin{table}[htb]
\centering
    \begin{subtable}[h]{\textwidth}
    \centering
    \begin{tabular}{lccc}
        \toprule 
        \multirow{2}{*}{Models} & \multirow{2}{*}{kNN} & \multicolumn{2}{c}{Linear} \\
        \cmidrule(l){3-4} && 1 layer & 4 layers \\
        \midrule
        DINOv2-small & 0.584 & 0.584 & 0.701 \\
        DINOv2-base & 0.617 & 0.663 & 0.732 \\
        DINOv2-large & 0.621 & 0.673 & 0.730 \\
        \midrule 
        BFE & 0.451 & \multicolumn{2}{c}{0.387} \\
        \bottomrule
    \end{tabular}
\end{subtable}
\caption{Segmentation results with probing on the test dataset measured as the IoU.}
\label{table_probing}
\end{table}

From a performance perspective, the boosted linear probing setup, which processes the last four layers of DINOv2, significantly outperforms both its 1-layer counterpart and the kNN classifier, especially for DINOv2-small (+0.117 IoU). This suggests that useful information is present not only in the final layer but also in the intermediate layers. However, the performance gain from processing multiple layers diminishes as the feature size increases; for example, it results in only a +0.057 IoU improvement for DINOv2-large. This is likely because higher-dimensional features inherently embed more information, making additional layers somewhat redundant. Moreover, processing more layers increases computational demands. Consequently, in practice, the number of processed layers should be considered an additional hyper-parameter, balancing computation cost against performance. It is also worth noting that for large feature dimensions ($\geq 768$), linear probing achieves better IoU results than kNN, which performs poorly with high-dimensional data. 

Overall, we found that a simple linear classifier trained on top of the frozen DINOv2 features achieves impressive results in segmentation tasks compared to a hand-engineered BFE. Additionally, DINOv2 has demonstrated remarkable robustness to noise compared to BFE-based models, which required an extra data denoising step. When using multi-layer features, DINOv2’s performance even approaches that of a fully supervised UNet model trained from scratch on this specific task and dataset ($0.78$ IoU). Furthermore, the t-SNE visualization offers valuable insights into DINOv2’s features and explains its surprising zero-shot generalization to geological data. However, DINOv2’s performance could be significantly enhanced with parameter-efficient fine-tuning techniques and a more sophisticated segmentation head (see Section \ref{sec:DINOv2-segmentation}).

\subsection{PCA visualization of DINOv2 features}\label{sec:pca_features}
\subsubsection{Experimental settings}
Prior research on self-supervised vision foundation models has demonstrated the emergence of properties such as semantic segmentation \cite{caron2021, oquab2024}, even though these models were not explicitly trained for such tasks. However, these phenomena have been observed primarily in natural images that are close to the training distribution. The question remains whether these spontaneous abilities extend to geoscientific data. Building on the original DINOv2 publication \cite{oquab2024}, we compute Principal Component Analysis (PCA) on the features predicted by DINOv2-base for CT images to explore this question. PCA is a dimensionality reduction technique that transforms high-dimensional data into a lower-dimensional space by identifying directions of maximum variance. We visualize the first three principal components by assigning each to a different color channel, creating a representation where similar areas are displayed in matching colors, similar to segmentation. 

We consider two settings: a non-fine-tuned DINOv2-base and the same model after fine-tuning. The non-fine-tuned setting allows us to visualize the extent to which spontaneous segmentation emerges with rock data. In contrast, the fine-tuned model helps us understand how the features evolve after task-specific training. Fine-tuning is performed using QLoRA, with a linear segmentation head that encourages the model to rely as much as possible on the backbone. In Figure \ref{fig:pca}, we illustrate PCA on the same 2D slice of rock from the carbonates dataset for both models.

\subsubsection{Results}\label{sec:results_pca}
\begin{figure}[h]
    \centering
    \begin{subfigure}[b]{0.24\textwidth}
        \centering
        \includegraphics[width=\textwidth]{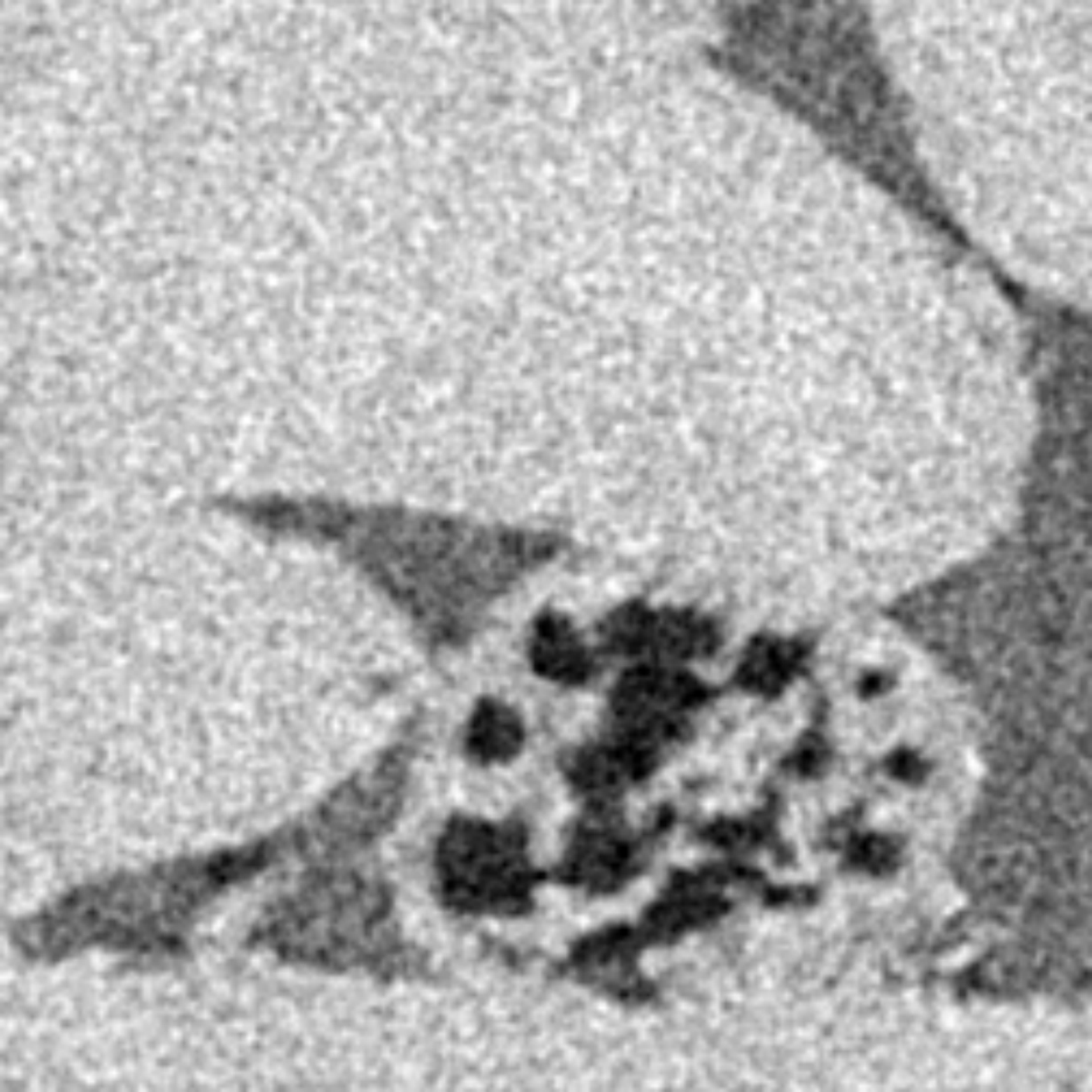}
        \caption{CT-scan}
        \label{fig:ct_scanner}
    \end{subfigure}
    \hfill
    \begin{subfigure}[b]{0.24\textwidth}
        \centering
       \includegraphics[width=\textwidth]{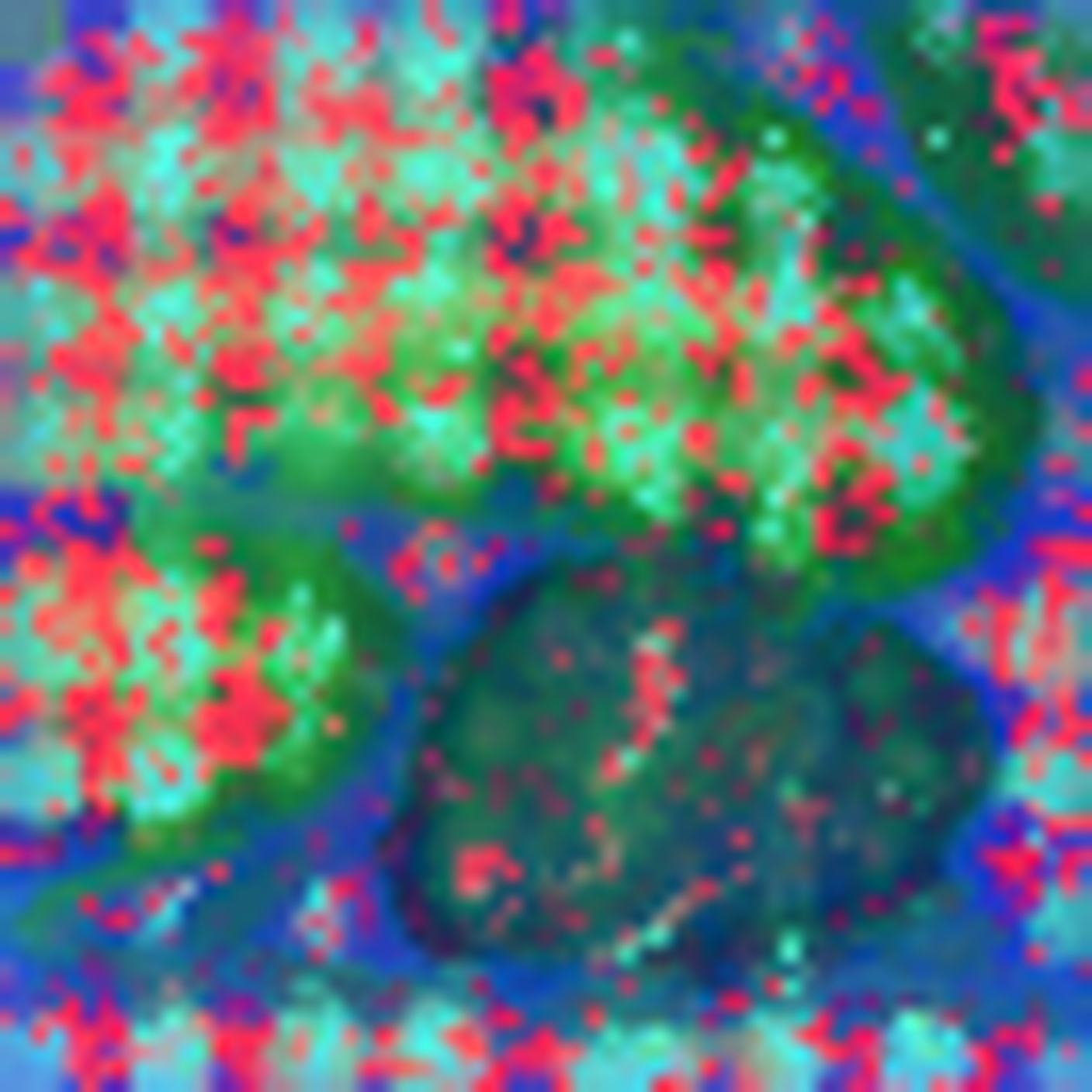}
        \caption{Before fine-tuning}
        \label{fig:frozen_backbone}
    \end{subfigure}
    \hfill
    \begin{subfigure}[b]{0.24\textwidth}
        \centering
        \includegraphics[width=\textwidth]{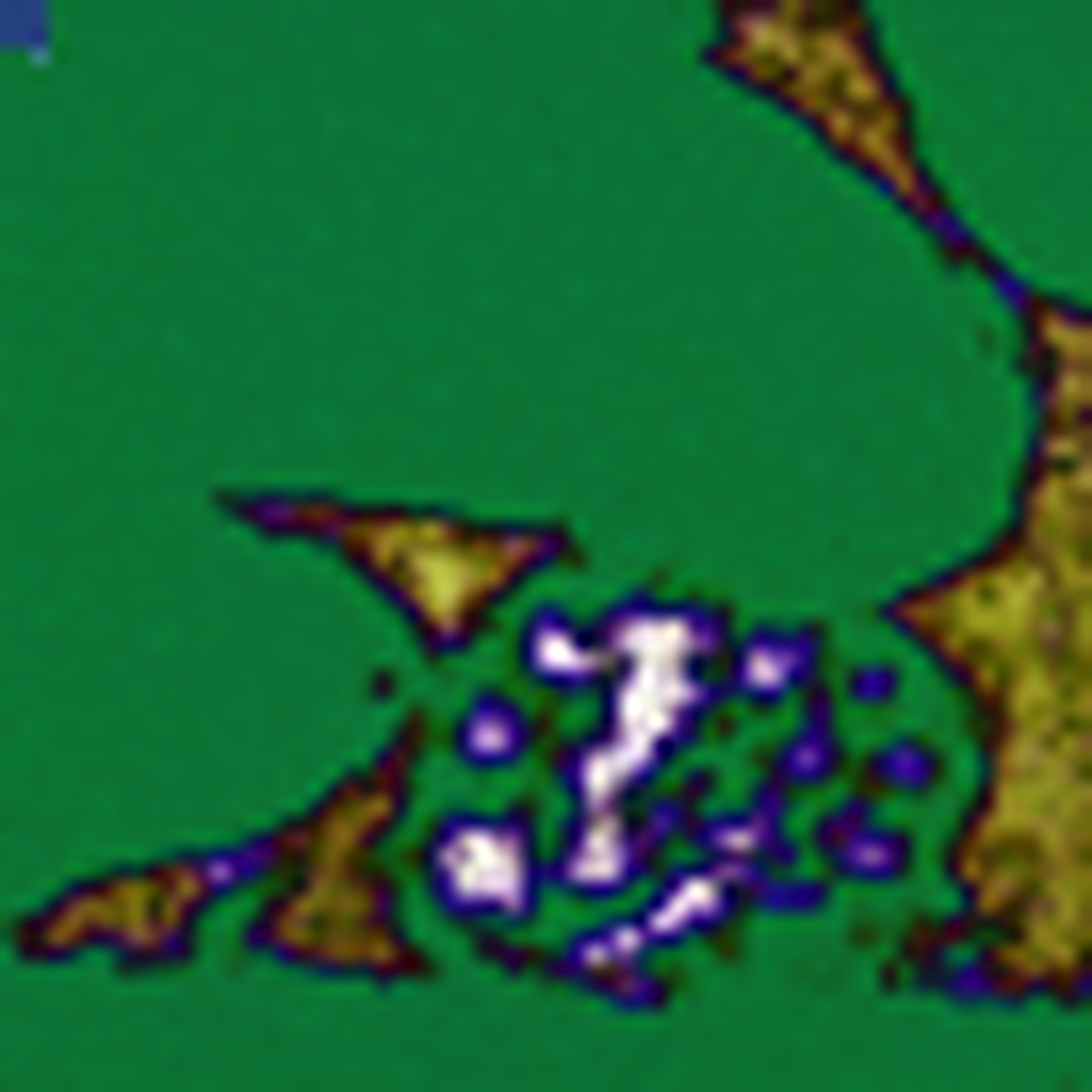}
        \caption{After fine-tuning}
        \label{fig:fine_tuned_pca}
    \end{subfigure}
    \hfill
    \begin{subfigure}[b]{0.24\textwidth}
        \centering
        \includegraphics[width=\textwidth]{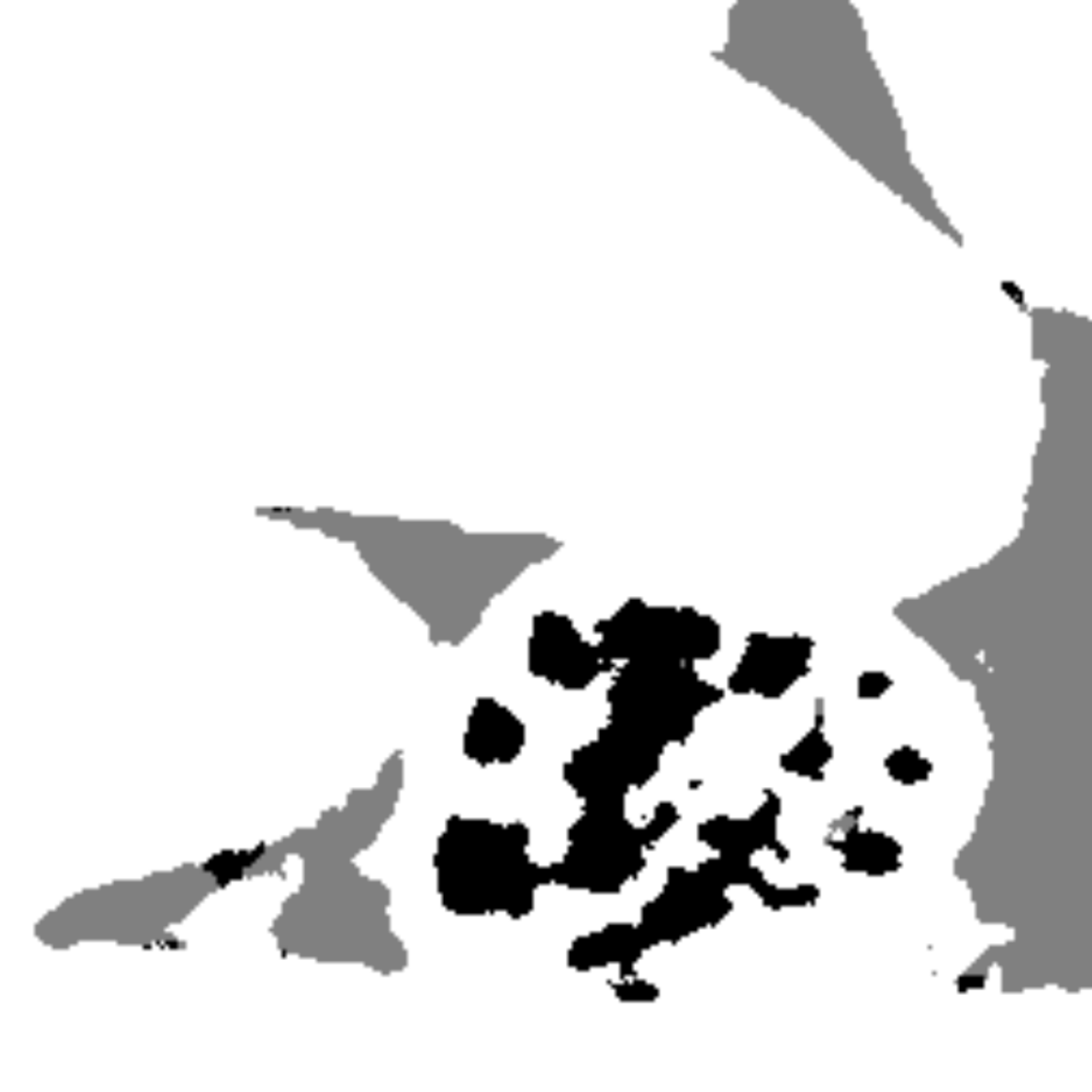}
        \caption{GT}
        \label{fig:ground_truth}
    \end{subfigure}
    \caption{Visualization of the 3 principal components of CT-scanned images respectively using the raw features of DINOv2 (b) and using a fine-tuned linear head (c). The CT-scanner (a) and the GT (d) are displayed for easy comparison.}
    \label{fig:pca}
\end{figure}

We observe a clear difference between Figure \ref{fig:frozen_backbone} and Figure \ref{fig:fine_tuned_pca}. The out-of-the-shelf DINOv2 outlines the most salient parts of the image but fails to segment the different phases (oil, brine, and rock matrix). It seems to focus on shapes and surfaces but largely disregards their semantic characteristics. This suggests that CT-scans of rock samples represent significant OOD data for the DINOv2 model. Non-fine-tuned PCA features do not produce spontaneous and meaningful segmentation masks, unlike in-distribution natural images. However, fine-tuned features, when displayed after PCA, accurately highlight various meaningful components in the original CT-scan and effectively target the three different classes as shown in Figure~\ref{fig:fine_tuned_pca}: crude oil (bright purple), brine (dark yellow), and rock matrix (green).

\section{Fine-tuning DINOv2 for high-quality segmentation}\label{sec:DINOv2-segmentation}
In this part, we focus our empirical analysis on the performance of DINOv2-base segmentation models after fine-tuning on the carbonates dataset. To better understand the contribution of each component to the overall performance, we first conduct an ablation study on our proposed model. Following this, we benchmark our best-performing variant against multiple widely used SOTA segmentation techniques for rock CT-scans to determine how it compares in terms of performance. In particular, we adopt an experimental protocol that assesses the ability of tested models to learn and generalize from potentially small amounts of noisy data. By doing so, we closely mimic the realistic and practical conditions researchers might face.

\subsection{Ablation study}
\label{sec:ablation}
\subsubsection{Experimental settings}
\begin{figure}[H]
\begin{center}
\includegraphics[scale=0.33]{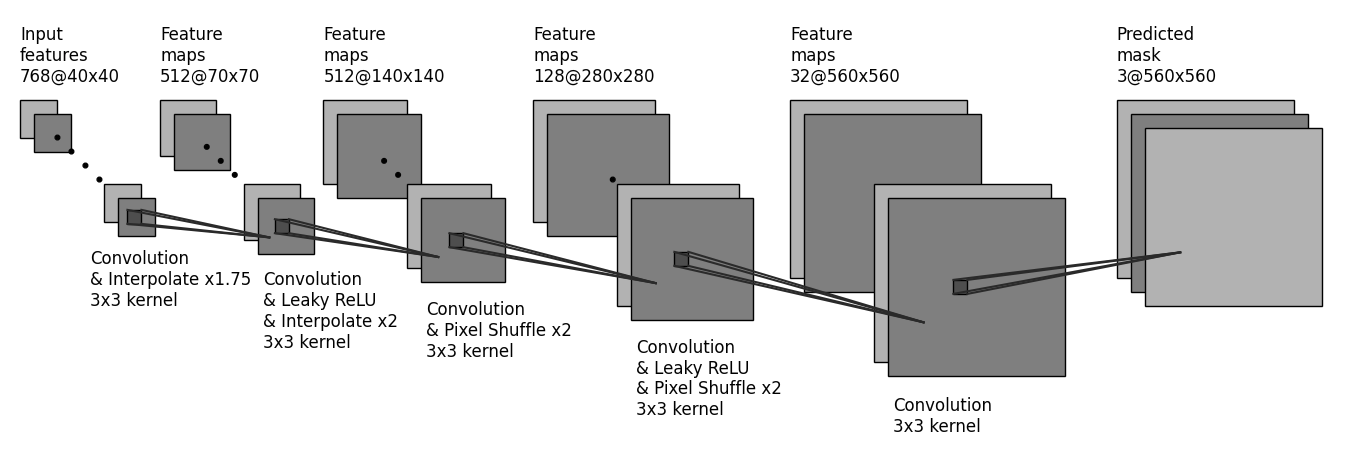}
\caption{Architecture for our convolution-based segmentation head. The spatial sizes displayed correspond to a DINOv2-base backbone.}
\label{fig:cnn_head}
\end{center}
\end{figure}

For both experiments, we train on the samples \textit{S1} and \textit{S2} datasets and validate performance using the IoU metric on the held-out sample \textit{S3}. All experiments are run using the DINOv2-base backbone, as it represents a good trade-off between computational cost and performance. In addition to our linear head (the same as in our linear probing setup in Section \ref{sec:probing}), we consider a segmentation head with 4 convolutional layers (which we refer to as ConvHead, or CH). This segmentation head starts with an initial convolutional layer that reduces the input features to 512 channels, followed by 4 decoder layers that progressively upscale and apply convolutions, with Leaky ReLU activation and dropout applied every two layers. The final segmentation map is produced by a convolutional layer that reduces the channels to the desired number of output classes. Each convolutional layer uses a 3x3 kernel with padding set to \textit{same}. Our CNN head is displayed in Figure~\ref{fig:cnn_head}. We arrived at this architecture through manual hyper-parameter search and insights from Section \ref{sec:pca_features}. As shown in Figure \ref{fig:fine_tuned_pca}, PCA can extract an accurate (although low-resolution) segmentation mask from a fine-tuned backbone. Since PCA is a linear projection, we hypothesize that a simple segmentation head should work well. Experimentally, we found that overly complex segmentation heads, such as UNet decoders or even four CNN layers with non-linear activation functions, lead to strong overfitting. This is further corroborated by prior work \cite{huix2024}, which found that linear heads work better than transformer-based segmentation heads. As a result, we aimed for the right balance between simplicity and modeling power. We consider 4 segmentation models based on DINOv2, depending on whether we fine-tune the backbone and which head is trained on top:

\begin{itemize} 
    \item \textbf{DINOv2-LinearHead}: The same model used for linear probing (see Section \ref{sec:probing}). We only fed the features from the last layer of the backbone to the linear head. This baseline setup provides insights into the intrinsic abilities of DINOv2 for rock segmentation. 
    \item \textbf{DINOv2-LinearHead-FT}: A DINOv2 model fine-tuned with a simple linear classifier simultaneously trained on top (using features from the last layer only). As such, the segmentation performance heavily relies on DINOv2's adapted features.
    \item \textbf{DINOv2-ConvHead}: An out-of-the-shelf DINOv2 backbone with a ConvHead fitted on top of the frozen features. This configuration is used to assess the impact of a more powerful segmentation head independently of a fine-tuned backbone. 
    \item \textbf{DINOv2-ConvHead-FT}: The full model where both the backbone and the \say{powerful} segmentation head are fine-tuned together. 
\end{itemize}

Fine-tuned models are quantized to 4 bits and further refined using LoRA. We also consider three additional baselines to better contextualize IoU performances: a light-weight UNet (UNet-small) with roughly the same number of trainable parameters as DINOv2-LinearHead-FT (respectively 7.8M and 5.3M), a large UNet (approximately 31M parameters), and a QLoRA fine-tuned ResNet152 encoder with a ConvHead. We conducted experiments on these models while maintaining consistent hyperparameters (see Section \ref{sec:data_proc_aug}). Each network is trained over 20 epochs with 1,000 images, evenly sampled between samples \textit{S1} and \textit{S2} from the carbonates set (see Section \ref{sec:carbonates}). For the large UNet model, we reduced the batch size from 15 to 8 due to GPU RAM limitations.

\subsubsection{Results}
\begin{figure}[h]
    \centering
    \includegraphics[scale=0.5]{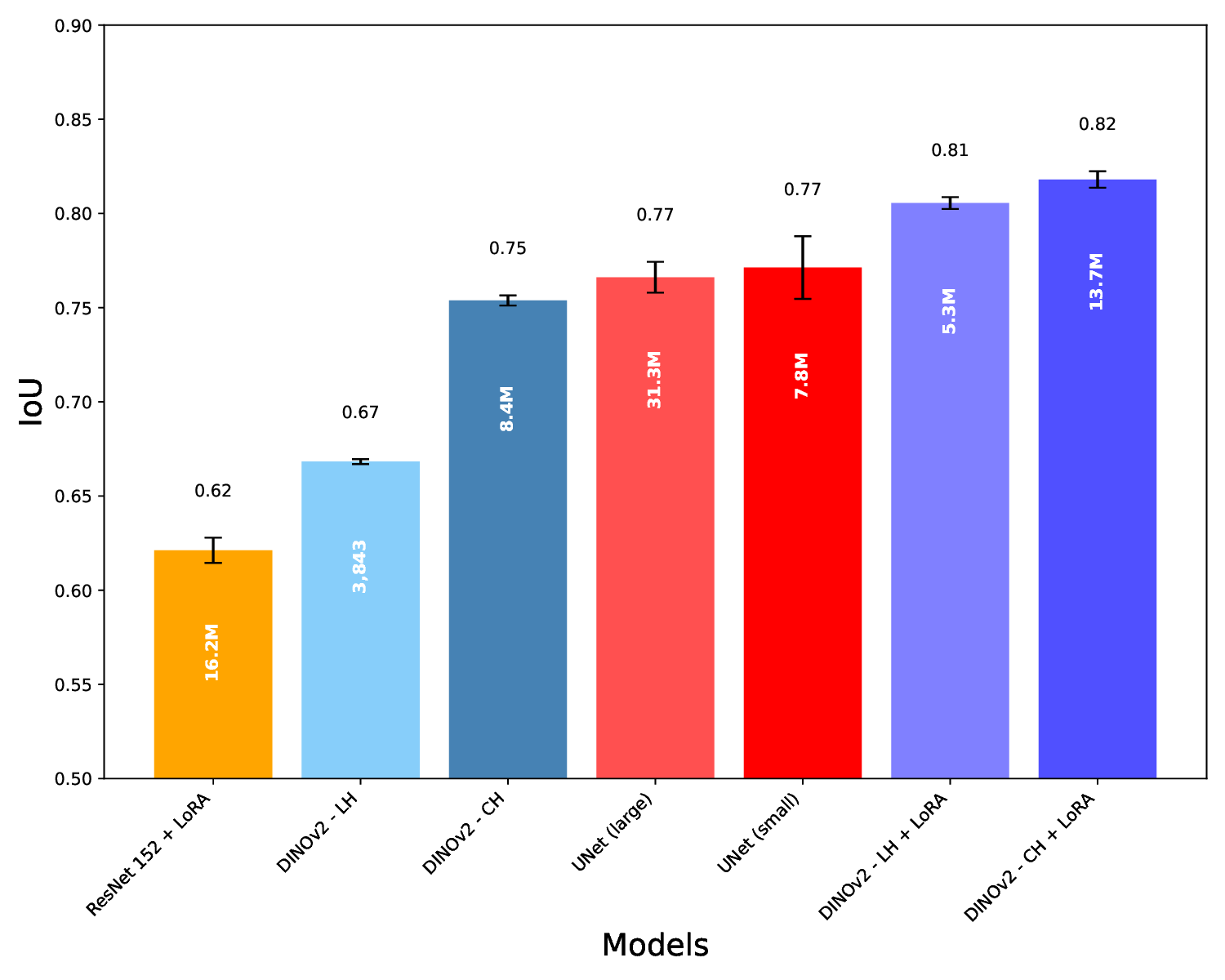}
    \caption{IoU values for various models and heads. Models compared are  ResNet152, UNet and DINOv2. Heads are the linear head (LH) and the slightly more complex ConvHead (CH). The values in each bar show the number of trainable parameters per model.}
    \label{fig:iou_models}
\end{figure}

Figure \ref{fig:iou_models} displays the mean IoU achieved after five runs for each model, using exactly 1,000 images of size 560x560 for training (from \textit{S1} and \textit{S2} samples). As shown, DINOv2-LinearHead still outperforms a LoRA fine-tuned ResNet152 despite having a significantly lower count of trainable parameters. Indeed, the ResNet152 is pre-trained on a standard supervised classification task, which likely produces overspecialized features that are harder to adapt to OOD data and unseen tasks. As expected, selecting a more sophisticated segmentation head on top of DINOv2's frozen features notably improves IoU performance (0.75 compared to 0.67) and is competitive with our UNet baselines. However, the biggest performance boost comes from fine-tuning the DINOv2 backbone with LoRA. In this setting, a simple linear head is still sufficient to achieve an IoU of 0.8, outperforming both UNets trained from scratch and underscoring the usefulness of DINOv2's features. Given that the high quality of the features produced by DINOv2's backbone is believed to be responsible for these results, only a small improvement is observed when transitioning from DINOv2-LinearHead-FT to DINOv2-ConvHead-FT, suggesting that most of the overall performance can be attributed to the fine-tuned backbone. We also note that increasing the parameter count of our UNet model from 7.8M to 31.3M does not improve performance and even leads to slight overfitting.

\subsection{Benchmark}
We evaluate the segmentation performance of each benchmarked model across different data regimes. For a review of the evaluated methods, please refer to Section \ref{sec:background}.

\subsubsection{Experimental settings}
Each supervised model is trained on the carbonates dataset described in Section \ref{sec:dataset}, using a varying number of images, ranging from 4 (2 images per sample) to 1,200 (600 images per sample). The training set consists of samples \textit{S1} and \textit{S2}, while testing is performed on sample \textit{S3}. For segmentation methods that do not require training (Otsu, K-means, and FCM), inference and evaluation are conducted solely on sample \textit{S3}. This approach aligns with our objective to achieve the best possible segmentation on \textit{S3}, as defined by the task requirements. Among the supervised models, we selected the two with the highest IoU scores from the UNets and DINOv2 variants: the DINOv2-ConvHead-FT model and the lightweight UNet, as discussed in Section \ref{sec:ablation}. Additionally, we included an RF model in the benchmark, chosen for its strong performance with the high-performance WEKA tool in Fiji \cite{schindelin2012}. The hyperparameters of the RF model were optimized using a grid search. Instead of using raw pixels as input, we enhanced the RF model by feeding it features extracted by the BFE from Section \ref{sec:probing}, thereby improving its performance and robustness.

\subsubsection{Results}\label{sec:results_benchmark}
The three unsupervised segmentation algorithms—Otsu, K-means, and FCM—produced similar results on sample \textit{S3}, with IoU values around 0.67. For comparison, we also segmented samples \textit{S1} and \textit{S2}, as shown in Table~\ref{tab:table_methods_seg}. Prior to segmentation, all images were denoised using a non-local means filter, as processing raw images led to significantly lower IoU values. Despite the intrinsic similarities between \textit{S1}, \textit{S2}, and \textit{S3}, traditional methods struggled with \textit{S2} and \textit{S3}, while achieving relatively better performance on \textit{S1}.

\begin{figure}[h!]
    \centering
    \includegraphics[width=0.8\textwidth]{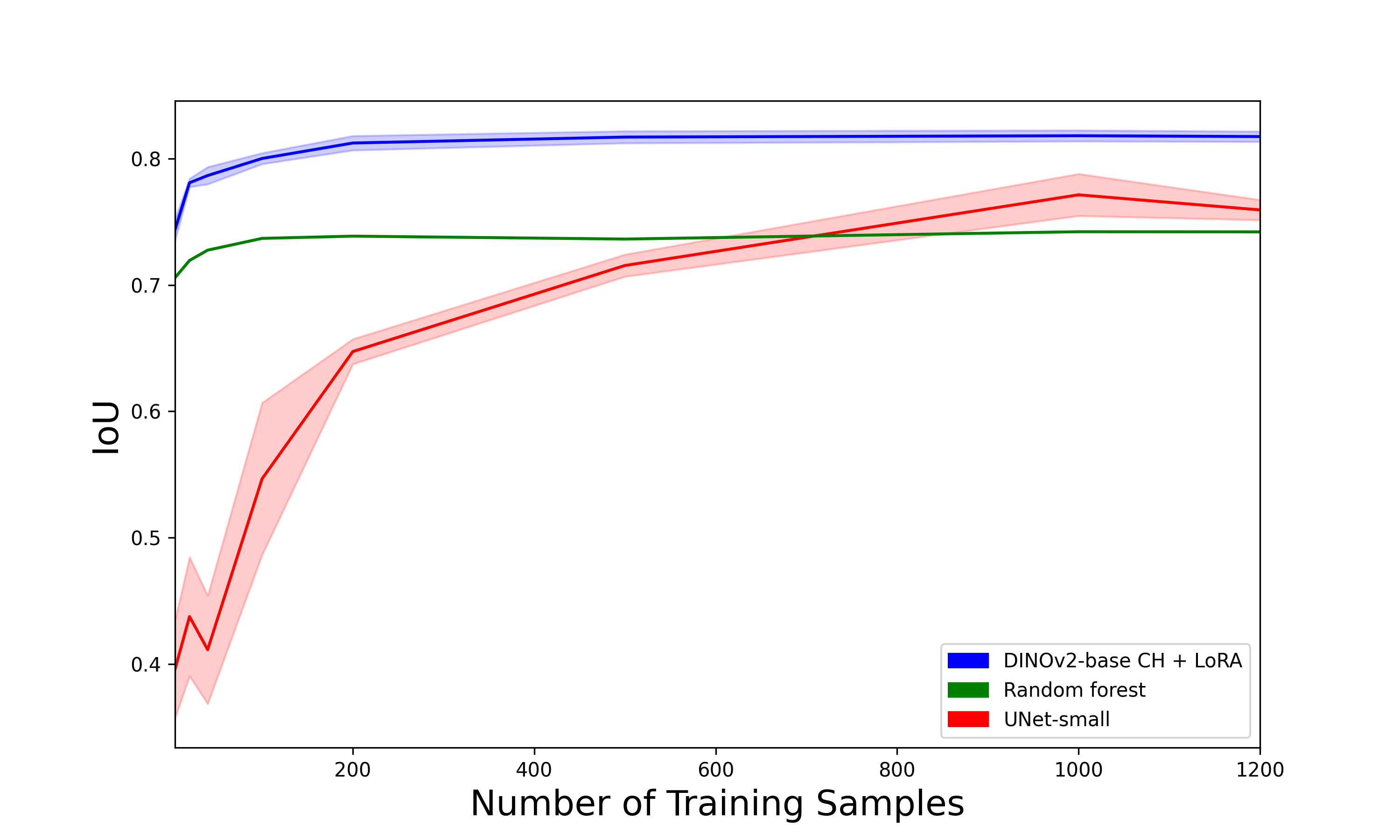}
    \caption{IoU as a function of the number of samples used during training of RF, UNet-small and DINOv2.}
    \label{fig:mean_iou_plot}
\end{figure}

\begin{table}[h]
\centering
\begin{tabular}{@{}>{\raggedright}p{2cm}ccc@{}}
\toprule
\multirow{2}{*}{Methods} & \multicolumn{3}{c}{Samples} \\ \cmidrule(l){2-4} 
                         & \textit{S1} & \textit{S2} & \textit{S3} \\ \cmidrule(r){1-4}
Otsu                     & 0.731   & 0.571   & 0.672   \\
K-means                  & 0.732   & 0.572   & 0.672   \\
FCM                      & 0.72    & 0.572   & 0.67    \\ \bottomrule
\end{tabular}%
\caption{IoU results for Otsu, K-means and FCM methods across the carbonates dataset.}
\label{tab:table_methods_seg}
\end{table}
We report the evolution of the IoU metric for each tested supervised model as a function of the amount of data used during training (Figure~\ref{fig:mean_iou_plot}). For quantitative results, readers can refer to Appendix \ref{appendix:bench}. It is important to note that, in this phase, training is performed using raw data (no denoising). As discussed previously, DINOv2-ConvHead-FT significantly outperforms our best UNet model, even in relatively \say{high} data regimes (above 1,000 training images). The performance gap widens rapidly as the amount of training data decreases. While DINOv2-ConvHead-FT retains most of its performance with as few as 200 training images, the UNet's IoU notably degrades, dropping from 0.78 to 0.65 with 200 training images, which represents a 17\% decrease. This disparity becomes even more pronounced in extremely limited data regimes (below 100 training images), where the UNet significantly overfits to the few training samples available, leading to a catastrophic fall in performance (up to a 50\% loss with only 4 images). Impressively, our DINOv2 segmentation model achieves a mean IoU of 0.74 with as few as 4 training samples for fine-tuning, a performance comparable to a UNet trained from scratch with 1,000 images. Regarding the RF model, it performs remarkably steadily across all data regimes but achieves lower IoU performance than DINOv2-ConvHead-FT. However, it outperforms the UNet model when trained with 500 data points or fewer. This highlights the trade-off between classical machine learning algorithms and deep learning: while the former is intrinsically limited by its simplicity but does not require large amounts of data to perform well, the latter can learn complex tasks but is very data-hungry. Given our results, it appears that DINOv2 combines the generalization capabilities of robust but simple machine learning algorithms with the modeling power of deep learning approaches. Overall, DINOv2-ConvHead-FT achieved the highest IoU scores across all data regimes, underscoring the strong generalization and adaptability of DINOv2 as a backbone.

\section{Discussion}
In one of our primary segmentation experiments, we applied traditional methods (Otsu, K-means, and FCM) to segment sample \textit{S3}. The results, presented in Section \ref{sec:results_benchmark}, reveal a strong dependency on the specific sample, with varying IoU values regardless of the method used. Although the main focus was on segmenting \textit{S3}, it is noteworthy that \textit{S1} achieved the highest IoU, followed by \textit{S3}, and finally \textit{S2}. This highlights the limitations of traditional segmentation methods, which can struggle with similar datasets, underscoring their sensitivity to dataset-specific characteristics.

\begin{figure}[H]
    \centering
    \begin{subfigure}[b]{0.32\textwidth}
        \centering
        \includegraphics[width=\textwidth]{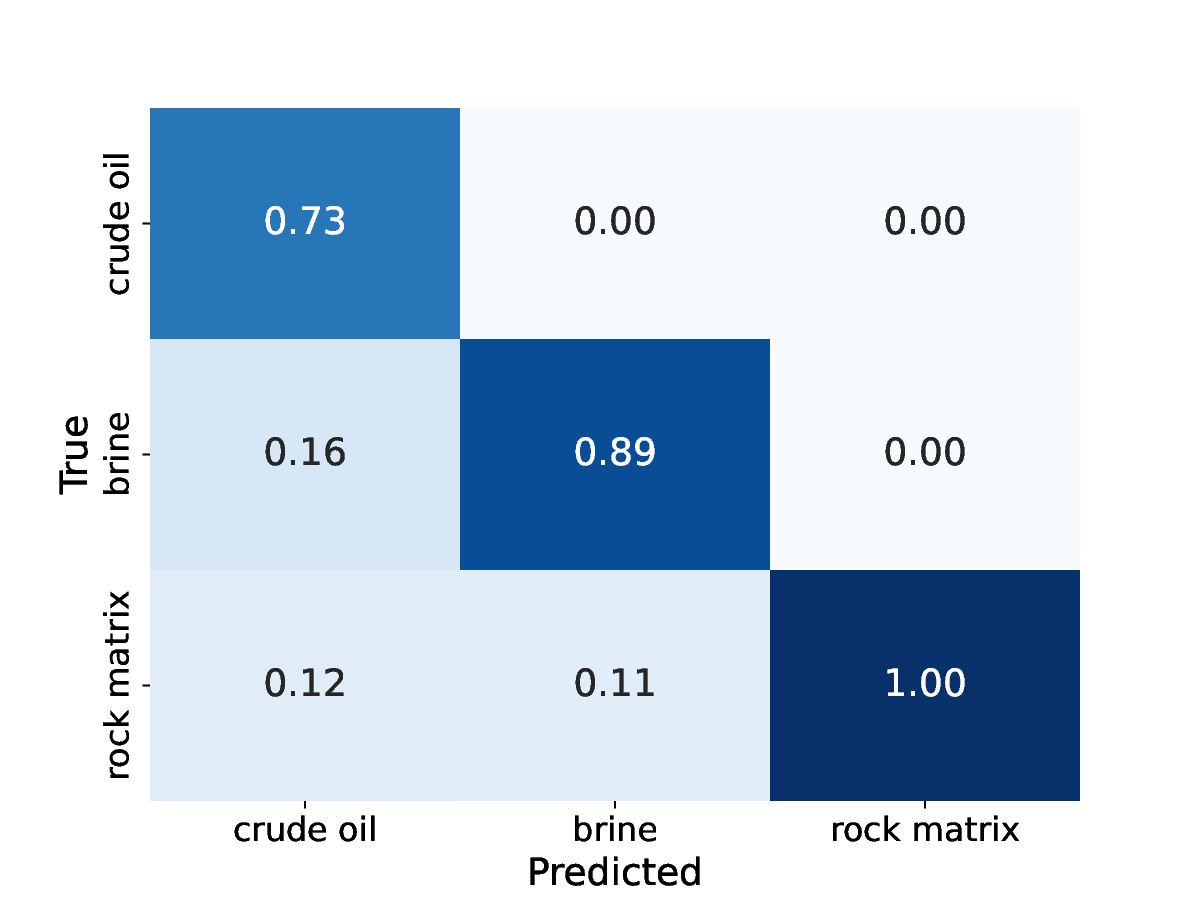}
        \caption{DINOv2}
        \label{fig:dino_confusion_matrix}
    \end{subfigure}
    \hfill
    \begin{subfigure}[b]{0.333\textwidth}
        \centering
        \includegraphics[width=\textwidth]{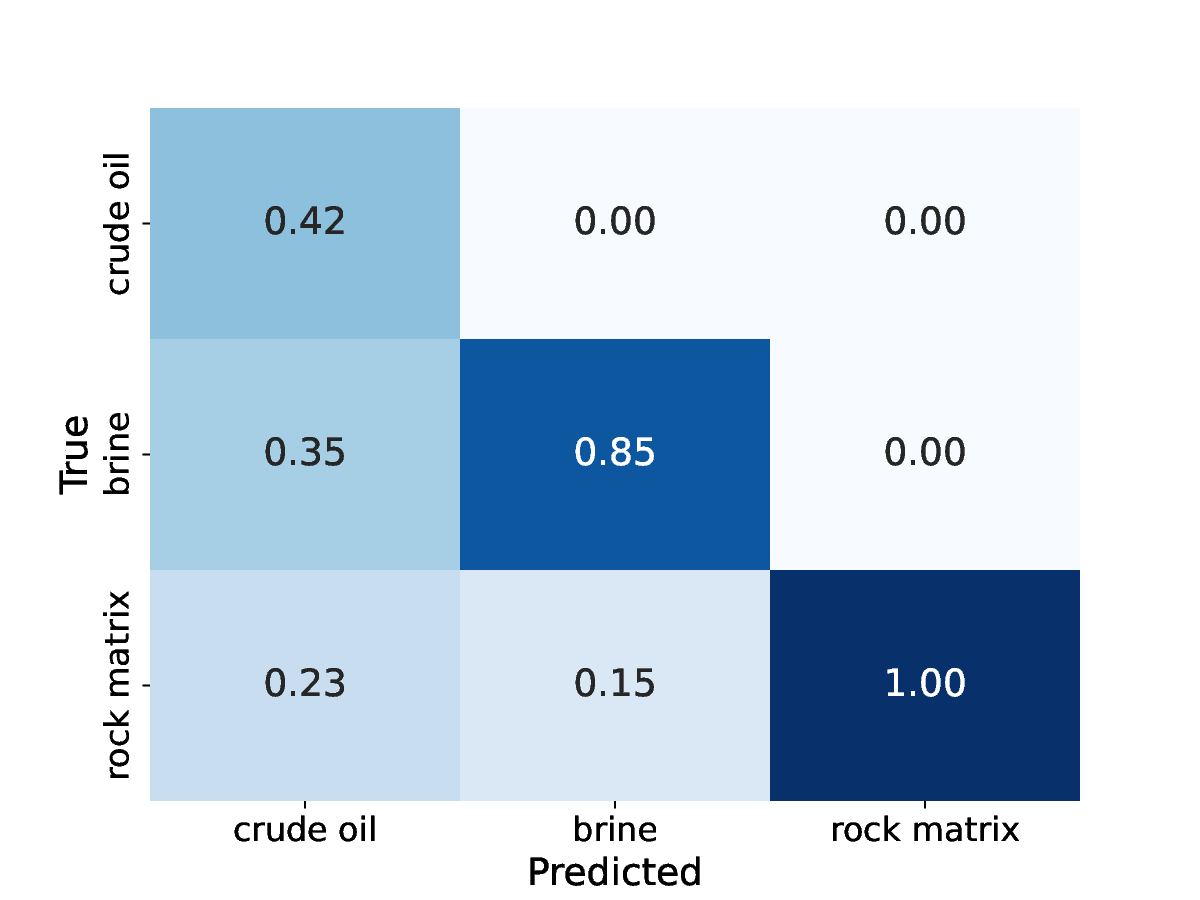}
        \caption{UNet-small}
        \label{fig:unet_confusion_matrix}
    \end{subfigure}
    \hfill
    \begin{subfigure}[b]{0.33\textwidth}
        \centering
        \includegraphics[width=\textwidth]{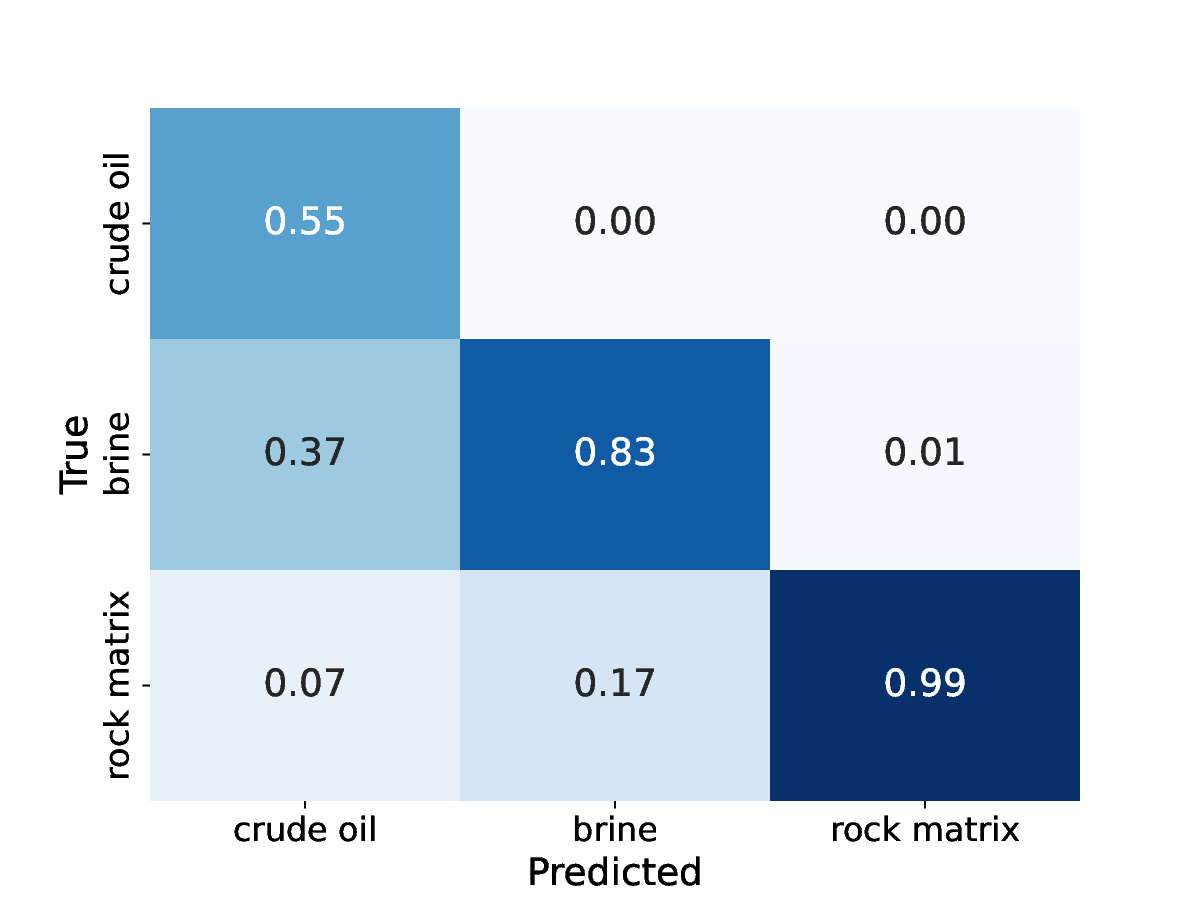}
        \caption{RF}
        \label{fig:RF_confusion_matrix}
    \end{subfigure}
    \caption{Confusion matrices for the segmentation predictions on sample \textit{S3}, trained with 500 images as inputs, for each benchmarked model.}
    \label{fig:confusion_matrices_bench}
\end{figure}

On the supervised front, each architecture was trained on two similar samples (\textit{S1} and \textit{S2}), to predict \textit{S3}. Here, results varied significantly and were much more architecture-dependent. As seen in Figure~\ref{fig:confusion_matrices_bench}, the most prevalent class (rock matrix) was generally the easiest to classify across all models. However, the brine and crude oil classes, with only 12.66\% and 2.41\% representation respectively, presented significant challenges. The DINOv2 model excelled in this regard, correctly classifying 89\% of brine and 72\% of crude oil pixels, despite the under-representation and low contrast in these classes (Cf. Section \ref{sec:dataset}). In contrast, the UNet and RF models achieved only 42\% and 55\% accuracy for crude oil, respectively. This highlights DINOv2's superior ability to generalize, especially with under-represented classes, demonstrating its robustness in handling complex datasets. However, specialized foundation models for segmentation such as SAM were not considered. Despite prior studies concluding that DINOv2 outperforms SAM in segmentation tasks, it has not been yet confirmed that this also holds for rock segmentation. We leave this for future work. Also, there are limitations to our approach. A key one is that we did not push the DINOv2 model's architecture to its full potential due to computational constraints. Our results suggest that performance correlates with feature size (Cf. Section \ref{sec:probing}), indicating that a larger DINOv2 backbone or more sophisticated model architecture (e.g., a hybrid ResNet-DINOv2 encoder with a UNet-like decoder as described in TransUNet \cite{chen2021transunet}) could further enhance performance. Incorporating intermediate layer features could also yield additional benefits by improving the capture of finer details in segmentation tasks (Cf. Table~\ref{table_probing}).

However, specialized foundation models for segmentation, such as SAM, were not considered. Despite prior studies concluding that DINOv2 outperforms SAM in segmentation tasks, it has not yet been confirmed whether this also holds for rock segmentation. We leave this for future work. Additionally, there are limitations to our approach. A key limitation is that we did not push the DINOv2 model's architecture to its full potential due to computational constraints. Our results suggest that performance correlates with feature size (see Section \ref{sec:probing}), indicating that a larger DINOv2 backbone or a more sophisticated model architecture (e.g., a hybrid ResNet-DINOv2 encoder with a UNet-like decoder as described in TransUNet \cite{chen2021transunet}) could further enhance performance. Incorporating intermediate layer features could also yield additional benefits by improving the capture of finer details in segmentation tasks (see Table~\ref{table_probing}).

A deeper reflection on DINOv2’s limitations arises from Section \ref{sec:results_pca}, where an analysis of its raw and fine-tuned features using PCA indicates that, before fine-tuning, the model struggles to distinguish the meaningful components of a CT-scan of rock sample. This observation suggests that while DINOv2's pre-trained features provide a solid foundation, further adaptation is necessary to optimize their performance for highly specialized tasks like rock segmentation. The reliance on fine-tuning highlights potential inadequacies in the raw model when applied to domain-specific data. Further experimentation could help determine whether fine-tuning across different data subsets consistently leads to improvements or yields divergent results.

In our classification study on the sandstones dataset, we found that DINOv2 effectively distinguished geological patterns using its latent representations (Cf. Figures \ref{fig:tsne-seg} \& \ref{fig:tsne-combined}). The t-SNE visualization of its feature space showed cohesive clusters that corresponded to samples from the same geological dataset, affirming the model’s ability to capture and represent geological similarities. These findings validate DINOv2’s sensitivity to subtle rock features, even without explicit training in this domain. The t-SNE plot further demonstrated a well-structured, interpretable feature space, with each sample clustering meaningfully and without misclassification. This highlights DINOv2’s robustness in rock classification, as its untreated features already contain much of the relevant information for further analysis. We emphasize 1) the strength of a non-fine-tuned DINOv2 in organizing geological datasets based on petromorphological traits, and 2) the potential for DINOv2 to uncover hidden geological features not easily detected through traditional methods. These results underscore the model's adaptability for geological analysis and its potential for broader applications in pattern recognition across various domains.

\begin{figure*}[!ht]
    \centering
    \begin{subfigure}{0.19\textwidth}
        \centering
        \includegraphics[width=\textwidth]{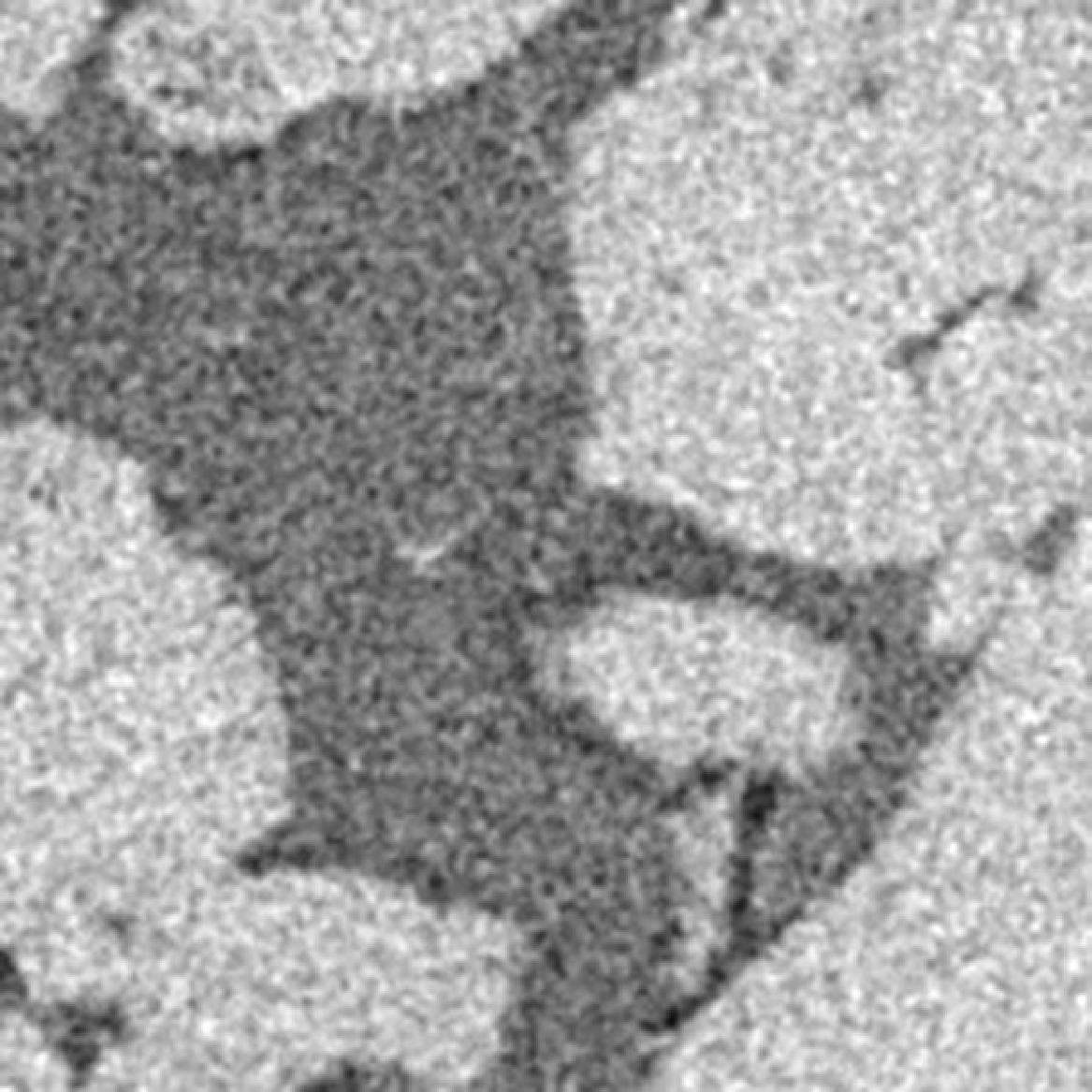}
    \end{subfigure}
    \hfill
    \begin{subfigure}{0.19\textwidth}
        \centering
        \includegraphics[width=\textwidth]{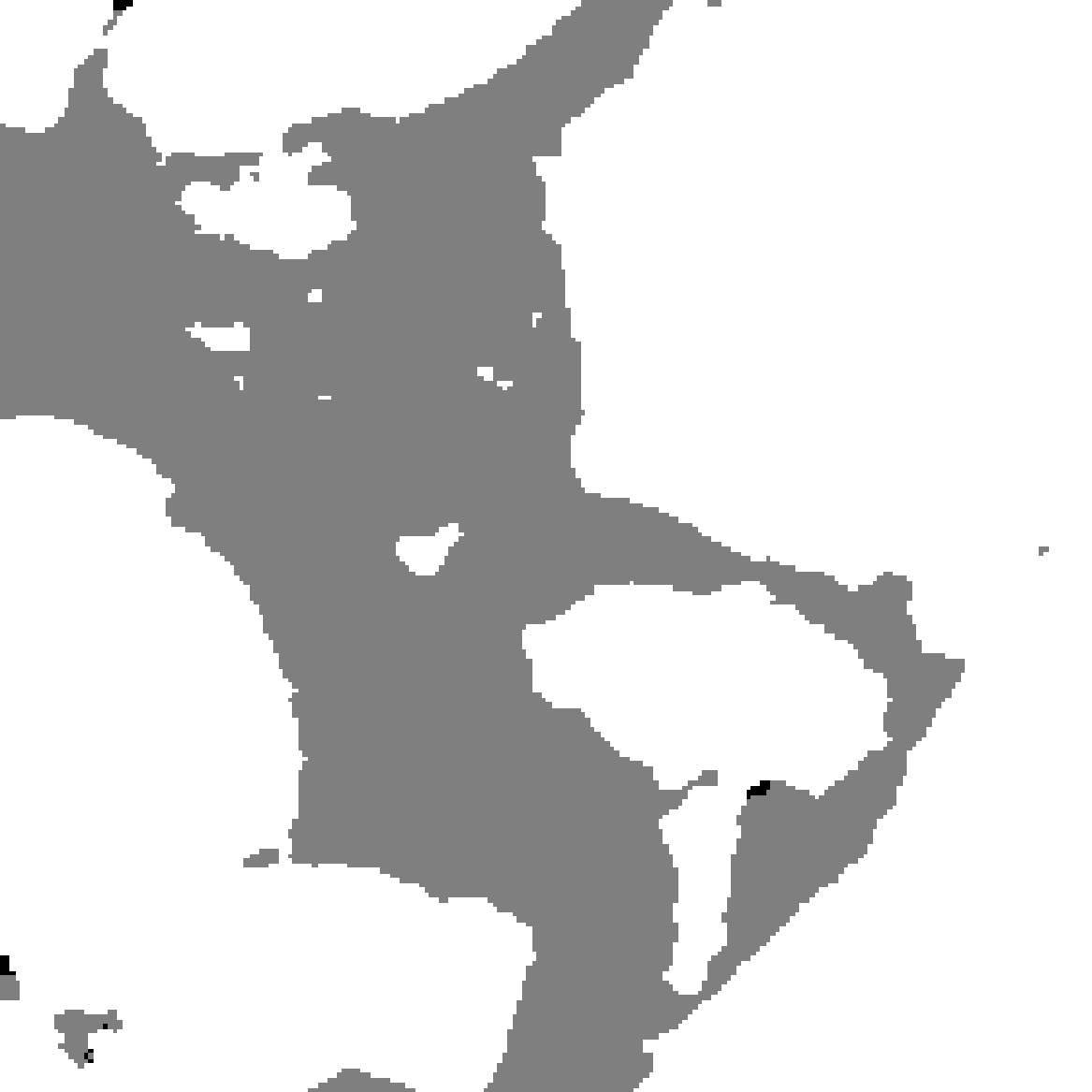}
    \end{subfigure}
    \hfill
    \begin{subfigure}{0.19\textwidth}
        \centering
        \includegraphics[width=\textwidth]{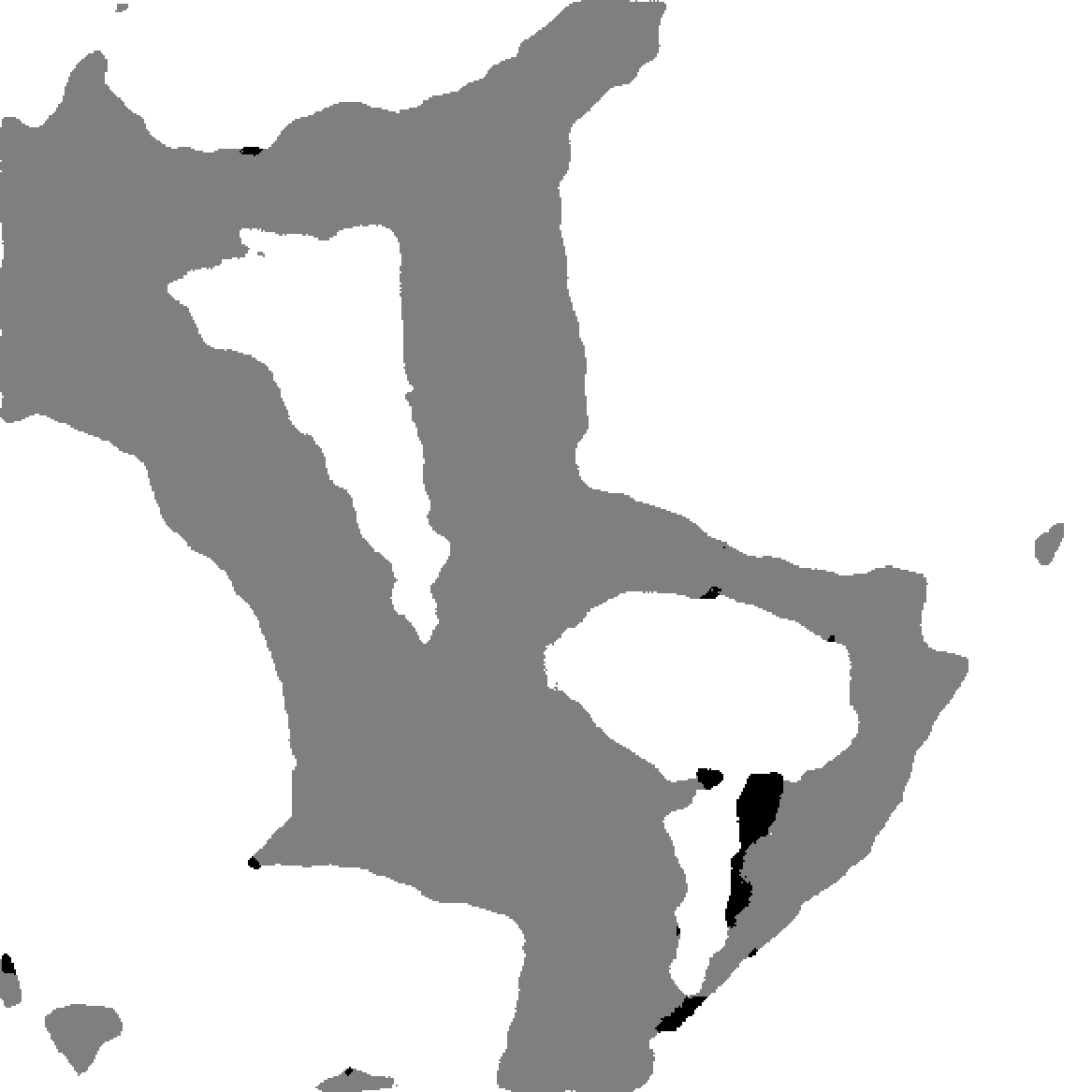}
    \end{subfigure}
    \hfill
    \begin{subfigure}{0.19\textwidth}
        \centering
        \includegraphics[width=\textwidth]{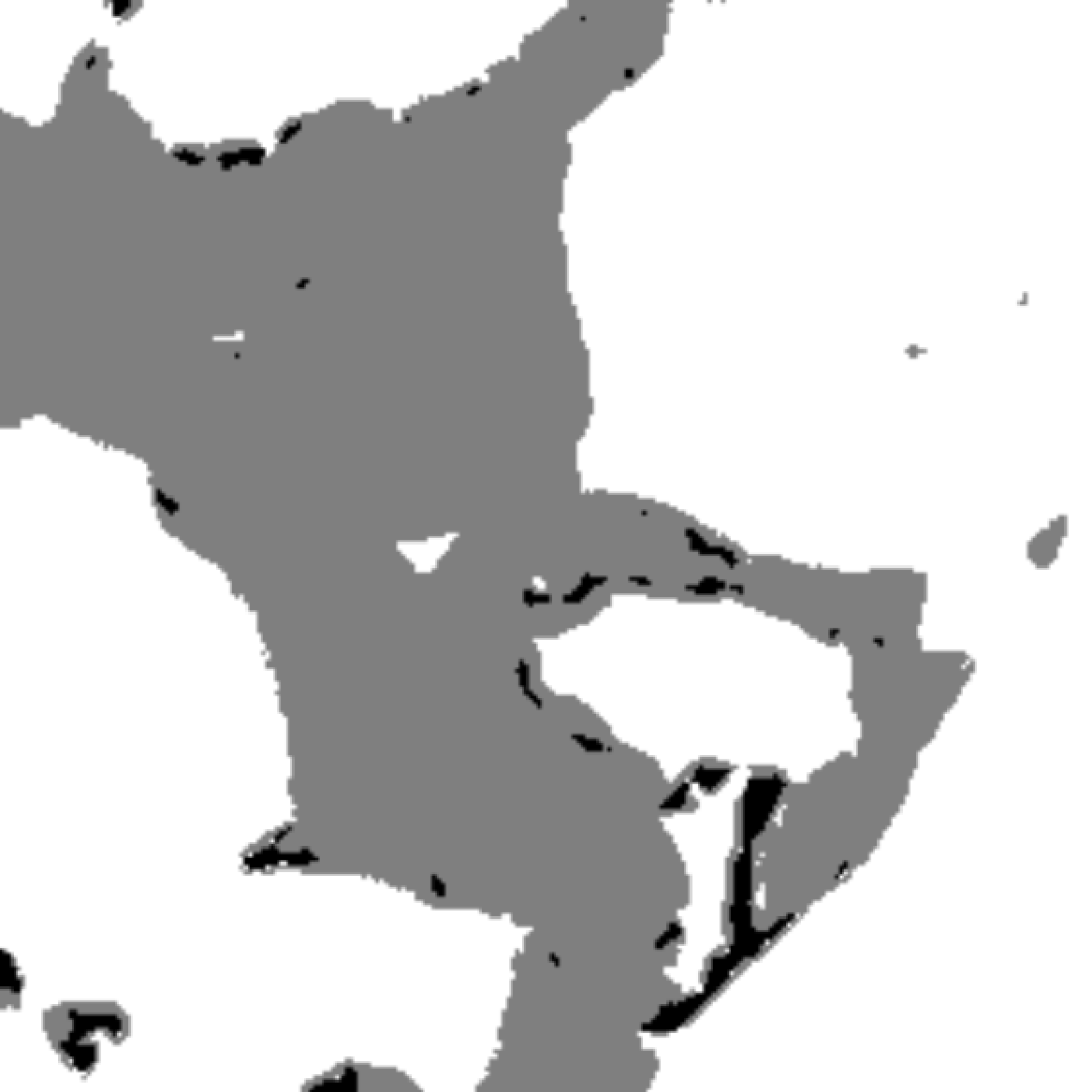}
    \end{subfigure}
    \hfill
    \begin{subfigure}{0.19\textwidth}
        \centering
        \includegraphics[width=\textwidth]{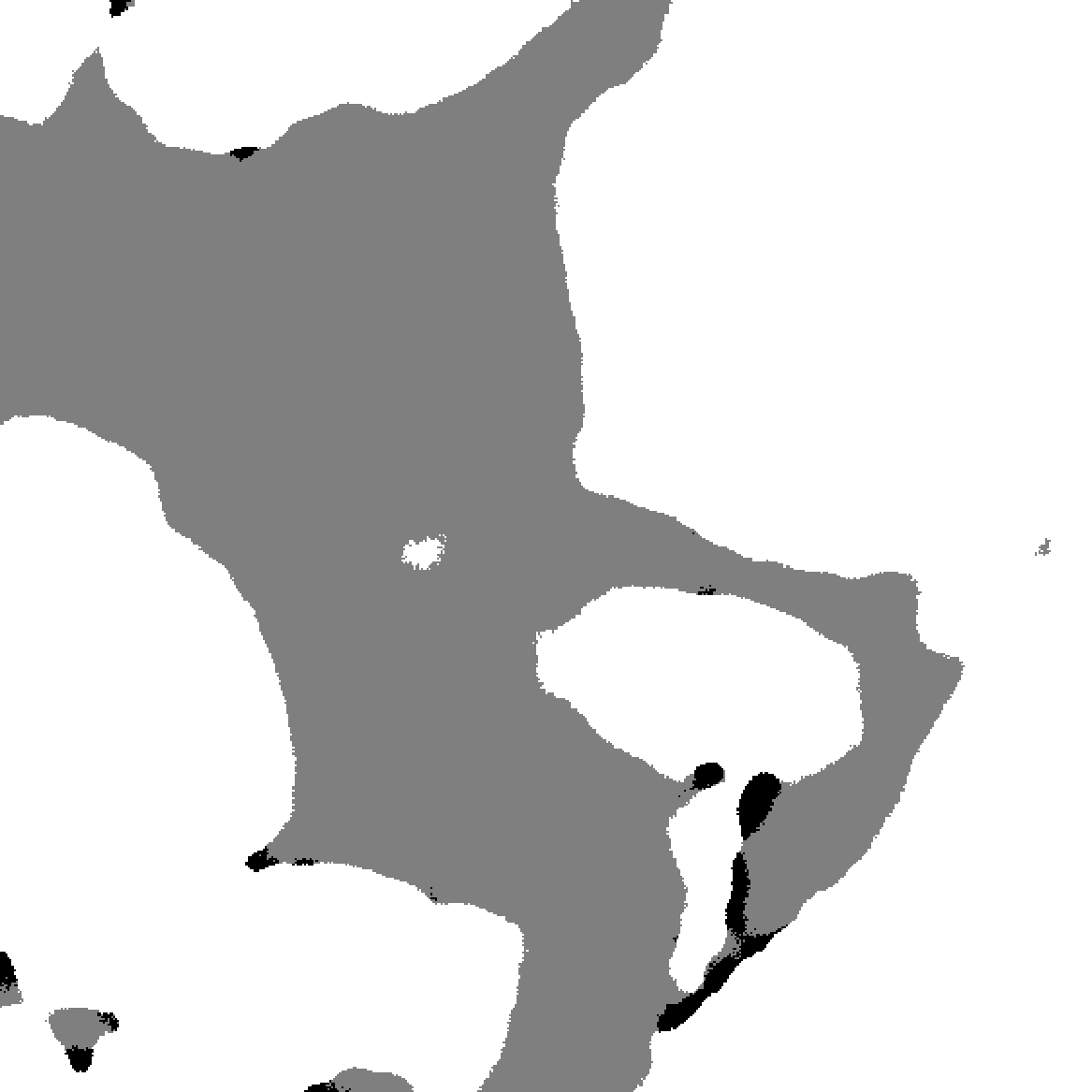}
    \end{subfigure}
    
    \begin{subfigure}{0.19\textwidth}
        \centering
        \includegraphics[width=\textwidth]{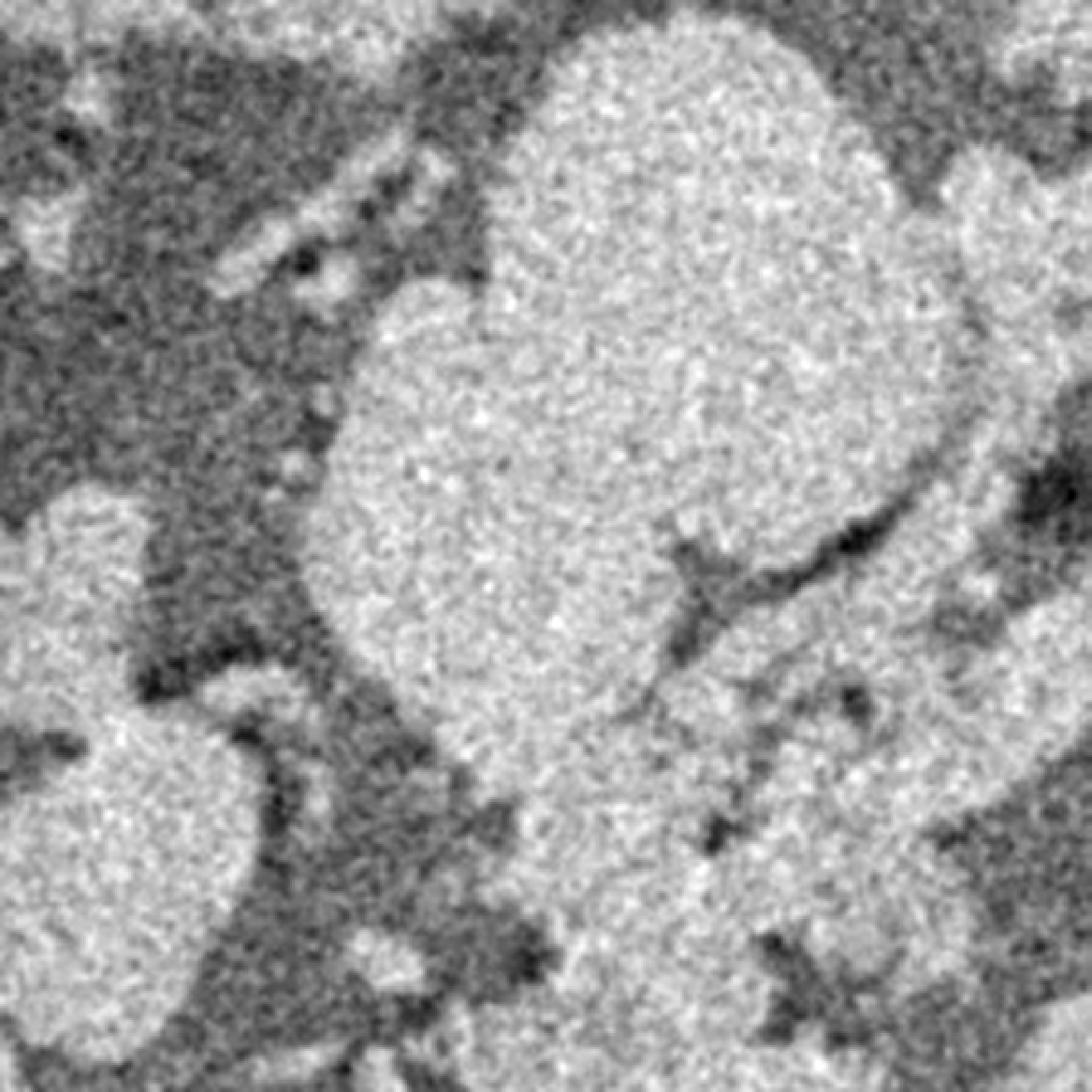}
    \end{subfigure}
    \hfill
    \begin{subfigure}{0.19\textwidth}
        \centering
        \includegraphics[width=\textwidth]{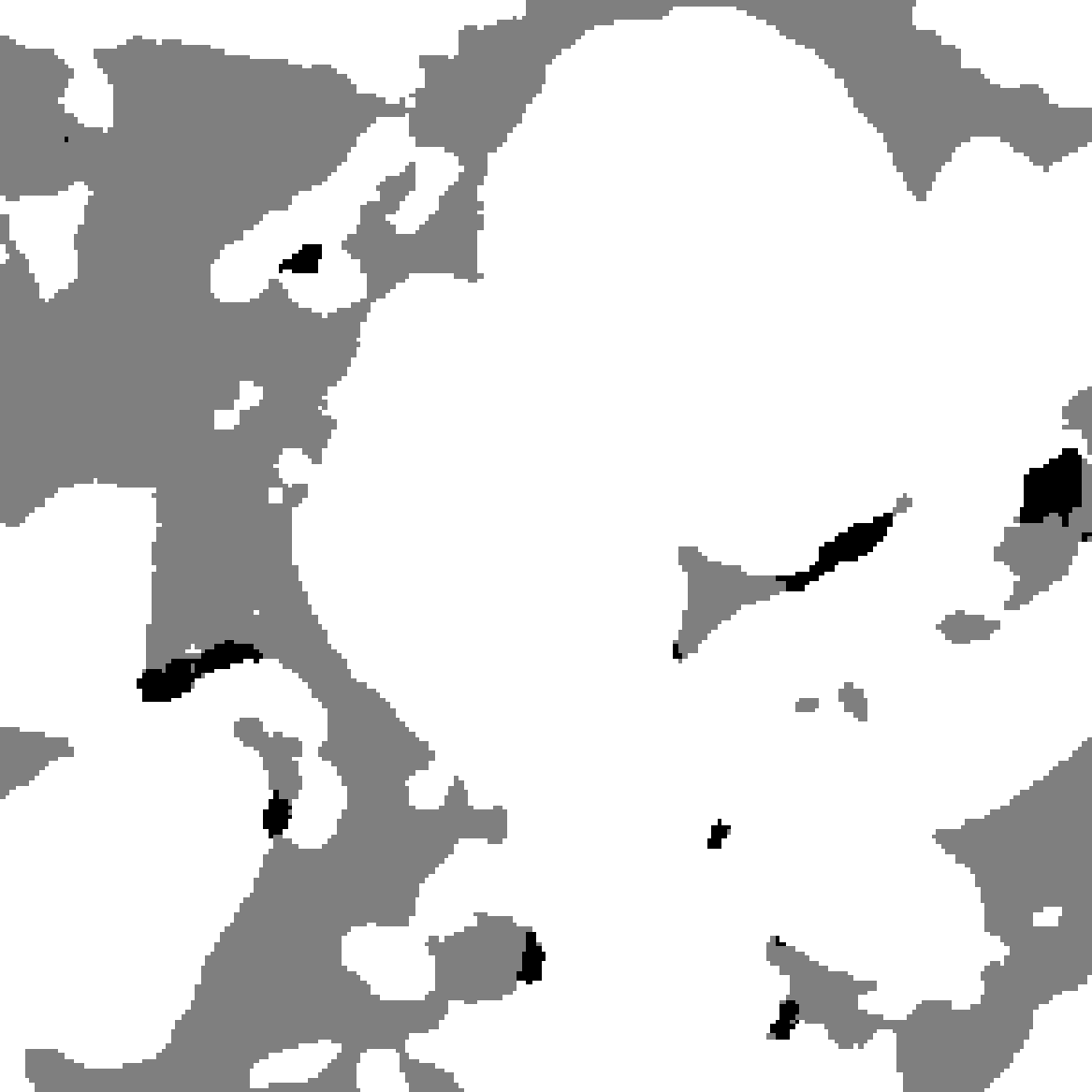}
    \end{subfigure}
    \hfill
    \begin{subfigure}{0.19\textwidth}
        \centering
        \includegraphics[width=\textwidth]{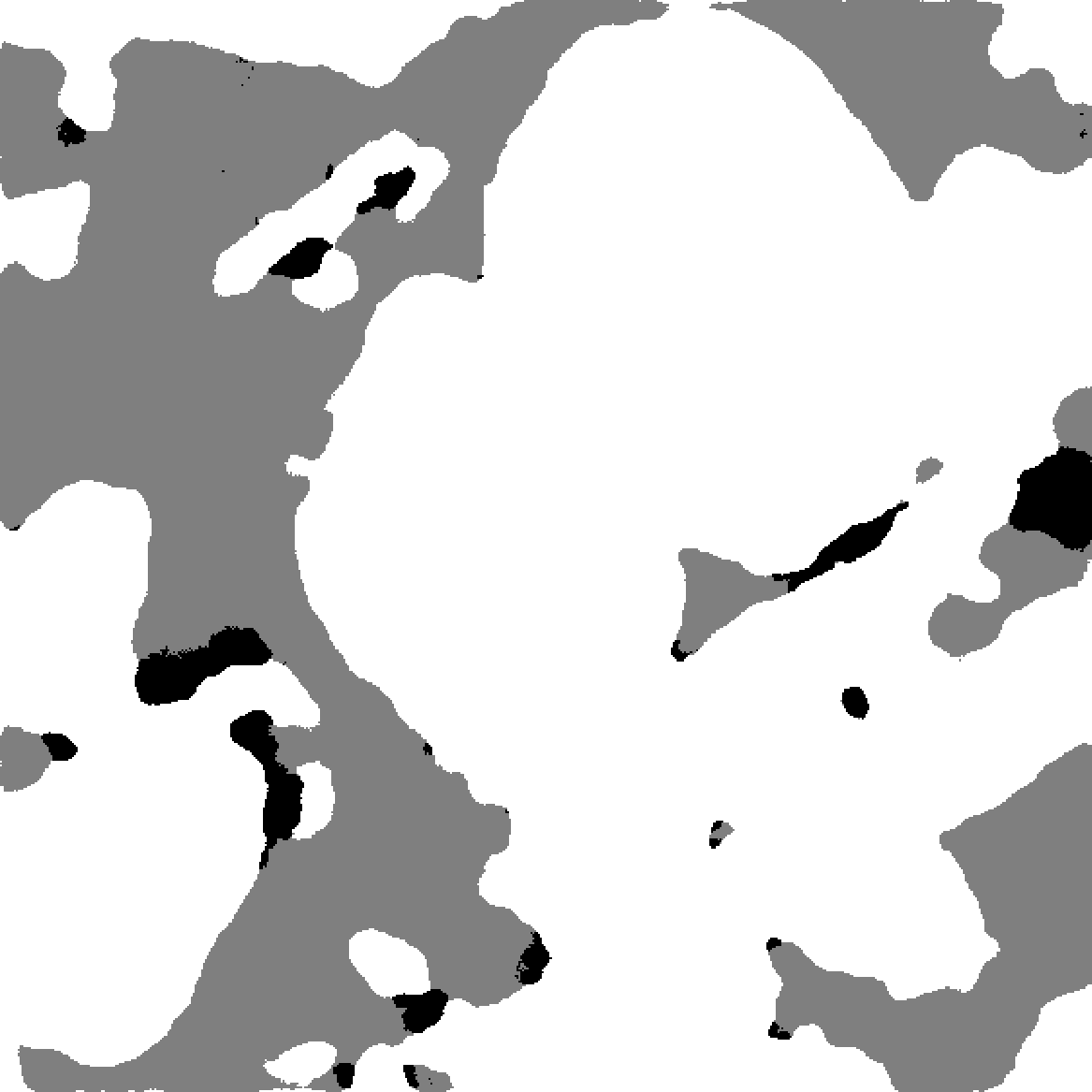}
    \end{subfigure}
    \hfill
    \begin{subfigure}{0.19\textwidth}
        \centering
        \includegraphics[width=\textwidth]{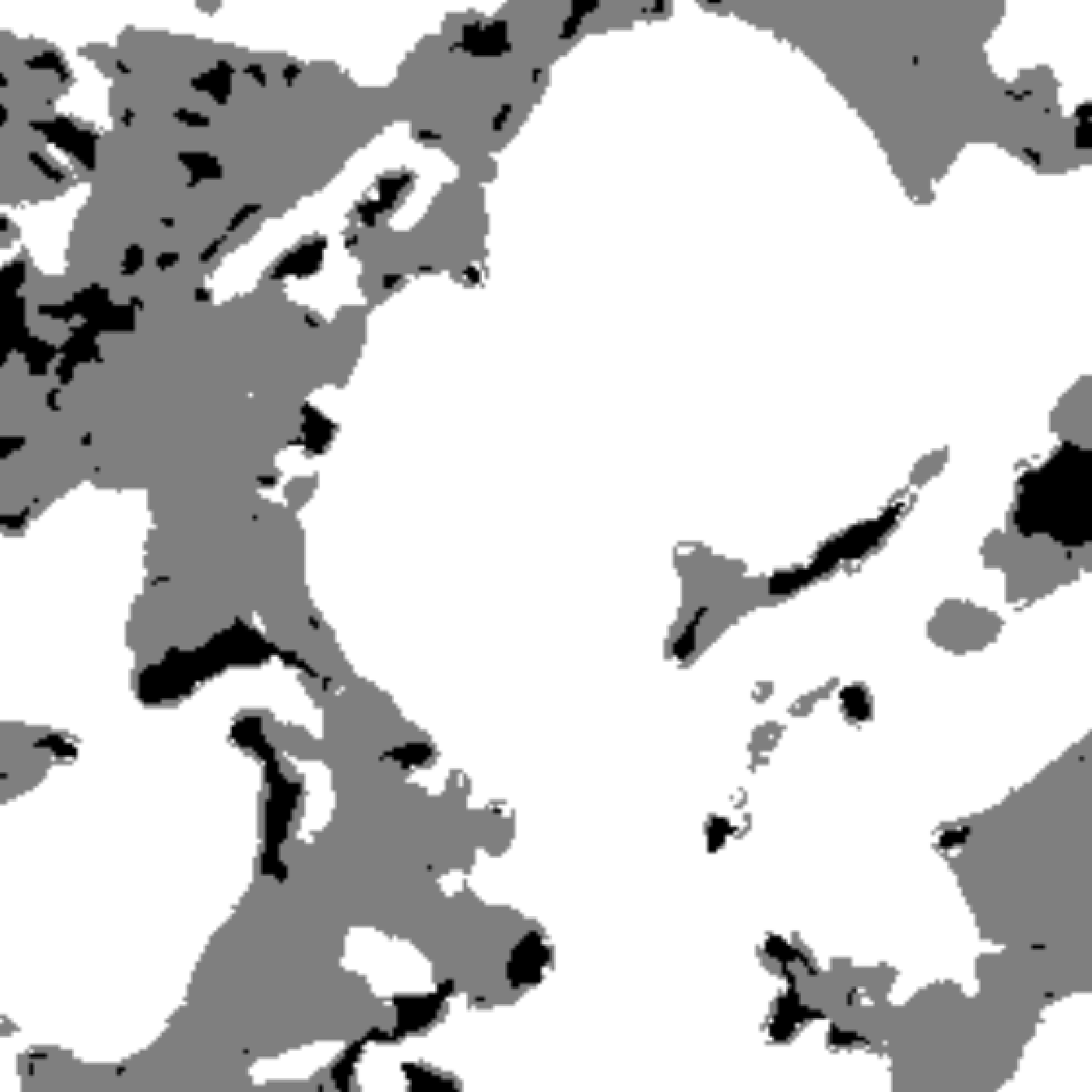}
    \end{subfigure}
    \hfill
    \begin{subfigure}{0.19\textwidth}
        \centering
        \includegraphics[width=\textwidth]{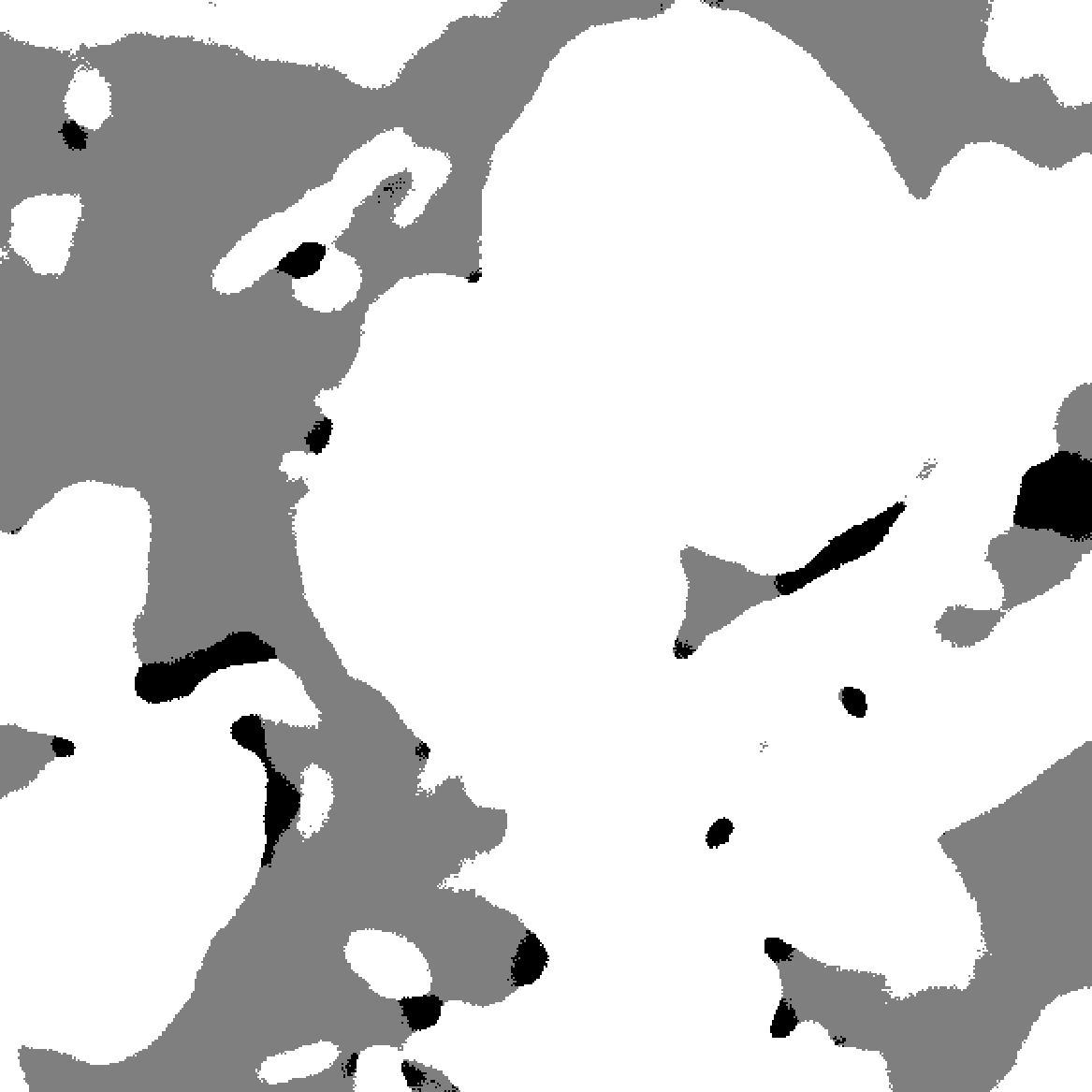}
    \end{subfigure}
    
    \begin{subfigure}{0.19\textwidth}
        \centering
        \includegraphics[width=\textwidth]{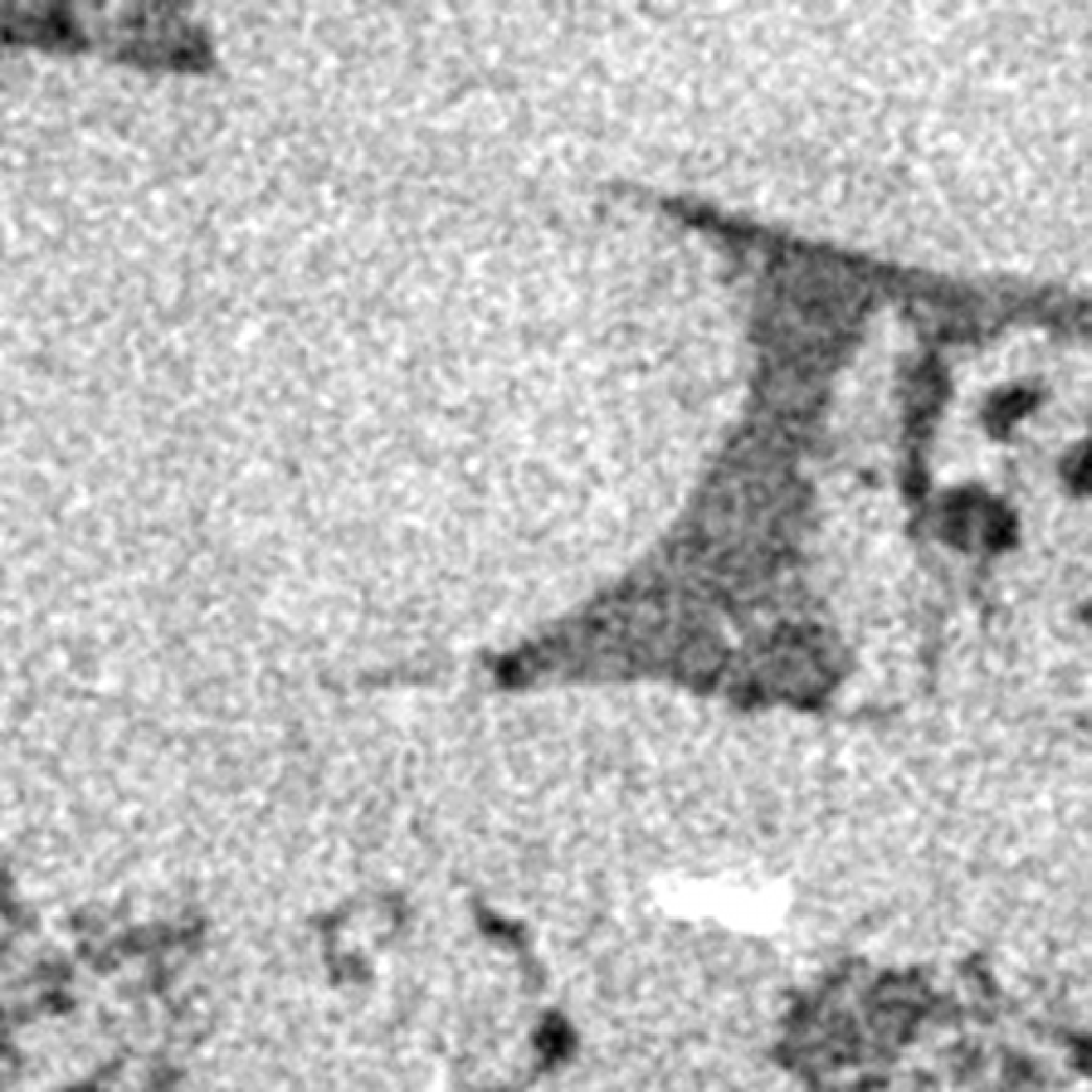}
        \caption{CT-scan}
    \end{subfigure}
    \hfill
    \begin{subfigure}{0.19\textwidth}
        \centering
        \includegraphics[width=\textwidth]{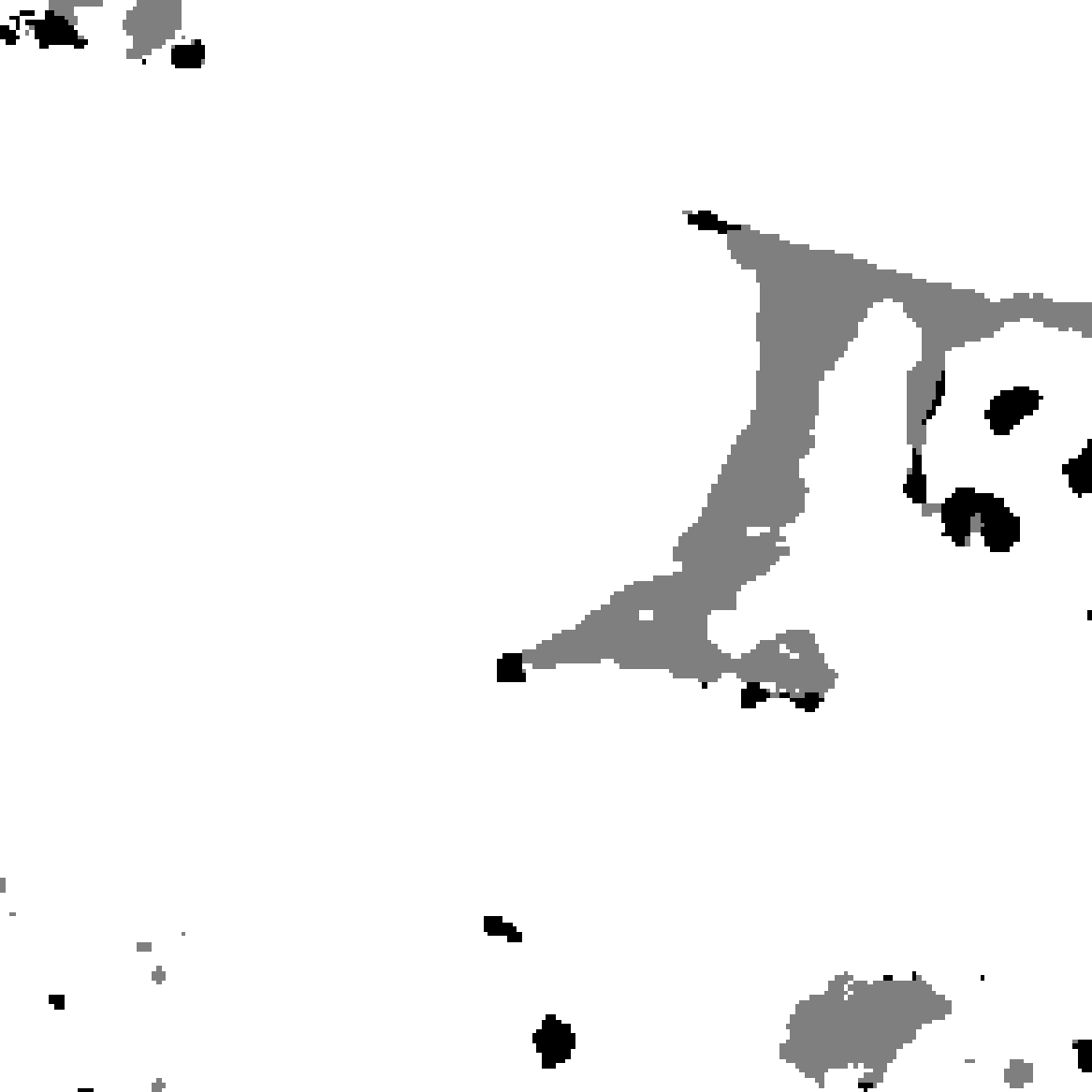}
        \caption{GT}
    \end{subfigure}
    \hfill
    \begin{subfigure}{0.19\textwidth}
        \centering
        \includegraphics[width=\textwidth]{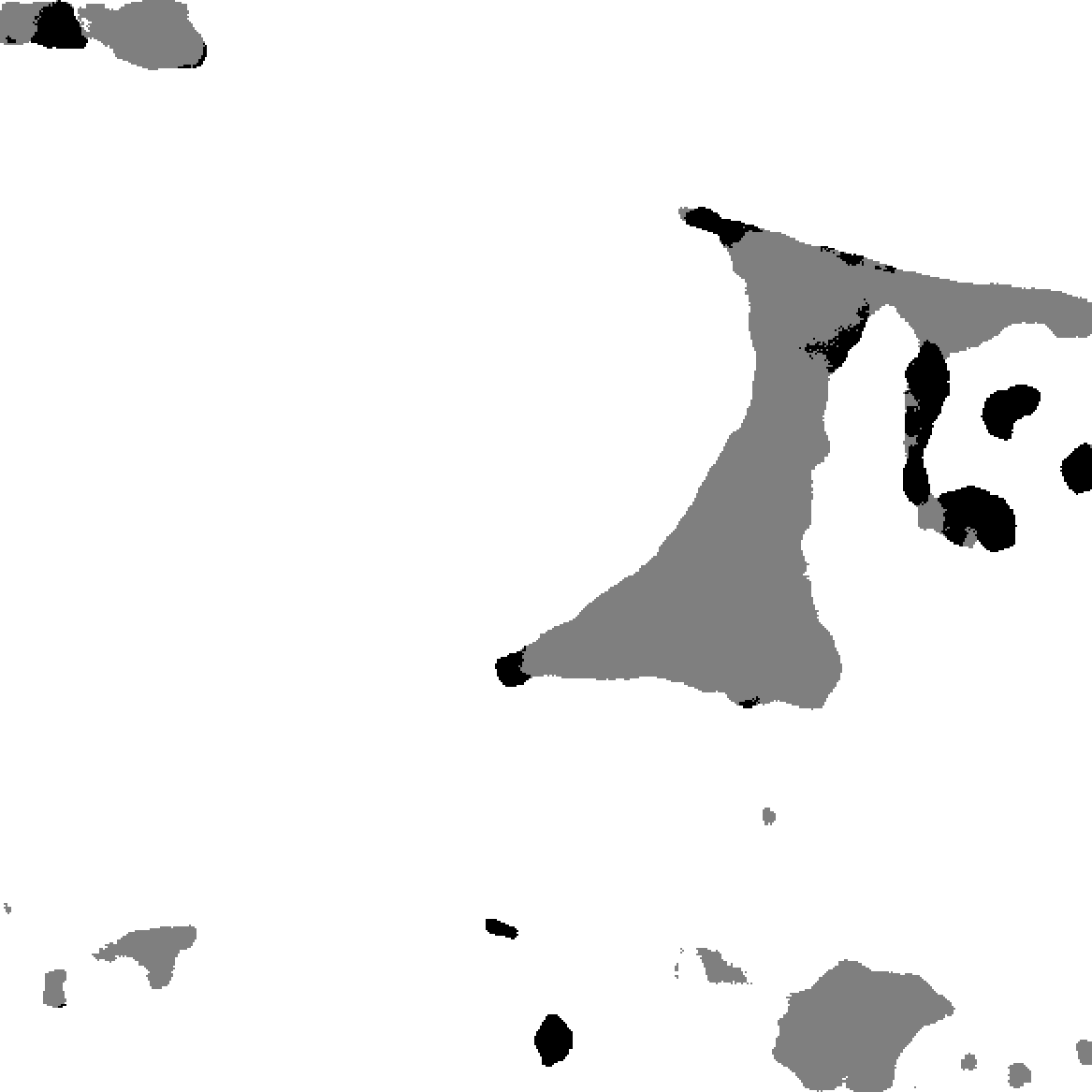}
        \caption{UNet}
    \end{subfigure}
    \hfill
    \begin{subfigure}{0.19\textwidth}
        \centering
        \includegraphics[width=\textwidth]{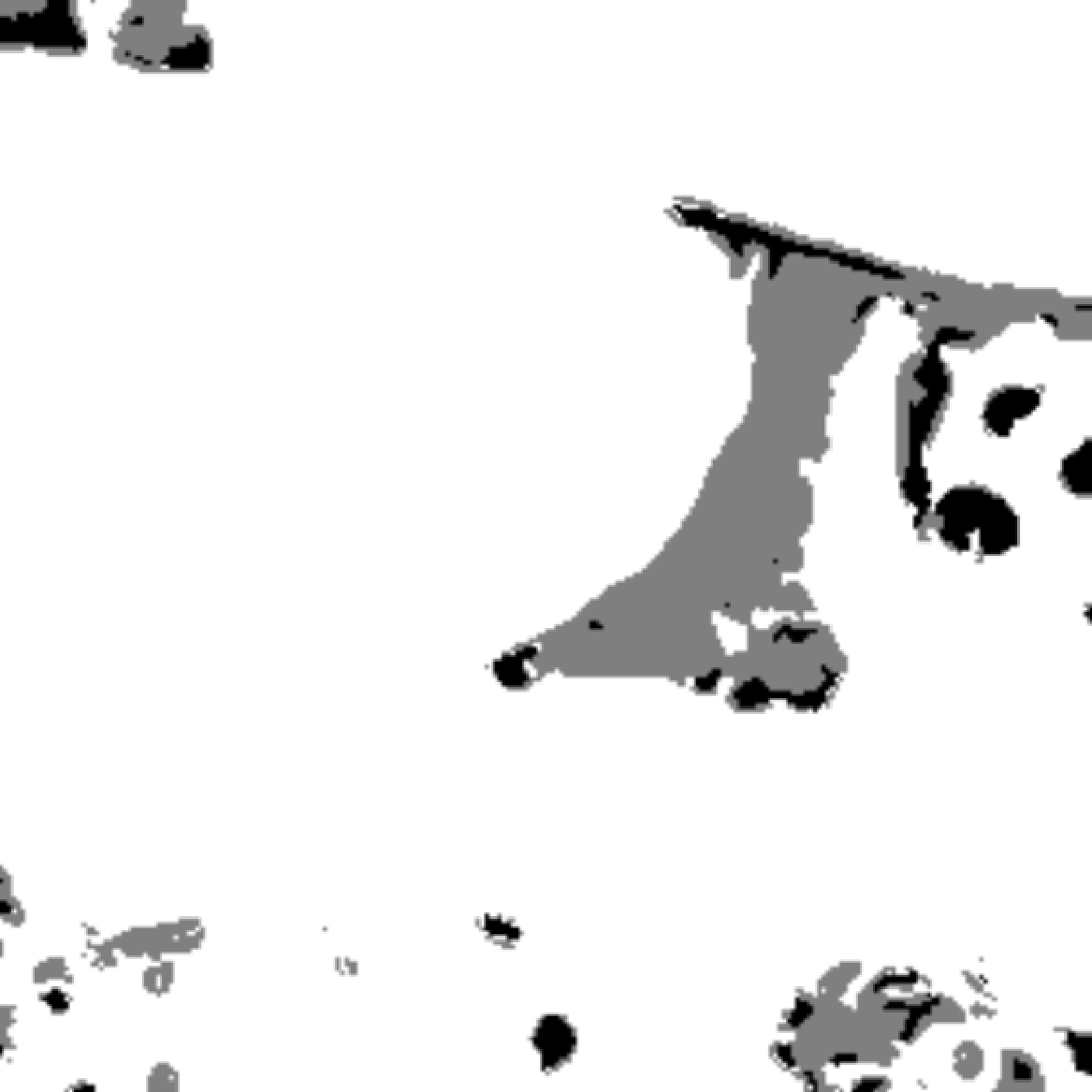}
        \caption{RF}
    \end{subfigure}
    \hfill
    \begin{subfigure}{0.19\textwidth}
        \centering
        \includegraphics[width=\textwidth]{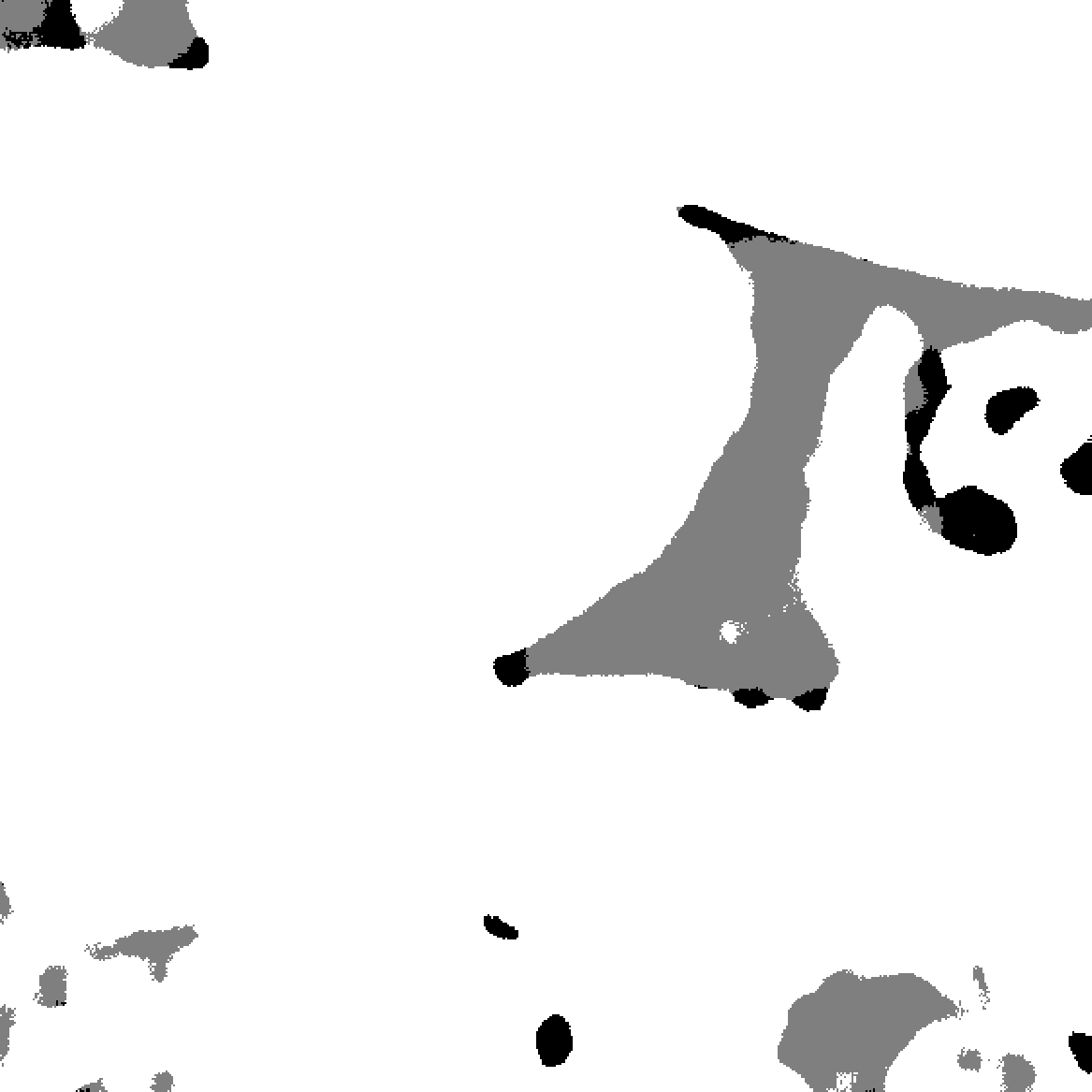}
        \caption{DINOv2}
    \end{subfigure}
    
    \caption{Predicted masks for the UNet model, the RF and our DINOv2 segmentation models for three different scanners with low-quality GT.}
    \label{fig:gt_reliability}
\end{figure*}

Finally, we also reflect on the reliability of the GT used in this study. Indeed, despite achieving the best IoUs using DINOv2, it is important to consider whether this translates into an accurate physical model. As previously noted in studies \cite{liang2022, da2021, pham2023}, the labels used for the segmentation workflow are never perfect. As such, the IoU metrics quantify the agreement between the predicted labeled images and the GT, rather than the exact distribution of the minerals. In some cases, we postulate that DINOv2's predicted labels can be qualitatively more accurate than the training labels (Figure \ref{fig:gt_reliability}). Moreover, the segmentation results from the DINOv2 model can be less affected by human subjectivity, which is a significant advantage for reliable physical properties estimation in systematic studies such as those involving binary and/or multi-class segmentation of rock CT-scans \cite{liang2022}.

\newpage
\section{Conclusion}
This study explores the applicability of DINOv2 for segmenting images, with a focus on assessing the model's effectiveness in µCT-scanned geological image analysis. Our results show that DINOv2's features effectively capture and distinguish subtle geological patterns, enabling accurate classification and segmentation tasks. Linear and kNN probing on DINOv2's features significantly outperformed BFEs, achieving near-supervised performance without fine-tuning. Additionally, t-SNE and PCA analyses highlight DINOv2's inherent ability to represent geological data, guiding the design of an efficient segmentation head. Our fine-tuning experiments demonstrate that a relatively simple segmentation head, when combined with the fine-tuned DINOv2-base model, achieves superior accuracy compared to more complex models like UNet and ResNet152. Finally, DINOv2 consistently outperforms traditional segmentation methods such as Otsu thresholding, K-means, FCM, and RF, particularly excelling in low-data settings and showing robustness to class imbalance.

Despite these promising results, several open questions remain, offering new research directions. Future work could explore hybrid models combining DINOv2 with advanced architectures like TransUNet, as well as sophisticated fine-tuning and domain adaptation techniques. Moreover, all supervised training was done without any processing of the images. Addressing the issue of noisy scanner images and assessing the physical accuracy of predicted masks are also promising areas for further investigation.

In conclusion, DINOv2 demonstrates substantial potential for enhancing geological image analysis, particularly in challenging data scenarios. We hope this study highlights DINOv2's capabilities and encourages its broader adoption within the geoscientific community.

\section*{Acknowledgment}
The Akkodis Group funded this research as part of the Semantic Classification of High-resolution Imaging for Scanned Materials (SCHISM) project. This study is an integral part of the SCHISM project, which is incorporated within the "Decarbonized Industry" innovation line of Akkodis Research. The project aims to explore and develop innovative solutions for \ce{CO2} management and storage, contributing significantly to the decarbonization of industry. We extend our gratitude to our colleagues at the Akkodis Research Department for their support and commitment. More specifically, we would like to thank Mehdi Mounsif, Zoran Adam Gaxotte, Arthur Rocha, Guillaume Vangilluwen and Michael Chepy for their valuable insights and feedbacks.

\appendix
\newpage
\section{Probing segmentation masks}
\label{appendix:probing}

\begin{figure}[h!]
    \centering
    \begin{subfigure}[b]{0.49\textwidth}
        \centering
        \includegraphics[width=1.25\textwidth]{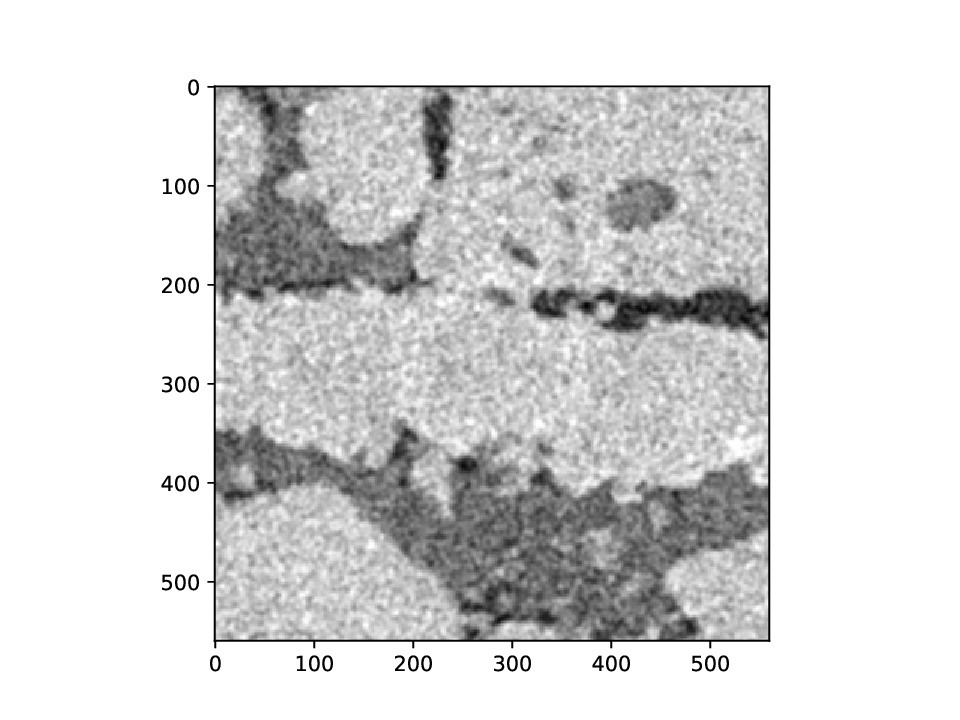}
        \caption{CT-scanner}
        \label{fig:ct_scanner_knn_probing}
    \end{subfigure}
    \hfill
    \begin{subfigure}[b]{0.49\textwidth}
        \centering
        \includegraphics[width=1.25\textwidth]{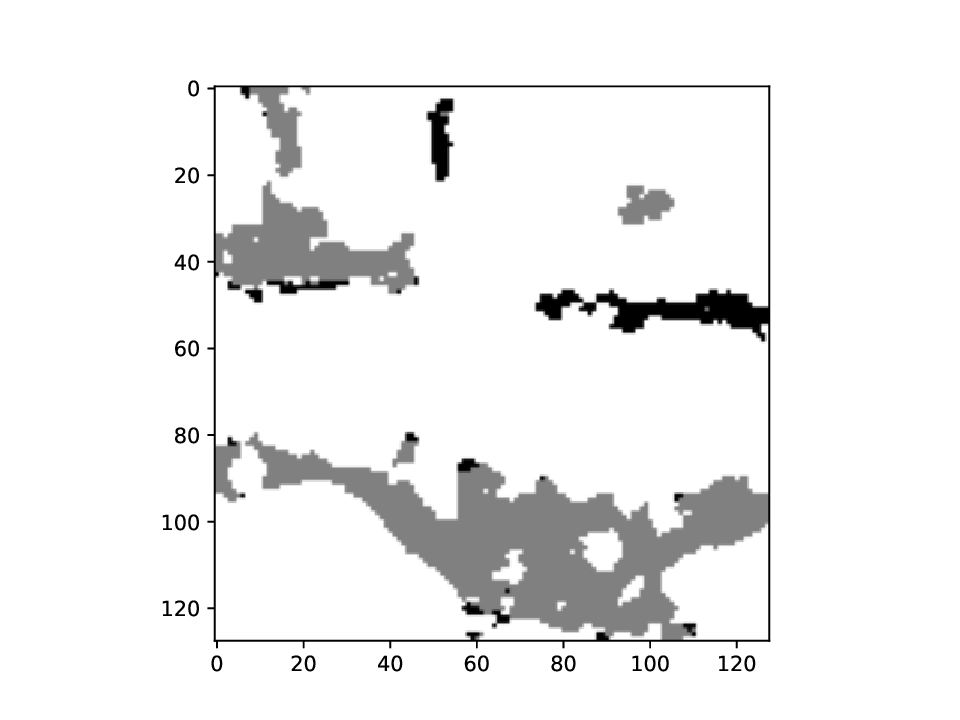}
        \caption{GT}
        \label{fig:knn_gt}
    \end{subfigure}
    \hfill
    \begin{subfigure}[b]{0.49\textwidth}
        \centering
        \includegraphics[width=1.25\textwidth]{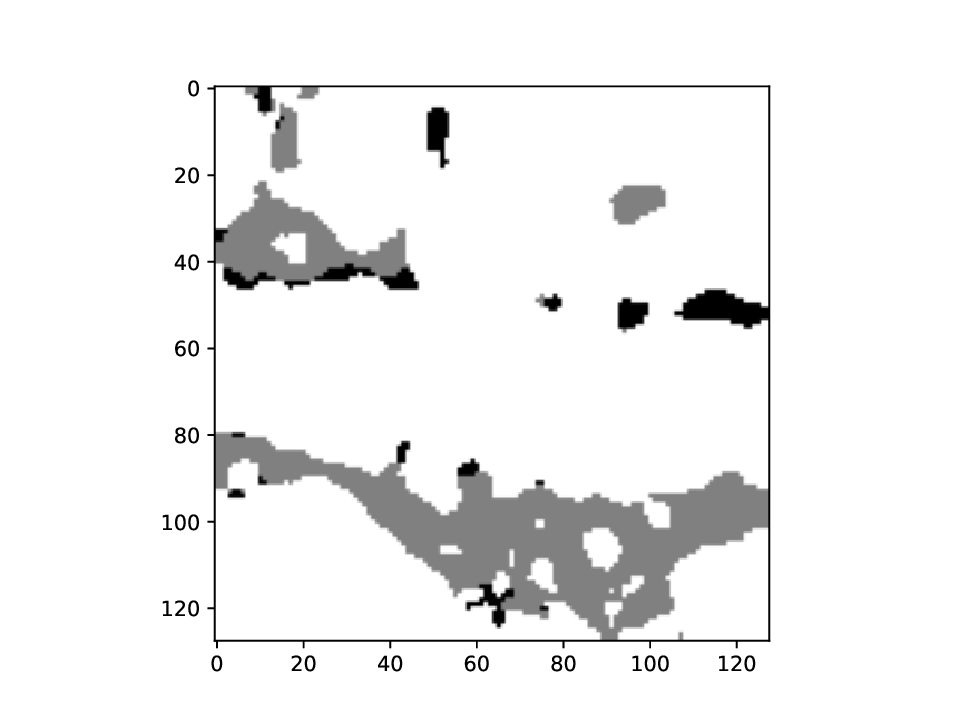}
        \caption{DINOv2}
        \label{fig:dino_knn_pred}
    \end{subfigure}
    \hfill
    \begin{subfigure}[b]{0.49\textwidth}
        \centering
        \includegraphics[width=1.25\textwidth]{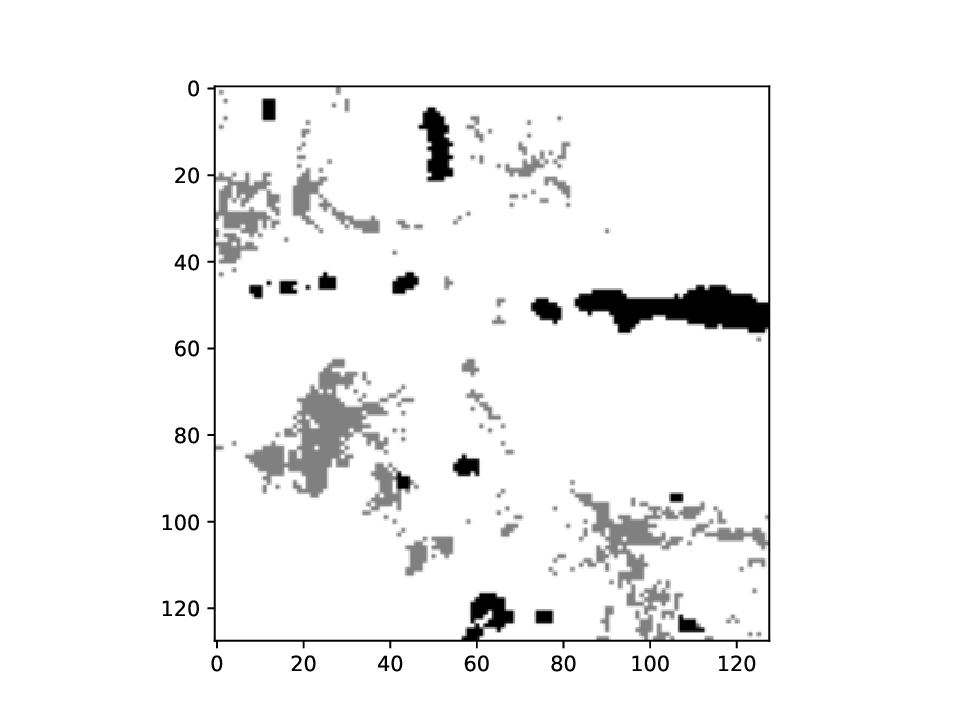}
        \caption{BFE}
        \label{fig:bfe_knn_pred}
    \end{subfigure}
    \caption{Prediction samples for kNN probing with DINOv2 and BFE as feature extractor.}
    \label{fig:knn_probing}
\end{figure}

\newpage
\section{Benchmark segmentation masks}
\label{appendix:bench}
\renewcommand{\arraystretch}{0.01}
\begin{figure*}[h!]
    \centering
    \resizebox{\textwidth}{!}{
    \begin{tabular}{ccccc}
        \begin{subfigure}[t]{0.2\textwidth}
            \includegraphics[width=1.35\linewidth]{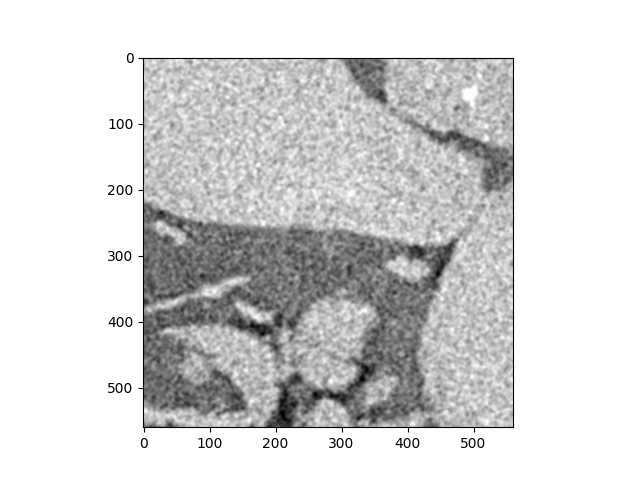}
        \end{subfigure} &
        \begin{subfigure}[t]{0.2\textwidth}
            \includegraphics[width=1.35\linewidth]{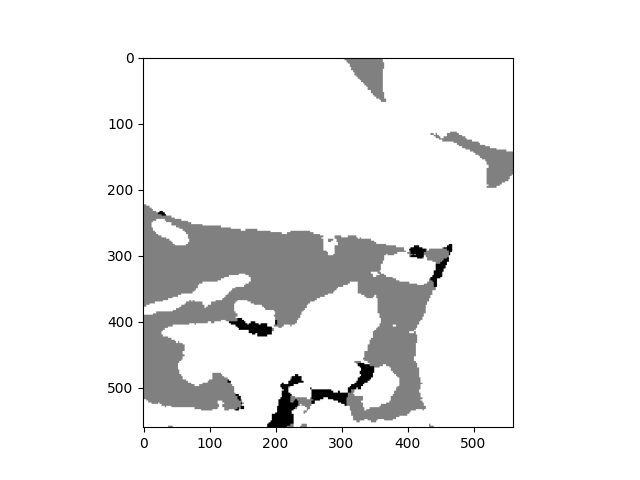}
        \end{subfigure} &
        \begin{subfigure}[t]{0.2\textwidth}
            \includegraphics[width=1.35\linewidth]{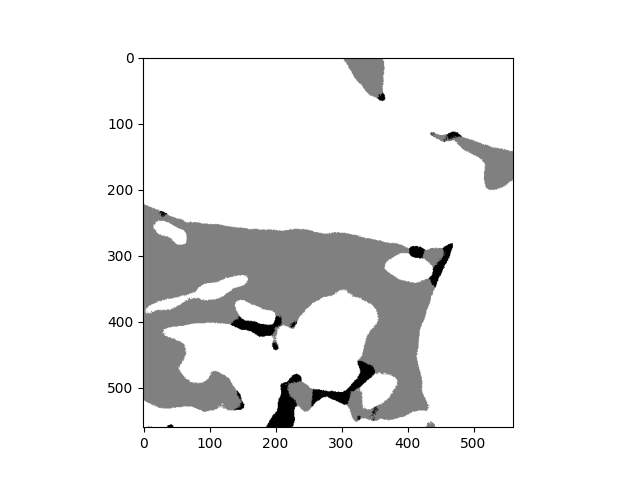}
        \end{subfigure} &
        \begin{subfigure}[t]{0.2\textwidth}
            \includegraphics[width=1.35\linewidth]{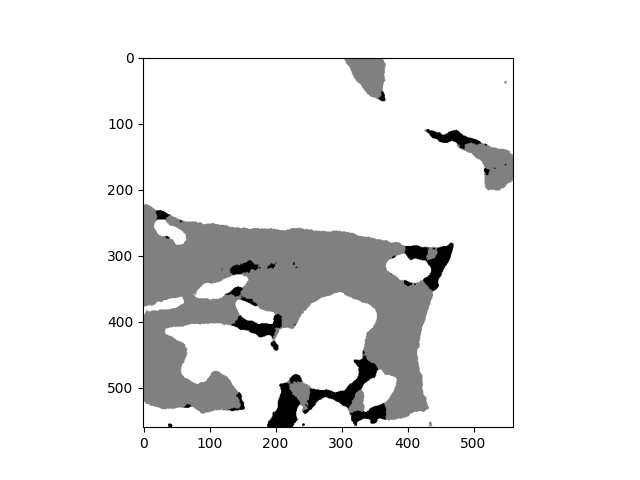}
        \end{subfigure} &
        \begin{subfigure}[t]{0.2\textwidth}
            \includegraphics[width=1.35\linewidth]{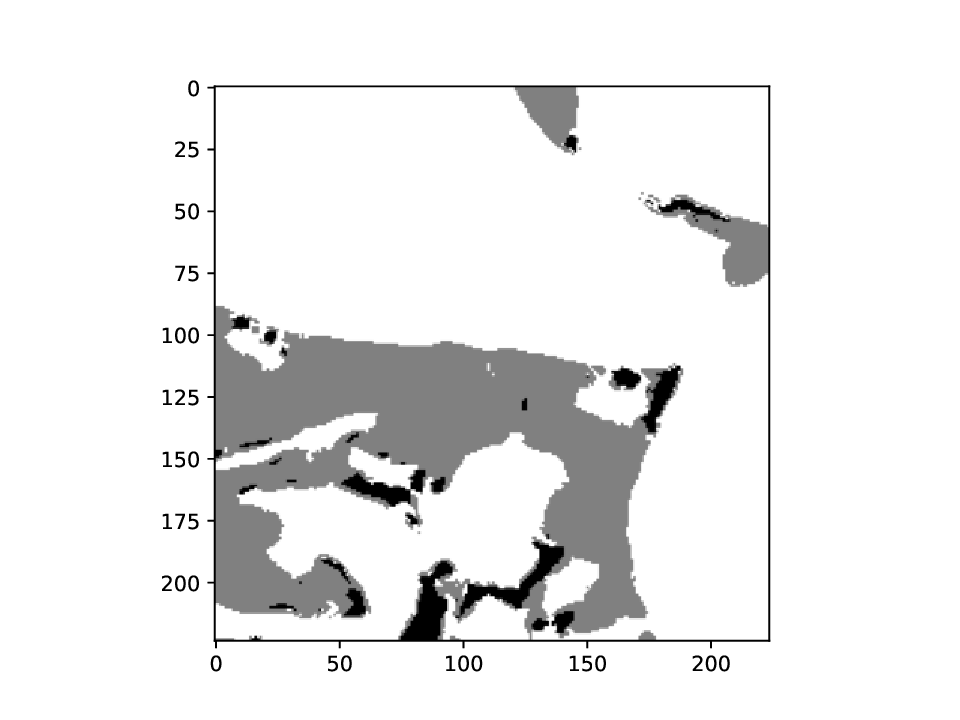}
        \end{subfigure} \\
        \begin{subfigure}[t]{0.2\textwidth}
            \includegraphics[width=1.35\linewidth]{images/appendices/bench/img.png}
        \end{subfigure} &
        \begin{subfigure}[t]{0.2\textwidth}
            \includegraphics[width=1.35\linewidth ]{images/appendices/bench/gt.png}
        \end{subfigure} &
        \begin{subfigure}[t]{0.2\textwidth}
            \includegraphics[width=1.35\linewidth ]{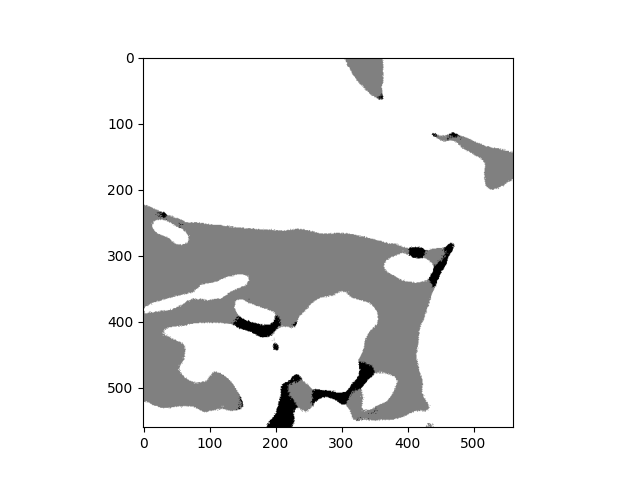}
        \end{subfigure} &
        \begin{subfigure}[t]{0.2\textwidth}
            \includegraphics[width=1.35\linewidth ]{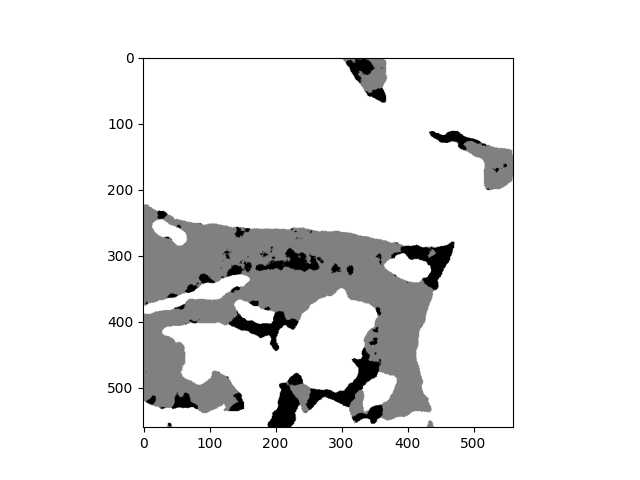}
        \end{subfigure} &
        \begin{subfigure}[t]{0.2\textwidth}
            \includegraphics[width=1.35\linewidth]{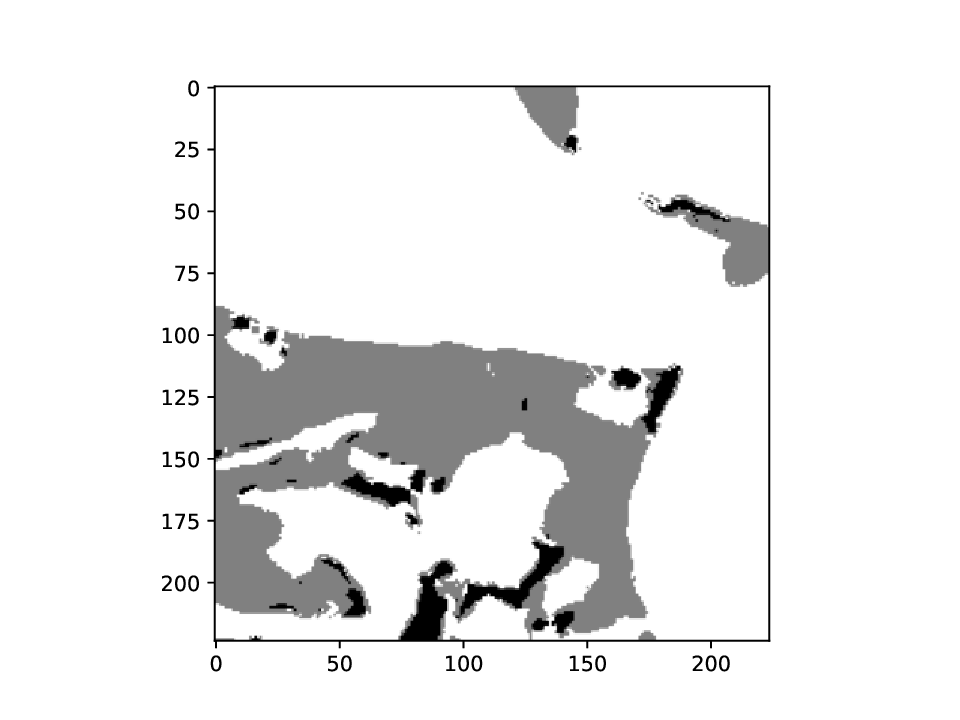}
        \end{subfigure} \\
        \begin{subfigure}[t]{0.2\textwidth}
            \includegraphics[width=1.35\linewidth ]{images/appendices/bench/img.png}
        \end{subfigure} &
        \begin{subfigure}[t]{0.2\textwidth}
            \includegraphics[width=1.35\linewidth ]{images/appendices/bench/gt.png}
        \end{subfigure} &
        \begin{subfigure}[t]{0.2\textwidth}
            \includegraphics[width=1.35\linewidth ]{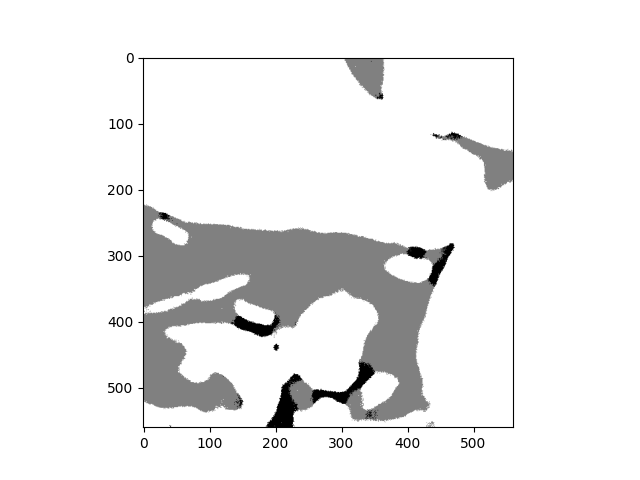}
        \end{subfigure} &
        \begin{subfigure}[t]{0.2\textwidth}
            \includegraphics[width=1.35\linewidth ]{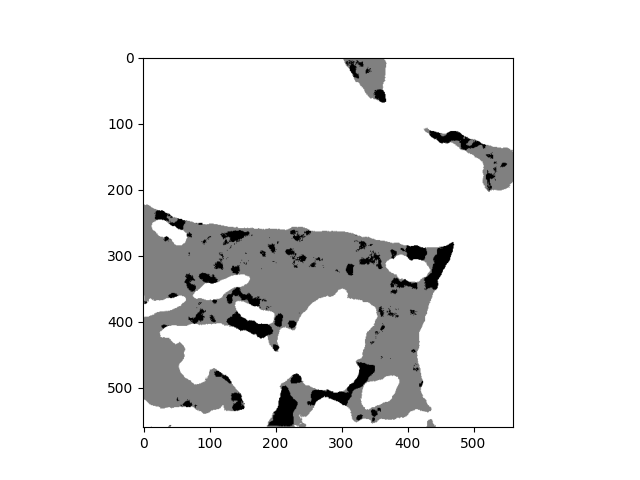}
        \end{subfigure} &
        \begin{subfigure}[t]{0.2\textwidth}
            \includegraphics[width=1.35\linewidth]{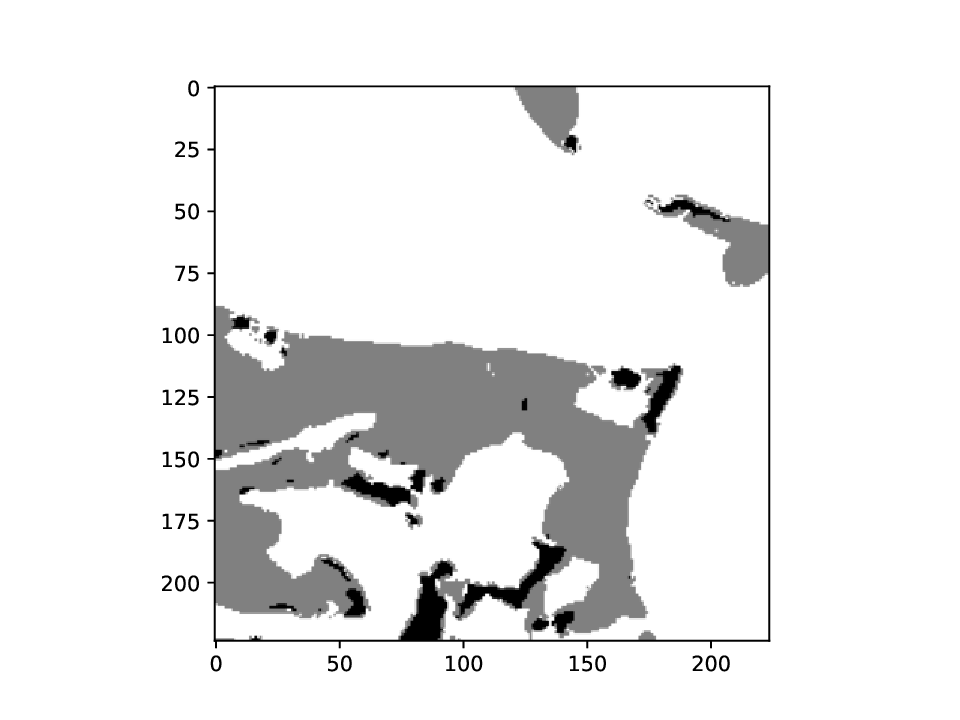}
        \end{subfigure} \\
        \begin{subfigure}[t]{0.2\textwidth}
            \includegraphics[width=1.35\linewidth ]{images/appendices/bench/img.png}
        \end{subfigure} &
        \begin{subfigure}[t]{0.2\textwidth}
            \includegraphics[width=1.35\linewidth ]{images/appendices/bench/gt.png}
        \end{subfigure} &
        \begin{subfigure}[t]{0.2\textwidth}
            \includegraphics[width=1.35\linewidth ]{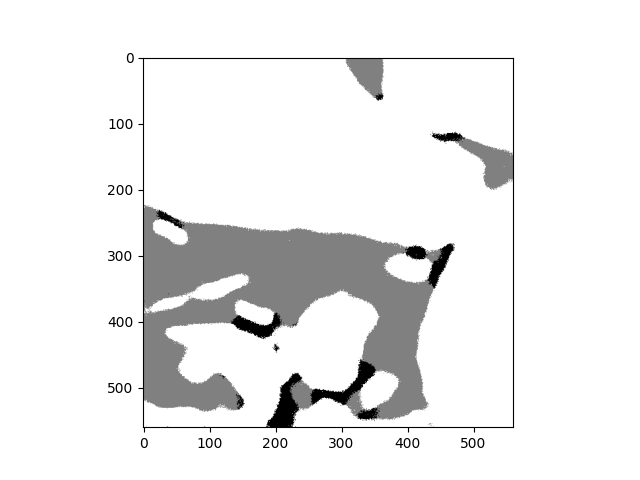}
        \end{subfigure} &
        \begin{subfigure}[t]{0.2\textwidth}
            \includegraphics[width=1.35\linewidth ]{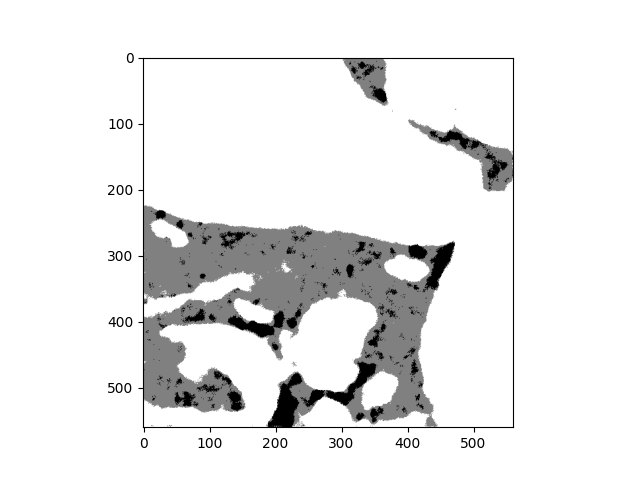}
        \end{subfigure} &
        \begin{subfigure}[t]{0.2\textwidth}
            \includegraphics[width=1.35\linewidth]{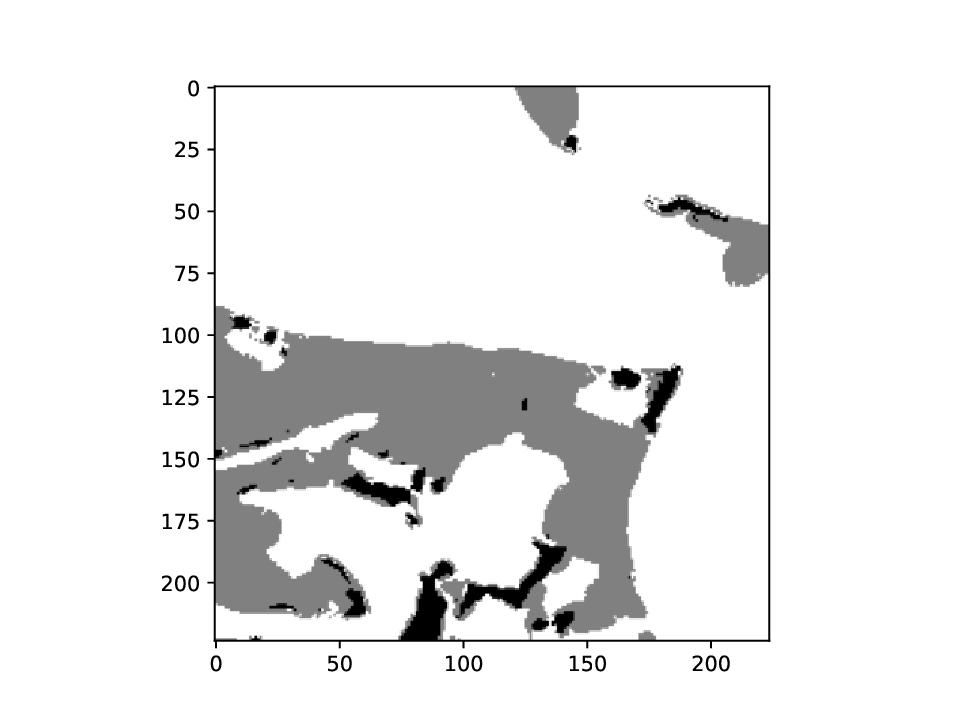}
        \end{subfigure} \\
        \begin{subfigure}[t]{0.2\textwidth}
            \includegraphics[width=1.35\linewidth ]{images/appendices/bench/img.png}
        \end{subfigure} &
        \begin{subfigure}[t]{0.2\textwidth}
            \includegraphics[width=1.35\linewidth ]{images/appendices/bench/gt.png}
        \end{subfigure} &
        \begin{subfigure}[t]{0.2\textwidth}
            \includegraphics[width=1.35\linewidth ]{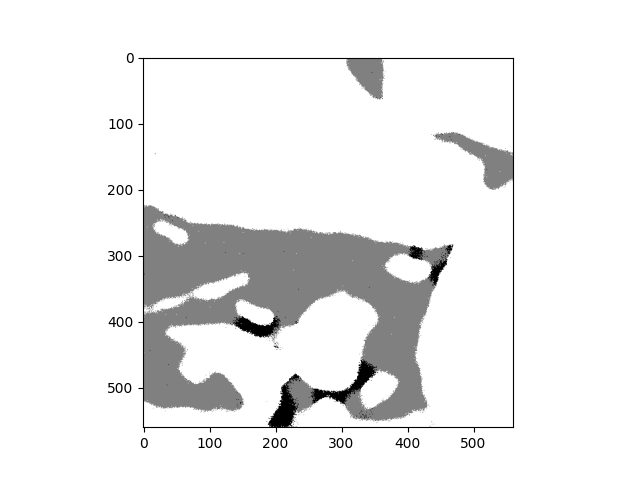}
        \end{subfigure} &
        \begin{subfigure}[t]{0.2\textwidth}
            \includegraphics[width=1.35\linewidth ]{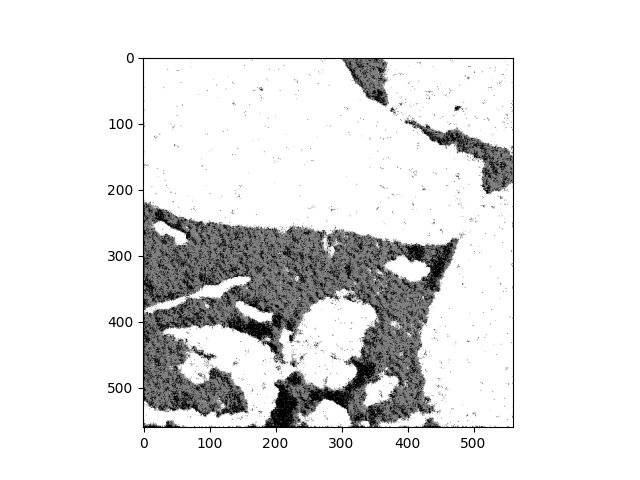}
        \end{subfigure} &
        \begin{subfigure}[t]{0.2\textwidth}
            \includegraphics[width=1.35\linewidth]{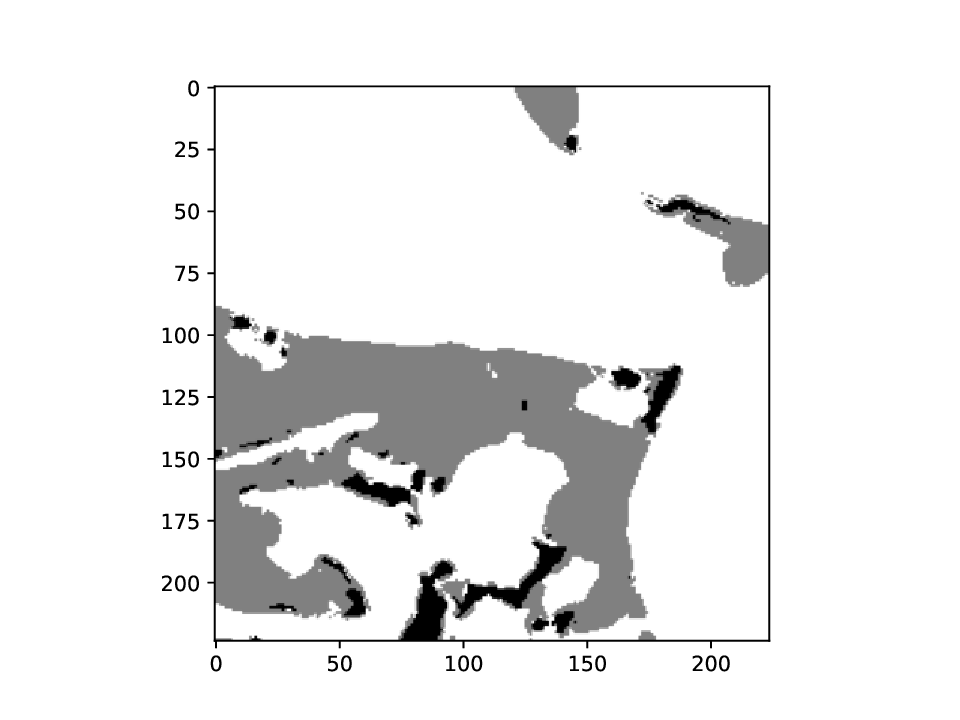}
        \end{subfigure} \\
        \begin{subfigure}[t]{0.2\textwidth}
            \includegraphics[width=1.35\linewidth ]{images/appendices/bench/img.png}
        \end{subfigure} &
        \begin{subfigure}[t]{0.2\textwidth}
            \includegraphics[width=1.35\linewidth ]{images/appendices/bench/gt.png}
        \end{subfigure} &
        \begin{subfigure}[t]{0.2\textwidth}
            \includegraphics[width=1.35\linewidth ]{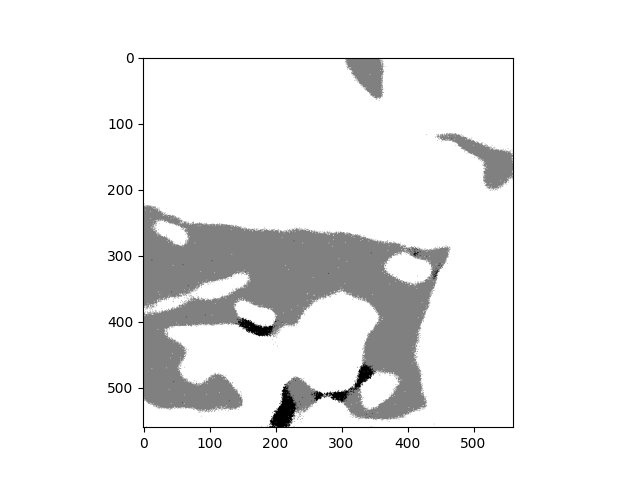}
        \end{subfigure} &
        \begin{subfigure}[t]{0.2\textwidth}
            \includegraphics[width=1.35\linewidth ]{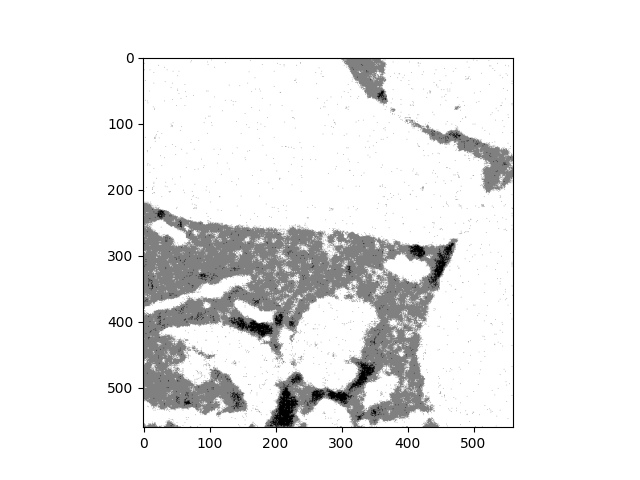}
        \end{subfigure} &
        \begin{subfigure}[t]{0.2\textwidth}
            \includegraphics[width=1.35\linewidth]{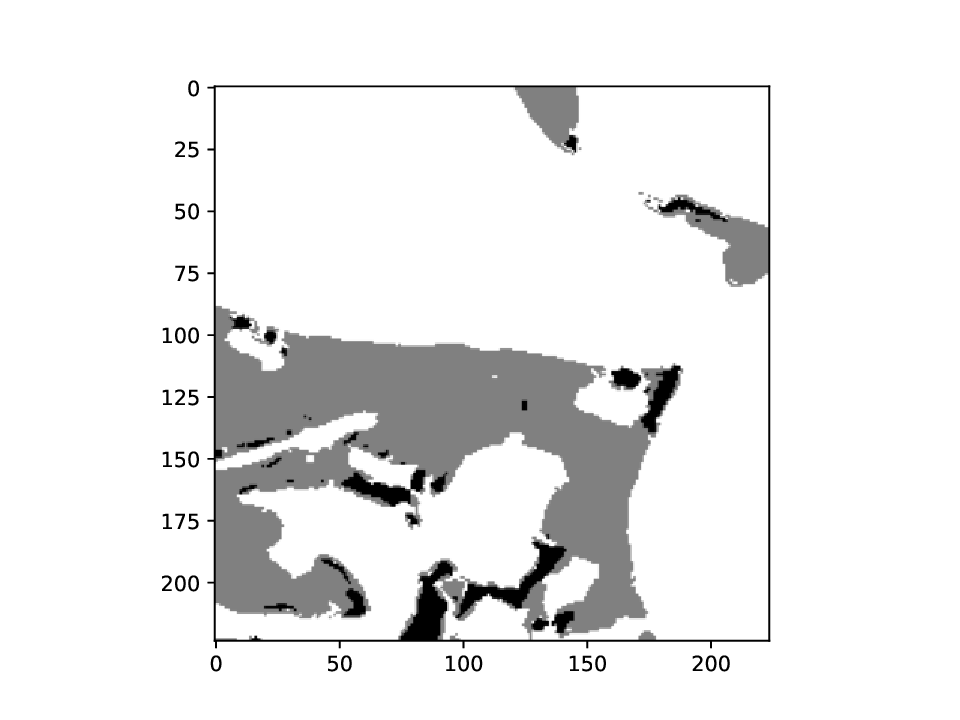}
        \end{subfigure} \\
        \begin{subfigure}[t]{0.2\textwidth}
            \includegraphics[width=1.35\linewidth ]{images/appendices/bench/img.png}
        \caption{Raw image}
        \end{subfigure} &
        \begin{subfigure}[t]{0.2\textwidth}
            \includegraphics[width=1.35\linewidth ]{images/appendices/bench/gt.png}
        \caption{GT}
        \end{subfigure} &
        \begin{subfigure}[t]{0.2\textwidth}
            \includegraphics[width=1.35\linewidth ]{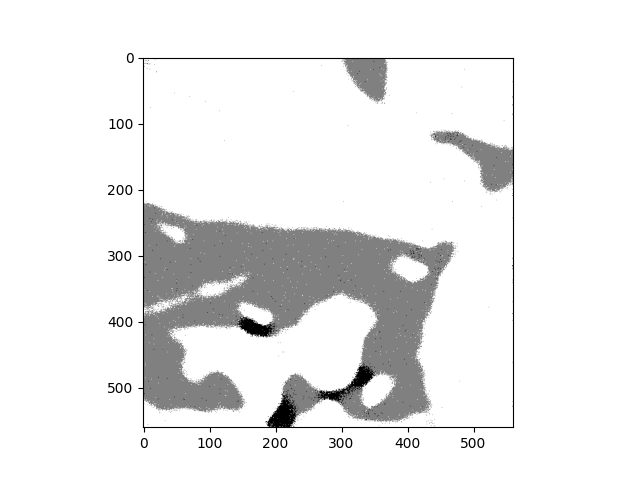}
        \caption{DINOv2}
        \end{subfigure} &
        \begin{subfigure}[t]{0.2\textwidth}
            \includegraphics[width=1.35\linewidth ]{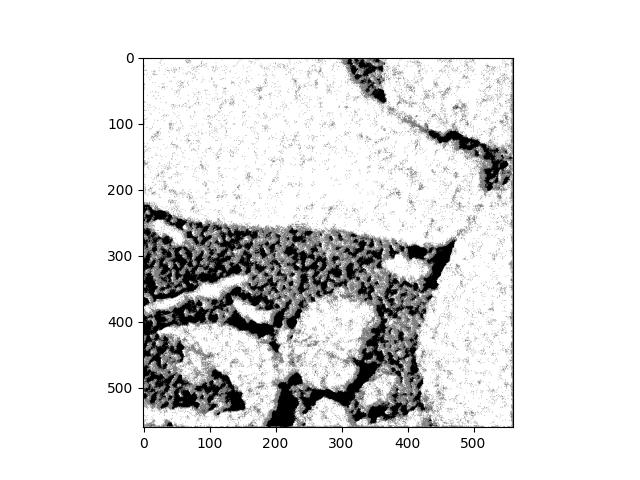}
        \caption{Unet-Small}
        \end{subfigure} &
        \begin{subfigure}[t]{0.2\textwidth}
            \includegraphics[width=1.35\linewidth]{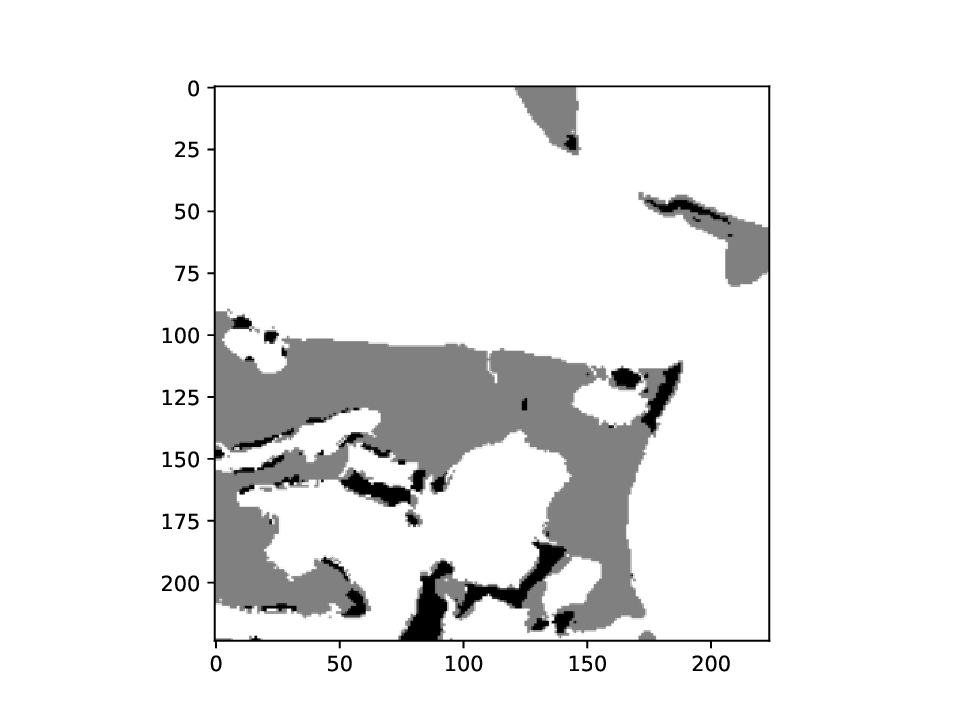}
        \caption{RF}
        \end{subfigure} \\
    \end{tabular}
    }
    \caption{Predicted segmentation mask for DINOv2 (c), UNet-Small (d) and RF (e) as a function of the number of training samples, with 1,000 training samples as the top row and 4 as the bottom row. The raw image (a) and the GT (b) are provided for comparison.}
    \label{fig:grid}
\end{figure*}

\bibliographystyle{abbrv}
\bibliography{biblio}

\end{document}